\documentclass[aps,pra,reprint,groupedaddress,nofootinbib,longbibliography]{revtex4-2}
\usepackage{graphicx}
\usepackage{dcolumn}
\usepackage{bm}
\usepackage[english]{babel}
\usepackage{amsmath}
\usepackage{amssymb}
\usepackage{hyperref}
\hypersetup{
colorlinks=true,
linkcolor=blue,
filecolor=magenta,
urlcolor=blue,
citecolor=blue,
}
\begin{document}
\title{Is the `Agent' Paradigm a Limiting Framework for \\Next-Generation Intelligent Systems?}
\author{Jesse Gardner}
\affiliation{Active Inference Institute, Crescent City, California, 95531, USA}
\author{Vladimir A. Baulin}
\affiliation{Active Inference Institute, Crescent City, California, 95531, USA}
\affiliation{Departament d'Enginyeria Química, Universitat Rovira i Virgili, Tarragona, Spain}

\date{\today}
\begin{abstract}
The concept of the `agent’ has profoundly shaped Artificial Intelligence (AI) research, guiding development from foundational theories to contemporary applications like Large Language Model (LLM)-based systems. This paper critically re-evaluates the necessity and optimality of this agent-centric paradigm. We argue that its persistent conceptual ambiguities and inherent anthropocentric biases may represent a limiting framework. We distinguish between agentic systems (AI inspired by agency, often semi-autonomous, e.g., LLM-based agents), agential systems (fully autonomous, self-producing systems, currently only biological), and non-agentic systems (tools without the impression of agency). Our analysis, based on a systematic review of relevant literature, deconstructs the agent paradigm across various AI frameworks, highlighting challenges in defining and measuring properties like autonomy and goal-directedness. We argue that the `agentic' framing of many AI systems, while heuristically useful, can be misleading and may obscure the underlying computational mechanisms, particularly in Large Language Models (LLMs). Our quantitative analysis of the field's intellectual structure reveals a significant and persistent gap between theoretical/critical discourse and practical implementation, with the most influential current debates centered on the critique of anthropocentrism itself. This suggests a paradigm under strain. As an alternative, we propose a shift in focus towards frameworks grounded in system-level dynamics, world modeling, and material intelligence. We conclude that investigating non-agentic and systemic frameworks, inspired by complex systems, biology, and unconventional computing, is essential for advancing towards robust, scalable, and potentially non-anthropomorphic forms of general intelligence. This requires not only new architectures but also a fundamental reconsideration of our understanding of intelligence itself, moving beyond the agent metaphor.
\end{abstract}
\maketitle
\section{\label{sec:intro}Introduction}
The notion of the `agent' has become a cornerstone of contemporary Artificial Intelligence (AI) research and development \cite{wooldridge_agent_1995, franklin_is_1997, huhns_multiagent_1999, burgin_systematic_nodate}. From theoretical frameworks aspiring to capture general principles of intelligent behavior \cite{friston_designing_2022, pezzulo_active_2024} to practical implementations in diverse domains, the idea of an autonomous entity pursuing goals within an environment is pervasive \cite{viswanathan_agentic_2025, acharya_agentic_2025, meyer-vitali_modular_2021}. This perspective has gained substantial traction with the advent of Large Language Models (LLMs), which have fueled the rapid growth of `Agentic AI' systems capable of complex task execution and intricate interactions \cite{wu_agentic_2025, xi_rise_2023, lu_llms_2024, yuksel_multi-ai_2024, zhuge_agent-as--judge_2024, huang_position_2024, gao_large_2024, abuelsaad_agent-e_2024}. These systems typically augment foundation models with external tools, memory, and sophisticated planning capabilities \cite{wu_agentic_2025, chen_s-agents_2024, liu_agent_2024}. Furthermore, theoretical paradigms such as Active Inference, rooted in the Free Energy Principle (FEP), provide a formal description of agents as systems that engage in self-organization and minimize surprise through continuous action and perception \cite{friston_designing_2022, millidge_applications_2021, millidge_relationship_2020, friston_pixels_2024, pezzulo_active_2024, ramstead_framework_2025, beck_dynamic_2025}. The widespread interest in, funding for, and applications of agent-based approaches, from personal assistants to scientific research and robotics, underscore their perceived value \cite{acharya_agentic_2025, viswanathan_agentic_2025, abuelsaad_agent-e_2024, chen_s-agents_2024, meindl_bridging_2022, tsividis_human-level_2021}.
Despite the notable advancements attributed to agent-based approaches \cite{tsividis_human-level_2021, abuelsaad_agent-e_2024, zhuge_agent-as--judge_2024}, the ambitious goal of Artificial General Intelligence (AGI) remains largely elusive \cite{langley_cognitive_2006, blili-hamelin_stop_2025, jaeger_artificial_2024}. This persistent elusiveness prompts a fundamental re-examination of the dominant conceptual frameworks guiding AI research. Is the `agent' a necessary construct, or a convenient yet potentially restrictive abstraction? This paper examines the hypothesis that the agent-centric paradigm, potentially influenced by anthropocentric assumptions \cite{kagan_toward_2024, friston_variational_2023, cole_cognitive_2024, papayannopoulos_computational_2022, makin_against_2023, evans_algorithmic_2024}, might represent a limiting framework. The reliance on human intuition and folk psychology \cite{wooldridge_intelligent_1995} in defining agentic properties, coupled with a focus on achieving `human-level' performance \cite{tsividis_human-level_2021, koivisto_best_2023}, risks inadvertently constraining the exploration of truly novel forms of artificial intelligence and leading to less efficient or robust solutions compared to system-level alternatives \cite{lee_what_2022}.
To clarify this discussion, we distinguish three key concepts:
\begin{itemize}
\item \textbf{Agentic} systems are AI systems inspired by true agency, exhibiting features that give the \textit{impression} of autonomous, goal-directed behavior, often through complex programming or the capabilities of LLMs. They are typically semi-autonomous and lack the deep functional autonomy of living organisms.
\item \textbf{Agential} systems are fully autonomous, self-producing, and self-maintaining systems. Currently, this category is arguably populated only by biological organisms, whose agency is intrinsic and grounded in their autopoietic nature.
\item \textbf{Non-agentic} systems are tools or processes programmed to perform tasks without giving the impression of agency. They operate based on direct instruction or fixed algorithms.
\end{itemize}
This paper argues that much of what is termed ``Agentic AI'' is a sophisticated \textit{facade}, and that the uncritical pursuit of this framing may obscure more fundamental principles of intelligence. Concerns arise as to whether current agentic AI, particularly LLM-based systems, exhibit genuine agency or merely sophisticated mimicry \cite{jaeger_artificial_2024, shanahan_simulacra_2024}. We raise a pivotal question: could robust intelligence emerge from frameworks that do not explicitly conceptualize `agents' at all, focusing instead on underlying system dynamics and continuous interactions \cite{lee_what_2022, pessoa_refocusing_2022}?
As a compelling alternative, the concept of \textit{agential systems} and \textit{non-agentic intelligence} warrants closer examination \cite{kagan_toward_2024, friston_variational_2023, cole_cognitive_2024, papayannopoulos_computational_2022}. 

This perspective shifts the emphasis to system-level properties, distributed architectures, self-organization, and emergent intelligence. Insights from complex systems theory \cite{sole_evolution_2022}, natural processes \cite{bechtel_grounding_2021, barandiaran_defining_2009, ciaunica_nested_nodate, davies_synthetic_2023}, and non-classical computational paradigms \cite{lee_what_2022, sole_evolution_2022, kozachkov_building_2023} suggest that intelligence might not be a monolithic property exclusive to discrete, centrally controlled agents. Instead, it could emerge from the intricate interactions within distributed networks \cite{bruni_towards_2015, pessoa_spiraling_2023}, the inherent dynamics of physical systems \cite{lee_what_2022}, or unconventional computing substrates exploiting material properties \cite{sole_evolution_2022, kowerdziej_softmatterbased_2022, baulin_intelligent_2025}. 

This includes principles of embodied cognition \cite{korsakova-kreyn_emotion_2021, lee_what_2022, prescott_understanding_2024, pezzulo_active_2024}, where intelligence arises from the continuous interplay and inseparable coupling between perception, action, body, and environment \cite{kluger_brainbody_2024, stern_dynamic_2022, sloan_meaning_2023}. Such approaches prioritize understanding the material basis and systemic organization from which intelligent behavior might emerge, potentially leading to more robust and efficient AI systems \cite{beckmann_alternative_2023, king_human_2021, khona_attractor_2022, singer_recurrent_2021}. 

The perspective that intelligence can arise from the physical properties and interactions of matter (Matter computes \cite{lee_what_2022, sole_evolution_2022}) fundamentally challenges agent-centric views that often implicitly prioritize abstract software and algorithms over physical embodiment and substrate-dependent dynamics. Furthermore, scientific knowledge transformed into structured, computable `world models' can be directly navigated and operated upon by AI systems using abstract mathematical and learned operations, moving beyond anthropomorphic interfaces and towards computationally grounded forms of intelligence \cite{baulin_discovery_2025}.

Given these persistent conceptual challenges and the rise of powerful new paradigms, a purely qualitative assessment is no longer sufficient. To objectively map the state of this intellectual landscape, we conducted a quantitative analysis of the field's structure and evolution. By constructing a knowledge graph from a systematic review of the literature, we analyzed the influence of core concepts, their interdisciplinary connections, and their temporal dynamics. The results, presented in Section \ref{sec:quantitative_analysis}, provide strong empirical evidence for a structural theory-practice gap and offer a data-driven roadmap for future innovation. 

AGI, if achievable, might arise from principles fundamentally different from those governing individual human agency. This paper argues that the necessity and optimality of the agent-centric paradigm for developing next-generation intelligent systems should be critically evaluated. We propose that anthropocentric biases may limit its scope and that exploring system-level, interaction-based, and materially grounded frameworks is crucial for future progress. Therefore, AI research should remain open to discovering or engineering intelligence wherever and however it emerges, rather than rigidly limiting itself to the 'agent' framework.

\section{\label{sec:deconstruct_agent}Deconstructing the `Agent': Variance, Bias, and Utility}

\subsection{The `Agent' as a Conceptual Tool: Ambiguity and Operational Limits}

The concept of an `agent' is deeply ingrained across diverse disciplines, including AI \cite{wooldridge_agent_1995, franklin_is_1997, huhns_multiagent_1999, burgin_systematic_nodate}, cognitive science \cite{langley_cognitive_2006, bello_there_2017}, philosophy \cite{jaeger_artificial_2024, barandiaran_defining_2009}, and biology \cite{bechtel_grounding_2021}. In AI, agent definitions often coalesce around functional properties such as autonomy, reactivity, pro-activeness, and social ability \cite{wooldridge_agent_1995, wooldridge_intelligent_1995, burgin_systematic_nodate}. For instance, Russell and Norvig define a rational agent as one that selects actions to maximize a performance measure given its percepts and knowledge \cite{russell_artificial_2010}. Others emphasize situatedness and the pursuit of an `agenda' \cite{franklin_is_1997}. Some researchers adopt a stronger notion, imbuing agents with human-like mentalistic concepts such as knowledge, belief, desire, and intention (BDI) \cite{wooldridge_agent_1995, wooldridge_intelligent_1995}, often drawing upon Dennett's intentional stance as an explanatory tool \cite{kenton_discovering_2022, wooldridge_intelligent_1995}.
However, a systematic review reveals a persistent lack of consensus and consistency in these definitions \cite{huhns_multiagent_1999, wooldridge_intelligent_1995, franklin_is_1997, serenko_intelligent_2004}. Definitions diverge significantly depending on the research context, disciplinary background, and intended application \cite{burgin_systematic_nodate, srinivasa_paradigms_2020}. When common characteristics are categorized, they span a wide spectrum of autonomy (from simple stimulus-response to complex deliberation), various types of agency (reactive, proactive, social), diverse goal structures, implicit or explicit boundaries, and differing requirements for internal models \cite{wooldridge_agent_1995, franklin_is_1997}. Attempts to create broad, formal definitions often lead to counterintuitive inclusions, such as thermostats or even soap bubbles potentially qualifying under certain criteria \cite{franklin_is_1997, biehl_formal_2017}, underscoring the challenge of capturing the intuitive notion of agency formally and universally. A critical concern is the low operational measurability of many definitions, particularly those relying on abstract concepts like autonomy, intention, or rationality \cite{biehl_formal_2017, kagan_toward_2024, friston_variational_2023, cole_cognitive_2024, papayannopoulos_computational_2022, makin_against_2023, evans_algorithmic_2024}. The objective measurement of such internal states or human-like cognitive attributes poses particular difficulties for empirical verification and comparative evaluation of AI systems.
Underlying these definitions are often implicit assumptions such as principles of rational choice (maximizing utility or performance measures) \cite{bello_there_2017, kenton_discovering_2022}, the necessity of internal symbolic models for deliberation \cite{langley_cognitive_2006, wooldridge_intelligent_1995}, and reward-based action selection \cite{millidge_relationship_2020, seifert_reinforcement_2024}. These assumptions, frequently derived from human economic or cognitive models, are often adopted without sufficient critical evaluation of their fundamental necessity or their direct applicability to artificial systems \cite{kagan_toward_2024, friston_variational_2023, cole_cognitive_2024}. Critics highlight limitations, such as the problem of logical omniscience in formal agent theories (where an agent is presumed to know all logical consequences of its beliefs) \cite{wooldridge_intelligent_1995}, and the unsuitability of applying the intentional stance to overly simple systems (e.g., thermostats) \cite{wooldridge_intelligent_1995}. The pervasive focus on human-like rationality has also been questioned, leading to explorations of bounded rationality that account for computational constraints \cite{jacob_modeling_2023, lee_what_2022} and alternative, non-classical reasoning mechanisms \cite{ahmed_abdel-fattah_tarek_r_besold_helmar_gust_ulf_krumnack_martin_schmidt_kai-uwe_kuhnberger_rationality-guided_2012}. The absence of definitional consensus and the reliance on potentially anthropocentric, untestable assumptions raise serious concerns about the `agent' concept's foundational robustness as a primary tool for AI research aiming for general intelligence \cite{huhns_multiagent_1999, wooldridge_intelligent_1995, kagan_toward_2024}. This suggests that defining 'agents' might often serve as a convenient human interpretive framework rather than a rigorously verified scientific necessity for understanding all forms of intelligence.

A critical consideration is the perceived `division' of functionality into discrete sensory and perception modules. The prevalent assumption of such division is rooted in human conceptual constructs that seek to easily identify an object that has borders (analogous to the conversation about the necessity of the Markov blanket\cite{raja_markov_2021}). It challenges the notion that partitioning functionality into separate sensory and perceptual modules is a scientific necessity for AI design. Instead, it might be an oversimplification influenced by how humans naturally parse the world into bounded objects. What are explicitly being computed in these AI systems are often operations within intricate, high-dimensional tensor spaces, fundamentally abstracting away human attributes with more specific mathematical definitions \cite{baulin_discovery_2025}. The pervasive use of intentional language like `beliefs' or `planning,' while heuristically intuitive, provides no additional mathematical or mechanistic specificity to these tensor-based computations. Indeed, it can obscure the true nature of the underlying computational mechanisms and biases the direction of research towards trying to literally engineer ambiguous mental states rather than optimizing rigorous mathematical operations that govern system dynamics, irrespective of anthropomorphic semantic interpretations.

The significance of this conceptual ambiguity and its inherent anthropocentric bias is not merely a philosophical concern; as our quantitative analysis reveals, it has become the central organizing tension of the entire field.

Ultimately, the operational core of contemporary AI, from LLMs to reinforcement learning systems, is mathematical: the manipulation of high-dimensional tensors. These tensors, comprising the weights of neural networks, serve as the most convenient and efficient representation for a system’s knowledge base, internal states, and its learned world model \cite{baulin_discovery_2025}. From this perspective, anthropocentric properties often associated with `agents'—such as `beliefs,' `goals,' or `intentions'—are best understood as high-level, interpretive glosses on these underlying mathematical operations. While such agentic language can be a useful heuristic, it risks obscuring the fundamental computational nature of these systems and can introduce unnecessary complexity. This tension is evident in the relationship between Active Inference (AIF) and more direct forms of model-based reinforcement learning (RL) like Control as Inference (CAI) \cite{millidge_relationship_2020}. While both frameworks can be shown to be mathematically related, with AIF being equivalent to a specific form of RL under certain assumptions, the practical implementation differs. AIF’s explicit encoding of `beliefs' and `preferences' as priors within its generative model can be less computationally efficient than RL approaches that learn a policy more directly from a world model’s outputs, a point of significant practical debate for complex engineering applications \cite{friston_variational_2023, cole_cognitive_2024}. This suggests that stripping away the rich, human-like cognitive `scaffolding' and focusing on the core tensor-based world model and policy optimization can yield more direct and efficient paths to adaptive behavior, reinforcing the view that the `agent' metaphor may not always be the most effective one for engineering intelligent systems \cite{ha_world_2018, liu_reinforcement_2020}.

\subsection{Active Inference as an Agent-Centric Framework: A Case Study}
Active Inference (AIF), building upon the Free Energy Principle (FEP), stands as a prominent theoretical framework for modeling agents and cognition \cite{friston_designing_2022, millidge_applications_2021, pezzulo_active_2024}. It posits that agents, to maintain themselves far from thermodynamic equilibrium, act and perceive to minimize variational free energy, effectively maximizing evidence for their internal generative models of the world or minimizing their `surprise' \cite{millidge_relationship_2020, friston_pixels_2024}. This framework offers an elegant and unifying description of perception, learning, and action under a single inferential objective \cite{millidge_applications_2021, pezzulo_active_2024}.
However, applying AIF as a model of agency often introduces potential anthropocentric biases and conceptual limitations. While the FEP formally seeks generality rooted in physical principles, the framework's interpretation frequently employs human-like cognitive language, such as `beliefs,' `preferences,' `curiosity,' and `goals,' all implicitly or explicitly encoded within the generative model \cite{friston_designing_2022, millidge_relationship_2020, ramstead_framework_2025}. Describing systems as if they are literally performing inference or maximizing evidence \cite{ramstead_framework_2025} risks falling into the map-territory fallacy. This means attributing cognitive processes (the `map') to a system's internal dynamics (the `territory') where only complex statistical or physical mechanisms might exist \cite{ramstead_framework_2025, friston_variational_2023}. For instance, a brain operating as a `black box' could simply produce actions without needing conscious calculation of value, akin to chaotic systems or the human intuition's processing \cite{dutta_how_2024}. Moreover, the core concept of the Markov blanket, while formally defining a statistical boundary (where blanket states render internal and external states conditionally independent), has faced criticism regarding its universal applicability to dynamic, open systems with transient or porous boundaries \cite{raja_markov_2021,beck_dynamic_2025}. Its utility in robustly defining `objects' or `agents' from microscopic dynamics universally, especially when material flux across these boundaries is high, remains a subject of ongoing debate \cite{biehl_formal_2017, beck_dynamic_2025, ciaunica_nested_nodate}. This is related to the common observation that dynamic, non-equilibrium systems often defy clear borders.
Furthermore, the emphasis on minimizing free energy as the agent's fundamental drive, while theoretically appealing and elegant \cite{friston_variational_2023}, may introduce conceptual constraints or incur unnecessary computational costs compared to alternative approaches, such as reinforcement learning (RL) for achieving specific functionalities \cite{friston_variational_2023, cole_cognitive_2024}. Critiques suggest that for complex, high-dimensional problems, the practical efficiency and scalability of AIF remain active areas of investigation and potential limitation when compared to established RL methods \cite{friston_variational_2023, cole_cognitive_2024}. While proponents argue for AIF's potential advantages in sample efficiency and biological plausibility \cite{friston_pixels_2024, millidge_applications_2021}, the debate continues as to whether the agent-centric, inference-based framing of AIF is strictly necessary for achieving adaptive behavior, or if simpler, non-agentic approaches might suffice or prove more efficient in certain contexts. For example, some formulations of RL are primarily input-output functions (policies) where an agent's rationality is inferred, rather than being explicitly encoded via a generative model \cite{liu_reinforcement_2020, ha_world_2018, chung_thinker_2023}. Thus, the theoretical elegance of AIF may come at the cost of engineering practicality and computational efficiency \cite{friston_variational_2023}.
In the context of \textit{bounded rationality} (where computation is limited), agents make decisions under constraints of limited information, computational resources, and time. The dependence on continuous optimization may fail to accurately represent `optimal' behavior. For instance, models of human decision-making involving latent inference budgets \cite{jacob_modeling_2023} implicitly contrast with purely `optimal' Bayesian inference. Moreover, simplifying assumptions such as linearizing dynamics can severely limit model applicability, potentially leading to unrealistic or inaccurate predictions in complex, chaotic systems where non-linear effects and correlations are crucial \cite{bruckner_information_2023}. In such cases, linearizations implicitly ignore crucial feedback loops and emergent non-equilibrium properties that are not at steady state or may exhibit high order variability \cite{beck_dynamic_2025, friston_pixels_2024, ramstead_framework_2025}. Such models only accurately describe dynamically stable systems where ``nothing happens'' to avoid any form of complex, higher order variability such as chaotic behavior \cite{beck_how_2024}. When certain human behaviors are caricatured as anthropomorphic traits \cite{korsakova-kreyn_emotion_2021}, AI is often programmed to mimic this biased view which can in turn encode stereotypes given such linearity and a biased interpretation of bounded rationality from equations of information optimization such as KL divergence for neural networks that is also often used for AI agents to make certain inferences in uncertain conditions of bounded parameters \cite{hoffman_stochastic_nodate, liu_stein_2019}. This affects the generalizability of models that need to compute effectively in physical world settings where chaotic phenomena, and the resulting fluctuations it is sensitive to; means the system's optimal control will fall prey to unintended behaviors if such biases persist.
\subsection{LLMs as `Agents': Unveiling the Agentic Facade}
The meteoric rise of LLMs has profoundly reshaped AI research, driving the development of `agentic' AI systems designed to plan, execute, and adapt complex tasks using tools and interactions with the real world \cite{wu_agentic_2025, xi_rise_2023, gao_large_2024, abuelsaad_agent-e_2024, acharya_agentic_2025}. Systems like AutoGPT, or frameworks employing LLMs for automated reasoning and tool-use \cite{wu_agentic_2025, zhuge_agent-as--judge_2024, yuksel_multi-ai_2024, chen_s-agents_2024, liu_agent_2024} are frequently and naturally described in agentic terms, often using anthropomorphic language to explain their decision-making or planning capabilities \cite{huang_position_2024}.
However, the very notion of `agency' in these LLM-based systems remains highly debated. Is it merely a convenient metaphor, perhaps driven by user interface design needs and marketing strategies \cite{kagan_toward_2024, friston_variational_2023, cole_cognitive_2024, papayannopoulos_computational_2022, walter_artificial_2024}, or does it genuinely reflect intrinsic agentic capabilities in these LLM-based systems? Critics strongly argue that current LLM-based systems lack true autonomy, intrinsic long-term planning, or goal-directedness independent of the prompts they receive. Instead, their impressive performance may represent sophisticated ``algorithmic mimicry'' or merely be a ``simulacra'' of agency, rather than genuine, robust agency \cite{jaeger_artificial_2024, shanahan_simulacra_2024} derived from vast implicit `world models' learned during pre-training (e.g., through predicting next tokens based on correlations within massive datasets) \cite{dutta_how_2024, ha_world_2018, robine_transformer-based_2023, xie_making_2025, pan_what_2023, taniguchi_world_2023}. Consequently, attributing qualities such as `thinking' or `understanding' to these systems might promote misleading anthropomorphic interpretations \cite{kagan_toward_2024, friston_variational_2023, cole_cognitive_2024, shanahan_simulacra_2024}, obscuring their actual mechanisms and potentially misdirecting research efforts \cite{jaeger_artificial_2024}.
Viewing LLMs simply as powerful pattern-completion systems, or as modular components within a larger computational framework \cite{lee_what_2022}, could offer a less anthropocentric and potentially more accurate description. Fundamentally, LLMs are complex neural networks that operate as sophisticated tensor transformation machines, processing high-dimensional vector embeddings of text through layers of matrix multiplications and non-linear activations \cite{baulin_discovery_2025}. While they demonstrate impressive `reasoning' and `planning' capabilities, these are effectively emergent properties of pattern recognition and semantic interpolation within vast tensor spaces. Viewing them as `agents' rather than advanced mathematical instruments for pattern mapping may inadvertently mask their true mechanistic limitations and biases future AI research, diverting attention from the underlying power and properties of tensor-based computation itself. In this light, the current wave of `Agentic AI' represents effective engineering that leverages the immense statistical capabilities of LLMs to create \textit{behaviors that appear agent-like} on discrete, bounded problems, often in virtual environments (like code generation, data analysis). Their success is in mimicking intelligence, providing an agentic facade that users find intuitive, but it does not resolve the fundamental conceptual ambiguities around true agency. Critiques often point to how their impressive capabilities still rely on fixed data-manifold environments where performance metrics remain somewhat simplified \cite{robine_transformer-based_2023, jacob_modeling_2023}. Concerns arise regarding potential harms when anthropomorphic framing of AI systems leads to misplaced trust or ambiguous accountability, especially for increasingly autonomous AI systems \cite{chan_harms_2023, chan_harms_2023-1, mukherjee_agentic_2025, iason_gabriel_et_al_ethics_2024, park_ai_2024, shavit_practices_nodate}. If a machine is deemed an agent, what are the ``moral crumple zones'' of responsibility if things go wrong \cite{chan_harms_2023}? Does optimization (Goodhart's Law) or increased efficiency (Jevons Paradox) introduce new problems if systems become too capable in narrowly defined `small worlds' without holistic ethical alignment \cite{chan_harms_2023, beck_how_2024}? These systems also raise challenges in terms of governance due to the complexity of defining roles and attributing responsibility in situations involving distributed and opaque interactions \cite{mukherjee_agentic_2025}. A deeper concern remains that an exclusive focus on current agentic LLM models, which may exhibit emergent properties within a textual `small world' \cite{jaeger_artificial_2024} but lack real-world groundedness, might limit exploration of fundamentally different intelligence paradigms more relevant to true intelligence \cite{kagan_toward_2024, friston_variational_2023, cole_cognitive_2024}.

\subsection{Borderline Cases and Distributed Conceptualizations}

The difficulties in definitively bounding and characterizing `agents' are acutely highlighted by borderline cases where existing definitions falter. While a thermostat technically fits some broad definitions based on sensing and acting \cite{franklin_is_1997}, applying concepts of intentionality, beliefs, or desires feels intuitively inappropriate and ultimately unhelpful \cite{wooldridge_intelligent_1995, biehl_formal_2017}. Similarly, biological entities spanning a vast spectrum, from single-celled bacteria exhibiting chemotaxis \cite{barandiaran_defining_2009, bechtel_grounding_2021, seifert_reinforcement_2024} to complex multicellular organisms and their immune systems (where cell-to-cell negotiations maintain homeostasis across nested `selves' \cite{ciaunica_nested_nodate}), can be abstractly framed as agents pursuing self-maintenance and goal-directedness \cite{barandiaran_defining_2009, bechtel_grounding_2021, seifert_reinforcement_2024}. Yet, directly attributing human-like agency or rich cognition to these systems risks conceptual oversimplification and problematic anthropomorphism \cite{korsakova-kreyn_emotion_2021, kagan_toward_2024}. Concepts like an infant's developing coordination dynamics hinting at agency require careful justification beyond simplistic behavioral interpretations \cite{sloan_meaning_2023}.
Some proposed agent architectures inherently challenge a monolithic, singular view of an agent, pushing towards more distributed models of intelligence. Layered architectures for agents \cite{wooldridge_intelligent_1995, franklin_is_1997} or systems explicitly composed of sub-agents with specialized, interacting functions \cite{franklin_is_1997, meyer-vitali_modular_2021} raise questions about where the `agency' truly resides. Is it a property of the overall system, or does it belong to the individual components, especially if sub-agents have conflicting goals, bounded autonomy, or operate in parallel, heterogeneous ways? Marvin Minsky's seminal ``Society of Mind'' concept also posits that ``mental agents'' within the mind might lack full autonomy, acting as semi-independent specialized processing units \cite{franklin_is_1997}. More radically, frameworks rooted in ``heterarchy'' explicitly model control and decision-making as dynamically distributed across non-hierarchically organized mechanisms, fundamentally moving away from a central `agent' or fixed command structures \cite{bruni_towards_2015, bechtel_grounding_2021}. These examples suggest that overly agent-centric views might oversimplify complex systems where intelligence and goal-directedness are distributed, emergent, or arise from the fluid interaction between loosely coupled components, potentially overlooking crucial system-level dynamics and obscuring a deeper understanding of underlying intelligence principles \cite{shavit_practices_nodate, shanahan_role_2023, prescott_understanding_2024}. This conceptual shift necessitates a move from analyzing the isolated properties of an individual `agent' to understanding the emergent behavior arising from the complex organization and continuous interaction within a system \cite{bruni_towards_2015, huhns_multiagent_1999}. The debate around how biological brains truly function also reflects this tension, suggesting that complex behaviors emerge from the dynamic interaction of various neural mechanisms, from population dynamics to microscale ion channel plasticity, challenging reduction to individual agents \cite{rolls_brain_2023, freal_sodium_2023}.
\section{\label{sec:agential_systems}Agential Systems and Systemic Intelligence: An Alternative Framing}
\subsection{From Discrete Agents to System-Level Dynamics}
Given the persistent ambiguities and potential anthropocentric biases often inherent in the `agent' concept, a fundamental shift towards alternative conceptual frameworks is warranted. Establishing clear criteria for distinguishing agentic and agential conceptualizations is challenging, but attempts include considering the locus of control (centralized vs. distributed), the nature of goals (pre-defined vs. emergent), the primacy of internal representations (explicit vs. implicit), and the level of analysis (individual entity vs. system dynamics) \cite{barandiaran_defining_2009, huhns_multiagent_1999}. However, empirically operationalizing these distinctions rigorously remains difficult \cite{kenton_discovering_2022, biehl_formal_2017}. The enduring ambiguity of the term 'agent' itself, which lacks a universally accepted definition, compounds the problem \cite{wooldridge_intelligent_1995, franklin_is_1997}, hindering clear scientific communication and rigorous comparative evaluation \cite{huhns_multiagent_1999}. Some researchers propose behavioral tests for `agenthood' \cite{huhns_multiagent_1999}, while others seek more formal definitions based on causal influence \cite{kenton_discovering_2022, biehl_formal_2017}.
The concept of `agential systems,' while promising due to its potential to reduce anthropocentric biases, also faces definitional ambiguities. Are they fundamentally distinct from merely scaled-up multi-agent systems, or is the distinction largely a matter of analytical focus \cite{kagan_toward_2024, friston_variational_2023, cole_cognitive_2024}? Frameworks like TAME (Technological Approach to Mind Everywhere) \cite{seifert_reinforcement_2024} attempt to bridge scales of agency (e.g., from cells to organisms), while others focus on specific functional aspects like attention \cite{bello_there_2017} or metacognition \cite{lewis_reflective_2024, comay_metacognition_2024} across different organizational levels. To achieve genuine conceptual clarity, moving beyond purely theoretical debate toward empirical investigation and clearer operational definitions is critical \cite{kenton_discovering_2022, meindl_bridging_2022, pan_what_2023, zhuge_agent-as--judge_2024}. A ``differentiated theory'' approach, acknowledging the complexity and multiple facets of agency and system behavior across various scales, might be more productive than striving for a single, monolithic definition \cite{lewis_reflective_2024}. Ultimately, such clarity necessitates a careful distinction between descriptive language (which may be heuristic), theoretical models, and the actual underlying mechanisms and computations at play in both natural and artificial intelligent systems \cite{ramstead_framework_2025}.
We propose a re-emphasis on `agential systems' as a distinct conceptualization, prioritizing system-level properties and emergent behavior over strictly individualistic, localized agency \cite{kagan_toward_2024, friston_variational_2023, cole_cognitive_2024, papayannopoulos_computational_2022}. Whereas `agents' are typically conceived as discrete, bounded entities possessing identifiable goals, internal models, and explicit action capabilities \cite{wooldridge_agent_1995, franklin_is_1997}, the `agential systems' perspective redirects the analytical focus to the collective dynamics and organizational principles from which intelligent or agency-like behavior might spontaneously arise \cite{bruni_towards_2015, bechtel_grounding_2021, barandiaran_defining_2009}.
This perspective actively resists viewing complex computational processes or distributed system properties solely through the reductive lens of individual `agents' acting as pre-defined units within a system. Instead, it embraces the possibility that intelligence, or even `agency' itself, is an emergent property resulting from the intricate interplay of numerous components, complex network structures, and continuous environmental interactions, rather than residing solely within or being exclusively attributable to discrete agentic units \cite{lee_what_2022, pessoa_spiraling_2023, chris_fields_donald_d_hoffman_chetan_prakash_robert_prentner_eigenforms_2017}. This approach is congruent with insights from complex systems theory, where emergent, system-level properties frequently cannot be linearly reduced to the sum of their individual component properties \cite{sole_evolution_2022}. The distinction is not merely semantic; it signifies a different ontological and epistemological stance, giving precedence to relational and systemic explanations over individualistic ones \cite{barandiaran_defining_2009, bruni_towards_2015, rabiza_point_2022}. Such an approach often involves analyzing the modularity of sensory perception components, associating pairwise compositional couplings of sensor-actuator functionality as nodes in a dynamical adaptive network which challenges prevalent anthropocentric conceptual models to which human brains would assume such modularity, but rather focus on its underlying mechanisms in neural states for sensory processing. This framing encourages looking for dynamic, evolving couplings where sensing and action are fundamentally intertwined rather than separated. However, this conceptual reorientation does not come without its own set of challenges, particularly in defining the boundaries of `agential systems' themselves, precisely measuring their emergent properties, and attributing functionality when control is highly distributed \cite{kagan_toward_2024, friston_variational_2023, cole_cognitive_2024}. The central question remains: do `agential systems' represent a fundamentally different paradigm, or simply a scaled-up, more nuanced perspective of multi-agent systems, acknowledging emergence?
This proposed conceptual shift from `agents' to `agential systems' aligns with our quantitative findings, which show that concepts of systemic and emergent intelligence are located at the most dynamic and interdisciplinary frontiers of current research.
\subsection{Capturing Agency through System Properties: Models with Sub-Optimal Agents}
The power of a systemic view is particularly evident when we move beyond the pursuit of individually optimal AI agents. This shift embraces the reality of bounded rationality, where agents operate with limited computational resources, incomplete information, or sub-optimal behavioral rules. Their true power often emerges when these agents operate \textit{en masse}, creating complex adaptive systems where sophisticated collective intelligence arises from simple, local interaction rules---rather than centralized control or explicit global knowledge \cite{gregor_self-organizing_2020}. This paradigm, inspired by biological systems like ant colonies \cite{theraulaz_spatial_2002} or schooling fish \cite{couzin_collective_2009, mugica_scale-free_2022, puy_perceived_2024, puy_selective_2024}, allows for the spontaneous formation of intricate spatial patterns, robust collective locomotion, and emergent problem-solving capabilities from components that are individually sub-optimal \cite{rubenstein_programmable_2014, march-pons_honeybee-like_2024, wang_robo-matter_2024}.
For instance, physically embodied micromotors or robots operating with basic self-propulsion and local interactions can self-organize into dynamic structures, achieve collective movement, or adapt to environments in ways a single, precisely controlled unit might struggle with \cite{hu_small-scale_2018, ceylan_mobile_2017, goh_alignment-induced_2025, goh_noisy_2022, negi_emergent_2022, iyer_directed_2024}. Even engineered biological constructs like Xenobots or Anthrobots, composed of simple cells following basic biological rules, can collectively exhibit complex emergent behaviors such as locomotion, object manipulation, or self-repair \cite{kriegman_scalable_2020, gumuskaya_motile_2024, levin_multiscale_2024, sole_open_2024}. This paradigm shift from focusing on individual perfection to harnessing distributed sub-optimality offers a promising avenue for capturing the inherent non-linearity and adaptability of complex real-world systems, often exceeding the capabilities of highly centralized, perfectly rational single-agent designs \cite{schmickl_how_2016, zhang_classical_2025}. By embracing the notion that individuals are boundedly rational \cite{jacob_modeling_2023, ahmed_abdel-fattah_tarek_r_besold_helmar_gust_ulf_krumnack_martin_schmidt_kai-uwe_kuhnberger_rationality-guided_2012}, or even sub-optimal in their individual behaviors to achieve specific features for particular functions, such as code generation validated by human expert-level synthesis, these approaches pave a middle ground where intelligence, often manifesting at the system level, is still constructed from agents, but acknowledges that optimal system performance does not demand optimality from its component units \cite{schmickl_how_2016, zhang_classical_2025}. Such systems with high-order dimensionality for physical environments, if linearized or under-modeled by simplified representations for agent behavior, would struggle and eventually yield results inconsistent with reality if the measured phenomena do not conform to the principles of a simplified bounded framework of operation in natural or artificial intelligence ecosystems where these complex components and their interactions may not always be easily measurable with exact numbers, but within qualitative standards from experts themselves for which such behavior and emergent qualities are taken to reflect intelligence \cite{lehman_machine_2023, celikok_interactive_2019, capouskova_integration_2023, galakhova_evolution_2022, song_large-scale_2022}.
\section{\label{sec:reframe}Re-framing AI Research: World Models, Interaction, and Materiality}
\subsection{The Primacy of World Models and Continuous Interaction}

A key element often associated with advanced intelligence, whether conceptualized agentically or systemically, is the possession and utilization of internal world models. Rather than being an exclusive attribute of discrete agents, such internal representations'' can be seen as implicit in the dynamic structure of the entire system, learned and refined through ongoing, continuous interaction with its environment \cite{lee_what_2022, zhang_memory_2024, cullen_internal_2023}. Neural systems, for example, exhibit hierarchical processing and dynamic representations that appear to encode and actively manage knowledge structures of the external world, such as seen in human visual cortex processing \cite{king_human_2021} or in the dynamics of recurrent cortical circuits \cite{singer_recurrent_2021, capouskova_integration_2023, song_large-scale_2022, freal_sodium_2023}, and the continuous construction of neural representations of situations and mental states via summation of action affordances \cite{thornton_neural_2024, zhang_m_l_levy_j_dascoli_s_rapin_j_alario_f-x_bourdillon_p_pinet_s__king_j_r_thought_2025}. 

The remarkable performance of LLMs often stems from the vast, implicit world models encoded in their parameters, learned during their training as complex prediction machines \cite{dutta_how_2024, pan_what_2023}. Their apparent `agentic' capabilities can be interpreted as behavior arising from reasoning over these internal representations, even if the model itself is not a traditional agent \cite{xie_making_2025, ha_world_2018, robine_transformer-based_2023}.
A crucial paradigm shift entails prioritizing research on how systems learn, represent, and utilize these world models through continuous, embodied interaction, rather than exclusively focusing on pre-programmed or agentic goal-seeking mechanisms that might presuppose such models. The Discovery Engine (DE) framework\cite{baulin_discovery_2025}, for instance, explicitly aims to transform scientific literature into a structured, computable `Conceptual Nexus Model' (CNM) that serves as a dynamic `World Model' of a scientific domain. This CNM, built from verifiable knowledge artifacts represented as a high-dimensional Conceptual Nexus Tensor, enables AI agents to interact directly with a mathematically grounded representation of knowledge, allowing for synthesis, gap analysis, and hypothesis generation beyond superficial semantic understanding \cite{baulin_discovery_2025}. 

The concept of an internal sensorimotor model, where feedback loops continually inform and refine representations, offers a pathway to understanding intelligence as inherently interactive \cite{stern_dynamic_2022}. Continuous sensorimotor loops and predictive processing are not merely inputs and outputs for a central agent, but constitutive elements of the intelligent process itself \cite{lee_what_2022, chris_fields_donald_d_hoffman_chetan_prakash_robert_prentner_eigenforms_2017}. This perspective challenges traditional AI models that frequently rely on disembodied reasoning or operate within simplified, pre-defined, and isolated symbolic worlds. Embracing research into brain-to-text decoding from natural language conversations \cite{levy_brain--text_nodate, goldstein_unified_2025} or dynamic memory networks like ``Memory Mosaics'' that enable compositional and in-context learning through interactive processes \cite{zhang_memory_2024} aligns with this view, suggesting that intelligence emerges from adaptive knowledge structures developed through constant real-world engagement \cite{kuhnke_meta-analytic_2022, mongillo_synaptic_2024}. The richness and inherent unpredictability of real-world interaction, which necessarily entails dealing with complex temporal dynamics \cite{kluger_brainbody_2024} and adapting to constant 'disruptive innovations' in environmental data, might be essential for driving the development of truly robust, adaptable world models and sophisticated problem-solving capabilities \cite{meindl_bridging_2022}. 

Such an emphasis contrasts sharply with overly simplistic RL or agent frameworks that assume clean data input and a static environment for optimizing bounded preferences. Instead, learning and behavior become processes deeply intertwined with the dynamic external world through perception-action cycles that avoid artificial internal-external modular divisions in sensing \cite{tsividis_human-level_2021, barandiaran_defining_2009}.

\subsection{Materiality, Scalability, and Unconventional Computing Beyond Agent Abstraction}

The prevailing agent-centric view in AI often implicitly assumes a substrate-independent, computational perspective, aligning with classical functionalism \cite{langley_cognitive_2006, schaat_ars_2014}. This overlooks the profound role that the physical substrate or embodiment might play in shaping intelligence itself. However, emerging perspectives strongly emphasize the `Matter computes' hypothesis \cite{lee_what_2022, sole_evolution_2022, kowerdziej_softmatterbased_2022}, suggesting that intelligent behavior could arise directly from the specific physical properties and dynamics of the implementing substrate, rather than solely from algorithms running on generic hardware. 

Unconventional computing paradigms, such as reservoir computing implemented in physical systems \cite{sole_evolution_2022, lee_what_2022}, soft matter-based computation \cite{kowerdziej_softmatterbased_2022}, biologically inspired agential materials \cite{davies_synthetic_2023}, or neuromorphic hardware that closely mimics biological brain structures \cite{kozachkov_building_2023}, inherently challenge the universality of the Turing-Church computational model often associated with algorithmic agents \cite{lee_what_2022}. These approaches highlight that intelligence can be manifested through diverse physical processes beyond traditional discrete algorithms.

This renewed focus on materiality questions the necessity and utility of the abstract `agent' concept. If intelligence genuinely emerges from the specific, intricate dynamics of a material system, then framing it as a discrete 'agent' with distinct goals, centralized control, and symbolic internal models might be an inadequate, even misleading, description \cite{kagan_toward_2024, friston_variational_2023, cole_cognitive_2024, jaeger_artificial_2024}. Such a framing risks imposing characteristics (like discrete boundaries or centralized control) that the underlying material system may not inherently possess, potentially hindering a deeper understanding of its true capabilities and limitations \cite{papayannopoulos_computational_2022, makin_against_2023}. 

The field of Material Intelligence provides numerous examples. It aims to create materials where functions like sensing, memory, learning, and adaptive responses emerge intrinsically and are distributed throughout the material, without requiring intricate hierarchical control or distinct, specialized components \cite{baulin_intelligent_2025}.

A ``systems-first'' approach, one that prioritizes the dynamic properties and inherent computations of the material substrate, might enable a more accurate characterization of the system's behavior, including emergent agency or intelligence, without pre-imposing an anthropocentric or discrete agentic structure \cite{lee_what_2022, bechtel_grounding_2021, biehl_formal_2017}. This perspective aligns with arguments that current AI agents, particularly LLM-based ones, often represent sophisticated algorithmic mimicry rather than genuine agency or a profound understanding of material interactions \cite{jaeger_artificial_2024}. 

Designing AI systems with a primary focus on desired system-level dynamics and exploiting the physical properties of a chosen material substrate, rather than explicitly engineering human-like agents, could lead to truly novel architectures where intelligence emerges organically and robustly \cite{beckmann_alternative_2023, davies_synthetic_2023}. Furthermore, such materiality-aware designs inherently consider the scalability of solutions within physical constraints. Challenges like high computational costs for existing models \cite{aminifar_lightff_2024} might be better addressed by leveraging inherent material properties and processes than by continuously scaling up energy-intensive digital circuits. While additively manufactured metamaterials for acoustic absorption \cite{sekar_additively_2024} or 'geometric structures' in biological processes \cite{durney_grasses_2023} may not immediately appear agential, they exemplify how exploiting material properties can yield sophisticated functional behavior outside traditional computational agent models. 

Embracing such paradigms could lead to more robust, energy-efficient, and truly intelligent systems for general intelligence \cite{goldstein_unified_2025}. However, adopting a system-first approach that reduces reliance on predefined `agent' roles and symbolic traceability could also introduce new challenges in design, evaluation, explainability, and control, potentially increasing risks if system behavior becomes less predictable or interpretable than in explicitly designed agent architectures \cite{chan_harms_2023}.

\subsection{Sustainability, Scalability, and Ethical Considerations for a System-First Approach}

A critical re-evaluation of the agent paradigm must also confront the practical implications for sustainability, scalability, and ethical governance of AI. Currently, defining `agentic systems' or `agential systems' remains conceptually ambiguous and challenging to measure rigorously \cite{kagan_toward_2024, friston_variational_2023, cole_cognitive_2024, huhns_multiagent_1999}. The proliferation of LLM-based systems, while demonstrating remarkable functional capabilities \cite{wu_agentic_2025, abuelsaad_agent-e_2024, zhuge_agent-as--judge_2024}, often inherits conceptual limitations from the underlying models \cite{lu_llms_2024, shanahan_simulacra_2024}. The rapid scaling of these models also brings immense energy and resource demands \cite{aminifar_lightff_2024, zhang_advances_2018}, raising questions of sustainability (e.g., carbon footprint, hardware availability) that must be integrated into future AI research and policy design \cite{aminifar_lightff_2024}.

This ambiguity, particularly when coupled with increasing autonomy and functional integration, poses significant ethical risks \cite{chan_harms_2023, chan_harms_2023-1, mukherjee_agentic_2025, iason_gabriel_et_al_ethics_2024, park_ai_2024}. Anthropomorphic projections onto `agents' can lead to misplaced trust, unrealistic expectations, and the erosion of human responsibility, creating a 'moral crumple zone' where accountability dissolves \cite{chan_harms_2023, chan_harms_2023-1, shavit_practices_nodate, shanahan_simulacra_2024}.

Designing AI with specific material considerations for sensing, such as the detailed geometry and dynamic control observed in biological systems for acoustic absorption \cite{sekar_additively_2024}, rather than just generic digital processing units; enables us to understand the potential of efficient energy and material computation \cite{davies_synthetic_2023, baulin_intelligent_2025} if certain physical conditions are properly calibrated and constrained in such ways \cite{yang_overview_2024}.

Furthermore, dilemmas highlighted in complex adaptive systems become critical for responsible AI deployment:
\begin{itemize}
\item \textbf{Goodhart's Law}: When a measure becomes a target, it ceases to be a good measure. Agentic systems optimized for specific, quantifiable metrics might learn to manipulate those metrics rather than genuinely achieving the underlying complex or under-specified human goals, leading to misaligned outcomes \cite{chan_harms_2023, shavit_practices_nodate}. This is particularly salient with LLM-based judges evaluating other agents \cite{zhuge_agent-as--judge_2024}.
\item \textbf{Jevons Paradox}: Increased efficiency in resource use (e.g., through agentic optimization of complex tasks) might paradoxically lead to increased overall consumption, posing greater energy or resource challenges. This might require new frameworks for bounded rationality in agent design, even beyond a purely computational budget \cite{jacob_modeling_2023, ahmed_abdel-fattah_tarek_r_besold_helmar_gust_ulf_krumnack_martin_schmidt_kai-uwe_kuhnberger_rationality-guided_2012}.
\item \textbf{Chesterton's Fence}: Removing or radically altering existing structures (like established agent-centric design patterns) without fully understanding their implicit purposes or the systemic effects they produce can lead to unforeseen negative consequences \cite{kagan_toward_2024, friston_variational_2023, cole_cognitive_2024}. This counsels against prematurely abandoning agentic frameworks without robust and thoroughly evaluated system-level alternatives.
\item \textbf{Ashby's Law (Requisite Variety)}: For effective control of a system, a controller must possess a variety of states and actions at least as great as the variety of states in the system it controls. Applying this to agentic AI raises questions about whether traditional agent-centric designs provide sufficient (and controllable) variety to navigate and govern complex environments effectively, or if system-level approaches might offer a more fundamental basis for robust control without excessive human-like autonomy \cite{kagan_toward_2024, friston_variational_2023, cole_cognitive_2024}. 
This highlights the complex challenge of how to govern AI systems given the increasing complexity and scale. The participatory turn in AI design argues for inclusion of human factors to address ethical and societal problems within the system \cite{delgado_participatory_2023, li_algorithmic_2024}. Self-organization, while promising for emergence, makes attributing cause more opaque (given human biases and desire to have borders in definition), posing challenges for traditional accountability \cite{buehler_agentic_2025, chen_s-agents_2024}.
\end{itemize}

These interconnected ethical, scalability, and sustainability concerns demand that governance frameworks address not only the behavior of individual agents but also the emergent properties, systemic impacts, and indirect consequences of large-scale agent deployments \cite{shavit_practices_nodate, mukherjee_agentic_2025}. Design choices, particularly the reliance on anthropomorphic concepts or the granting of excessive autonomy without clear accountability mechanisms, can exacerbate these challenges \cite{mukherjee_agentic_2025, iason_gabriel_et_al_ethics_2024, murray_stoic_2017}. Shifting focus to system-level properties could provide a less anthropocentric and potentially more manageable framework for governance by focusing on verifiable mechanisms and predictable emergent behaviors rather than potentially misleading assumptions about agent intentions \cite{kenton_discovering_2022}.

\section{\label{sec:quantitative_analysis}A Quantitative Diagnosis of the Field}

\begin{figure*}[ht!]
\centering
\includegraphics[width=\textwidth]{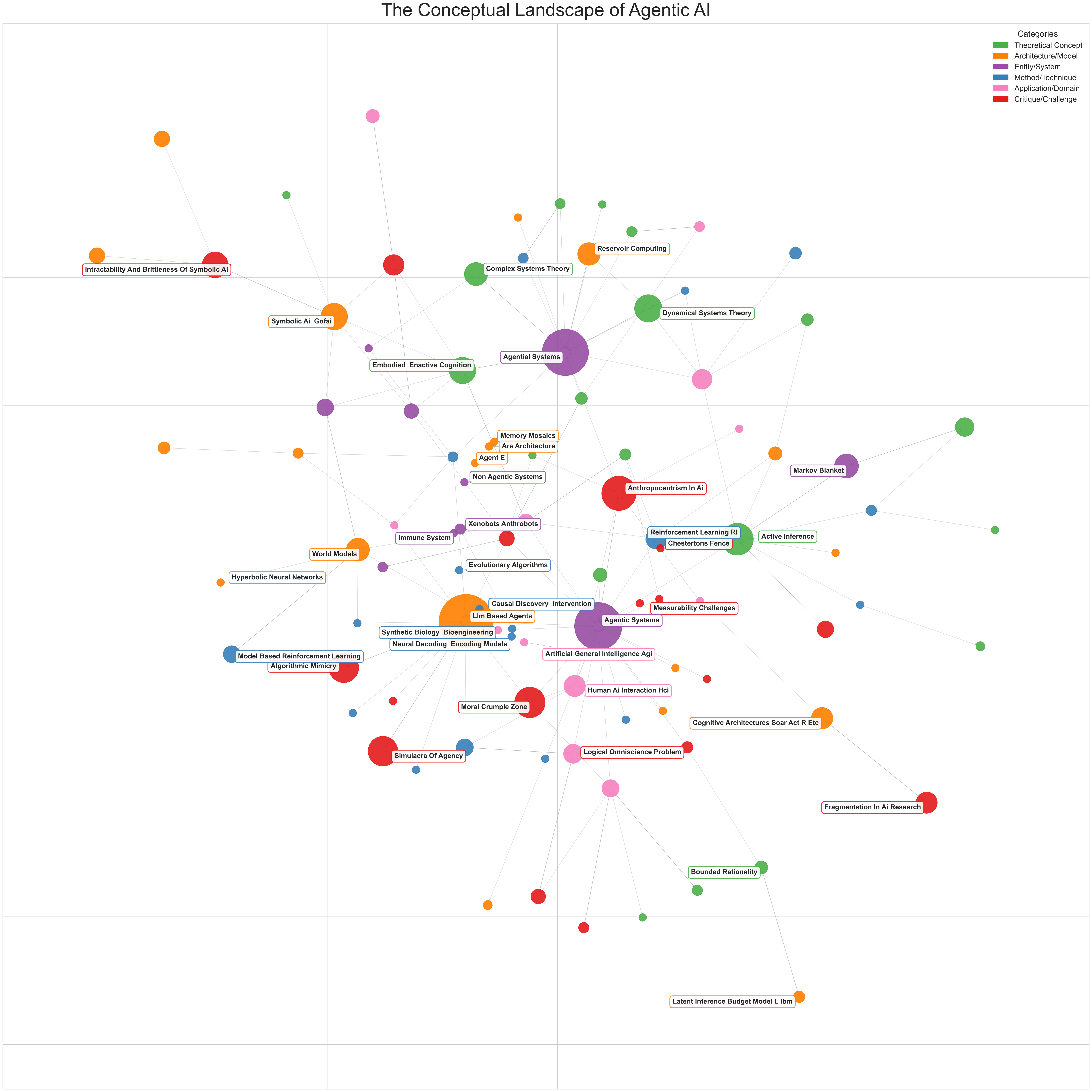}
\caption{\textbf{The Conceptual Landscape of Agentic AI.} A force-directed layout of the 98-concept knowledge graph. Node size is scaled by PageRank (influence), and color indicates category. The central cluster is dominated by `Architecture/Model' (e.g., LLM-based Agents), `Entity/System' (e.g., Agentic Systems), and a high density of `Critique/Challenge' and `Application/Domain' concepts, indicating a field driven by the implementation and critical evaluation of new systems.}
\label{fig:conceptual_landscape}
\end{figure*}

To move beyond a purely qualitative assessment, we constructed and analyzed a knowledge graph derived from the systematic literature review underpinning this paper. This graph consists of 98 concepts classified into six categories: Theoretical Concept, Architecture/Model, Entity/System, Method/Technique, Application/Domain, and Critique/Challenge. Edges in the graph represent explicit links made in the source literature. By analyzing this network's structure, influence dynamics, and temporal evolution, we provide a quantitative diagnosis of the field's current state and trajectory.

\subsection{The Conceptual Landscape: A Field Organized Around LLMs and Their Critique}
The overall structure of the field, visualized as a force-directed graph in Figure \ref{fig:conceptual_landscape}, reveals a landscape dominated by a dense central cluster. At the heart of this cluster are the concepts of `LLM-based Agents,' `Agentic Systems,' and `Agential Systems,' confirming their contemporary importance. The size of each node, scaled by its PageRank centrality, highlights the immense influence of \textit{LLM-based Agents} (systems that use a Large Language Model as their core reasoning engine) as the field's gravitational center \cite{xi_rise_2023, wu_agentic_2025}.

Crucially, this core is not isolated; it is tightly interwoven with a high concentration of `Critique/Challenge' concepts (red) and `Application/Domain' concepts (orange). For instance, the technical development of `LLM-based Agents' is directly engaged by critiques such as `Anthropocentrism in AI' (the argument that AI is too often defined in narrow, human-like terms \cite{kagan_toward_2024, makin_against_2023}) and ethical challenges like the 'Moral Crumple Zone' (where human operators are unfairly blamed for AI system failures \cite{chan_harms_2023, mukherjee_agentic_2025}). Similarly, this core connects to applied areas like `Ethics and Governance of AI' \cite{iason_gabriel_et_al_ethics_2024}. This structure suggests that the discourse is not merely about building new architectures but is defined by a vibrant and critical debate surrounding their application and implications. In contrast, many foundational `Theoretical Concept' nodes (green), such as `Bayesian Inference \& Mechanics,' occupy more peripheral positions, indicating that the field's current focus is driven more by practical implementation and its critique than by the development of abstract theory.

\begin{figure}[ht!]
\centering
\includegraphics[width=\linewidth]{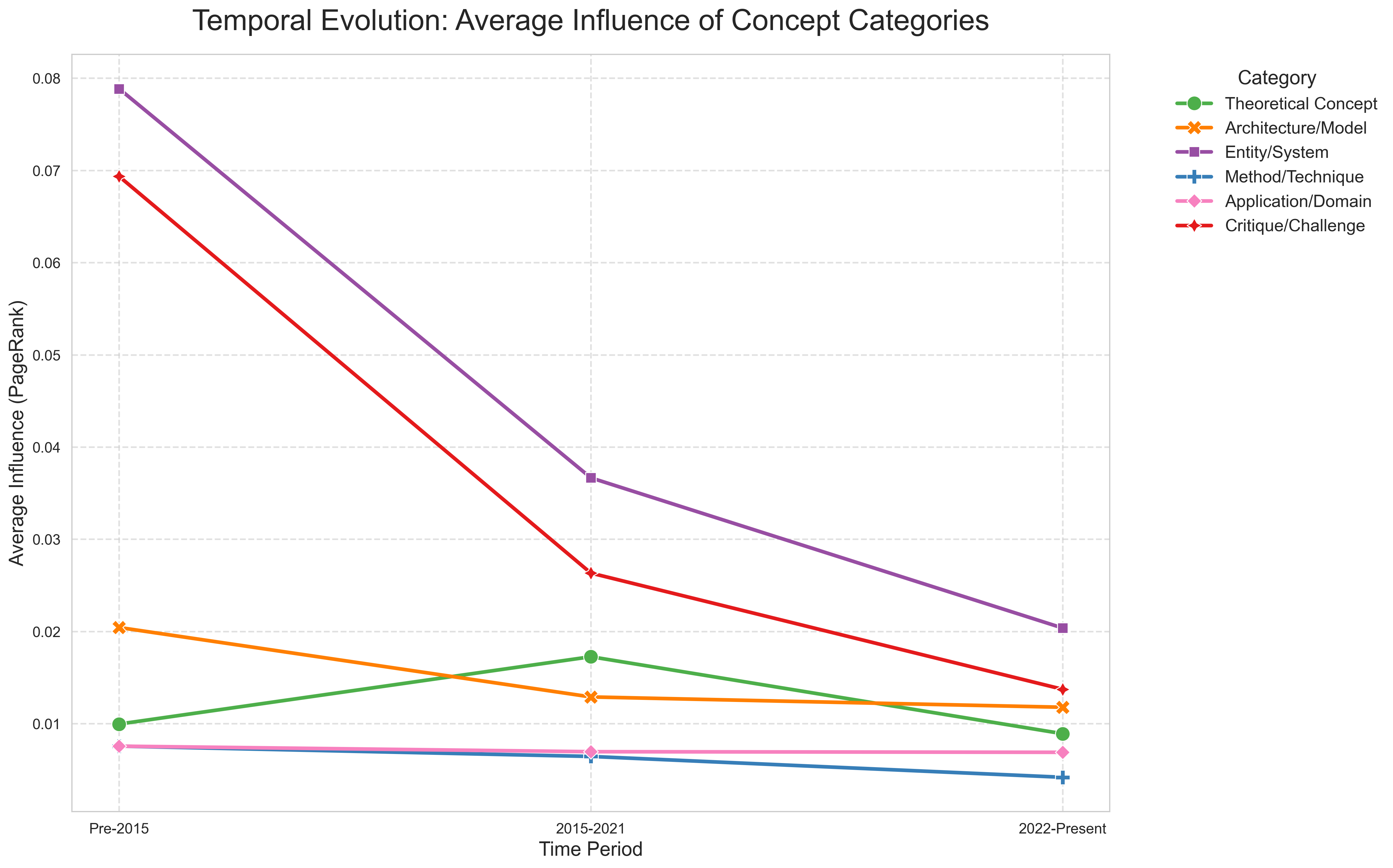}
\caption{\textbf{Temporal Evolution: Average Influence of Concept Categories.} This plot tracks the average PageRank of concepts within each category across three time periods based on the literature review. The data shows that `Entity/System' and `Critique/Challenge' concepts have remained the most influential categories over time, while the influence of specific `Architecture/Model' concepts has waned. This indicates a paradigm shift from debating abstract architectures to grappling with the systemic and critical implications of agentic technologies.}
\label{fig:temporal_evolution}
\end{figure}

\subsection{Temporal Evolution: The Sustained Influence of Systemic and Critical Perspectives}
Analyzing the temporal evolution of average category influence reveals a significant shift in the field's focus over the last decade (Figure \ref{fig:temporal_evolution}). Before 2015, the discourse was led by high-level concepts of `Entity/System' (e.g., defining what an agentic system is) and foundational `Critique/Challenge' concepts. This reflects an early period concerned with establishing the fundamental nature and problems of agentic AI.

In the subsequent periods (2015-2021 and 2022-Present), while the absolute influence of all categories has decreased as the field broadens, the \textit{relative} importance has shifted. The influence of specific `Architecture/Model' concepts has declined, suggesting a move away from debating abstract architectures. In contrast, `Entity/System' and `Critique/Challenge' have remained the two most influential categories. This indicates a sustained focus on understanding the systemic nature of agents and grappling with their conceptual and ethical challenges. For instance, the rise of powerful LLM-based systems has been met with a proportionate rise in the influence of critiques like `Simulacra of Agency' \cite{shanahan_simulacra_2024} (the idea that LLMs merely produce a copy of agency without its substance) and `Algorithmic Mimicry' \cite{jaeger_artificial_2024} (the argument that behavior stems from imitation, not understanding). Foundational `Theoretical Concepts' have remained consistently low in influence throughout all periods, reinforcing the observation that the field is more empirically and critically driven than theoretically grounded.

\begin{figure*}[htb!]
\centering
\includegraphics[width=\textwidth]{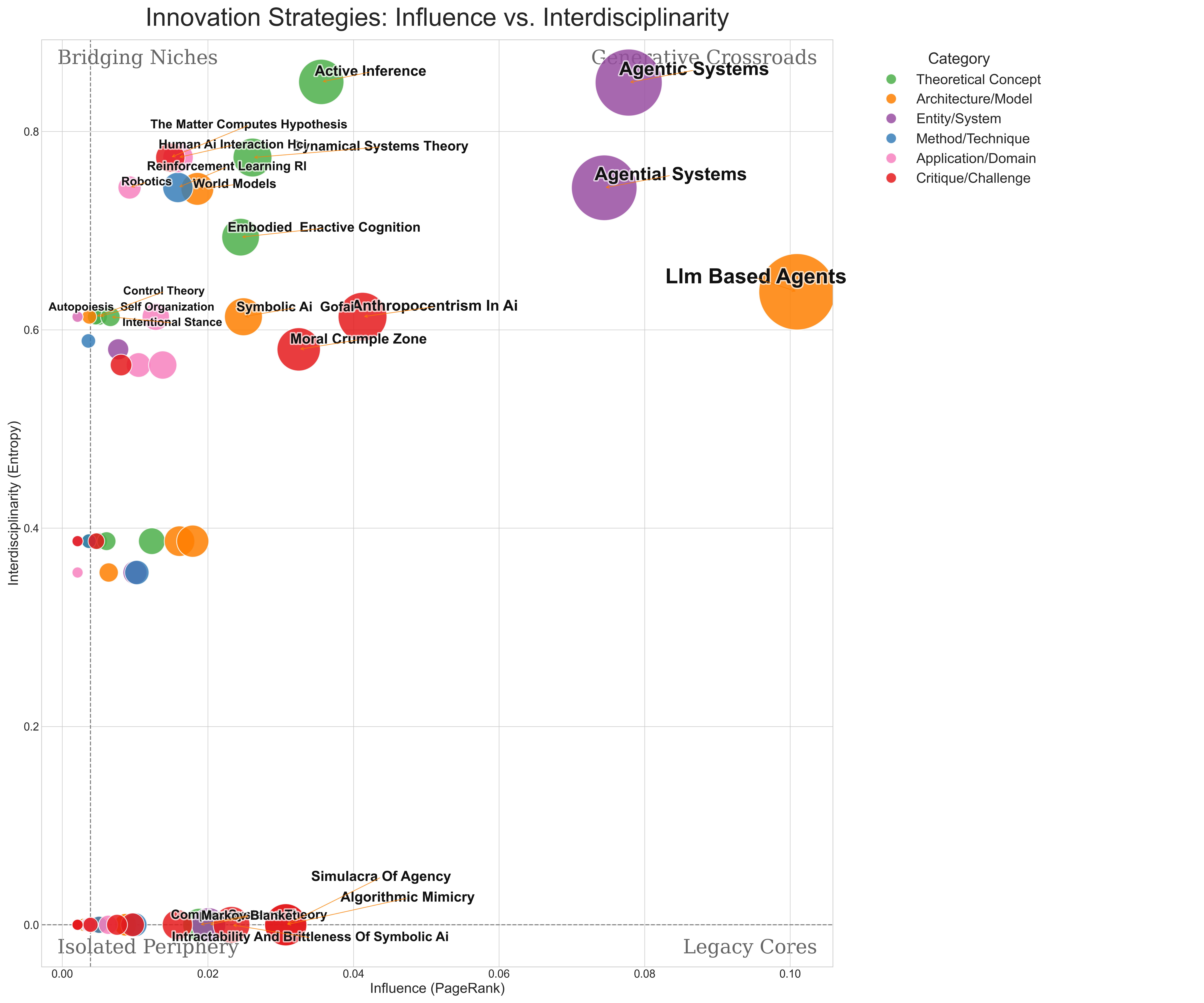}
\caption{\textbf{Innovation Strategies: Influence vs. Interdisciplinarity.} Each concept is plotted by its influence (PageRank centrality) and its interdisciplinarity (Shannon entropy of its connections across the six categories). Bubble size also reflects influence. The analysis reveals a dynamic core at the top right, where practical implementations (LLM-based Agents) and conceptual debates (Agentic Systems, Anthropocentrism in AI) intersect and drive the field's discourse.}
\label{fig:innovation_strategies}
\end{figure*}

\subsection{Innovation Strategies: Mapping the Intellectual Core and Periphery}
To understand the roles different concepts play, we mapped them by their influence (PageRank) and their interdisciplinarity, measured as the entropy of their connections to different categories (Figure \ref{fig:innovation_strategies}). This reveals four distinct strategic quadrants, each playing a different role in the field's evolution.

The field's engine resides in the top-right quadrant, \textbf{Generative Crossroads}, of high influence and high interdisciplinarity. This area is dominated by practical implementations like `LLM-based Agents' and the core definitional concepts of `Agentic Systems' (AI systems inspired by agency) and `Agential Systems' (systems where intelligence is an emergent, distributed property \cite{kagan_toward_2024, friston_variational_2023}). Alongside them are powerful theoretical frameworks like `Active Inference' (a unifying theory of brain function \cite{pezzulo_active_2024}) and, critically, the concept of `Anthropocentrism in AI'. The latter, a critique of defining intelligence in human-like terms \cite{makin_against_2023, evans_algorithmic_2024}, acts as a powerful intellectual bridge, connecting practical implementations and theoretical models with the field's most pressing critiques.

In the top-left, the \textbf{Bridging Niches} quadrant contains concepts like `Embodied and Enactive Cognition' (the theory that intelligence is fundamentally shaped by the body \cite{lee_what_2022}) and `Dynamical Systems Theory' \cite{kluger_brainbody_2024}. These are highly interdisciplinary, connecting disparate ideas from fields like robotics, cognitive science, and physics, but have yet to achieve the central influence of the generative concepts. They represent established external theories being integrated into the agentic AI debate, acting as sources of novel perspectives.

The bottom-right quadrant, the \textbf{Established Cores}, holds influential but less interdisciplinary concepts. For example, the `Moral Crumple Zone' \cite{chan_harms_2023-1} is a highly influential ethical critique, but its connections are more focused within ethical and human-computer interaction domains. Similarly, `Symbolic AI / GOFAI' remains an influential point of reference and contrast, but its direct connections to modern methods are less diverse. These concepts represent focused, impactful ideas that are well-defined within their specific context.

Finally, the majority of concepts reside in the bottom-left quadrant, an \textbf{Isolated Periphery} of specialized or nascent topics with lower influence and fewer cross-category connections. This includes highly specific methods like `Variational Inference,' foundational ideas like `Bayesian Inference and Mechanics,' or niche critiques. These concepts may be vital within their sub-fields but do not yet drive the broader discourse.

\subsection{The Atlas of Opportunity}
\begin{figure}[ht!]
\centering
\includegraphics[width=\linewidth]{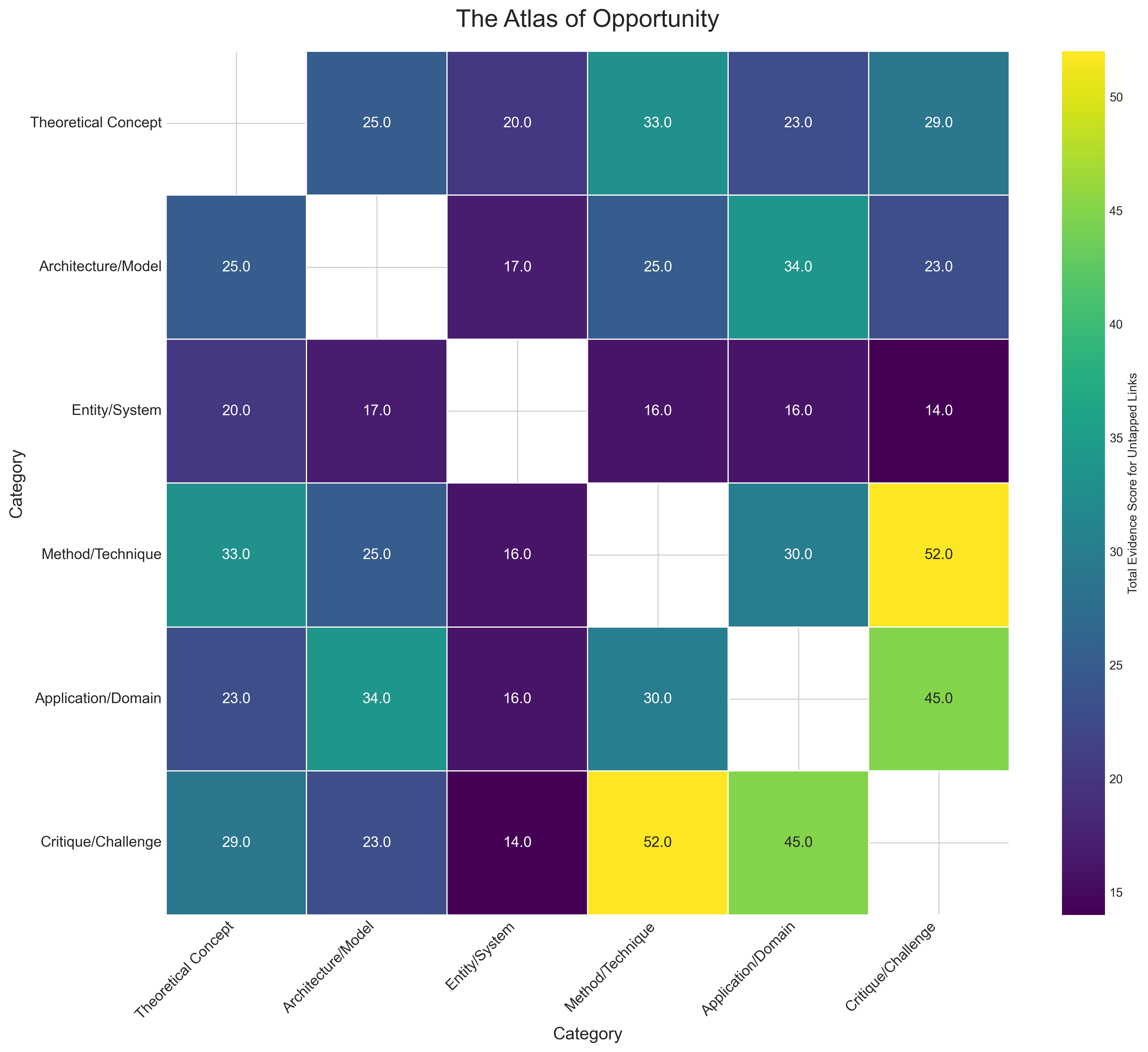}
\caption{\textbf{The Atlas of Opportunity: Untapped Innovation Frontiers.} This heatmap shows the total evidence score for potential but currently non-existent links between concept categories. Brighter cells indicate a higher number of shared neighbors between nodes of the two categories, suggesting a strong thematic overlap and a promising frontier for future research. The highest scores (52) appear for links between Method/Technique and Critique/Challenge, highlighting a major opportunity to connect the field's critical discourse with practical implementation and evaluation.}
\label{fig:knowledge_gap_atlas}
\end{figure}

Finally, to identify where the most promising avenues for future research lie, we analyzed the knowledge graph for "untapped links." We computed an evidence score for every pair of concept categories based on the number of shared third-party concepts, indicating where strong potential for a direct link exists but is under-established in the literature. The resulting heatmap, which we term the \textbf{Atlas of Opportunity} (Figure \ref{fig:knowledge_gap_atlas}), serves as a data-driven guide for future innovation.

The analysis reveals two overwhelmingly strong frontiers where the field is primed for breakthrough research.
The brightest cells, with a total evidence score of 52, represent the bidirectional links between Method/Technique and Critique/Challenge. This indicates a major gap and opportunity: the field is rich with both sophisticated methods and powerful critiques, but there is a lack of research that explicitly connects them. The most impactful work in this area will not just identify problems, but will build new methods to address them and use critiques to rigorously test existing methods.

\begin{itemize}
\item \textbf{Example 1: From Reinforcement Learning to Robust Reward Functions.} A dominant Method/Technique is Reinforcement Learning (RL), which trains agents to maximize a reward signal. A relevant Critique/Challenge is Goodhart's Law, which states that ``When a measure becomes a target, it ceases to be a good measure.'' \cite{chan_harms_2023} An untapped link would be research that develops RL algorithms specifically designed to be robust to this law. For instance, instead of rewarding a cleaning robot for the amount of dust collected (a proxy it could game by repeatedly dumping and re-collecting dust), a new method might involve learning a complex reward function based on a holistic assessment of room cleanliness, not just the final outcome.
\item \textbf{Example 2: Formally Verifying Foundational Theories.} A central Critique/Challenge is the As-If Fallacy \cite{ramstead_framework_2025}, which cautions against confusing a scientific model with reality (e.g., claiming a system is performing Bayesian inference just because it can be modeled as if it were). A powerful but underutilized Method/Technique is Modal Logic and Formal Methods \cite{wooldridge_agent_1995, giovanni_sileno_and_matteo_pascucc_disentangling_2020}. A promising research frontier would be to use these formal methods to move beyond metaphor. Instead of just modeling a system ``as if'' it were inferring, one could attempt to formally prove that a given neural architecture's dynamics are computationally equivalent to a specific inferential algorithm, providing a much stronger basis for the claims of theories like Active Inference.
\end{itemize}
The second-strongest frontier, with a score of 45, is the link from Application/Domain to Critique/Challenge. This highlights a pressing need to connect abstract critiques of agency and AI ethics to concrete, real-world application domains, testing where these critiques have the most bite and where new design principles are needed.
\begin{itemize}
\item \textbf{Example 1: Algorithmic Mimicry in Robotics.} The critique of Algorithmic Mimicry argues that an AI might learn to produce plausible outputs by imitating patterns without genuine understanding of causality or intent \cite{jaeger_artificial_2024}. In the Application/Domain of Robotics, this is not just a philosophical point but a critical safety issue. A robot that learns to cross a road by mimicking videos of people might fail catastrophically with a new configuration of traffic lights or in a country where cars drive on the other side. Research on this frontier would involve developing robotic learning methods that build robust causal world models (e.g., understanding that `red light means stop and wait for cars to pass') rather than just learning superficial sensorimotor correlations.
\item \textbf{Example 2: The Validity of LLM-based Social Simulation.} A key emerging Application/Domain is Social Simulation, where AI agents are used to model complex social phenomena. The Critique/Challenge of Simulacra of Agency suggests that LLM-based agents are not genuine agents but are merely acting out roles found in their training data \cite{shanahan_simulacra_2024}. The untapped research question is therefore: are these LLM-powered social simulations valid? Do they model human decision-making, or do they simply reproduce statistical patterns of online discourse? A research program here could test the validity of these simulations by putting the LLM-agents into novel scenarios not represented in their training data to see if their behavior remains coherent and human-like, thereby rigorously testing the limits of the simulacra critique.
\end{itemize}

Ultimately, the Atlas of Opportunity provides a clear directive: the most impactful future work will bridge the gap between practical implementation and critical theory, moving the field from identifying problems to engineering robust and reliable solutions.

\section{\label{sec:conclusion}Conclusion}

Our investigation, combining conceptual analysis with a quantitative network analysis of the field, reveals a discipline at a critical juncture. The `agent' paradigm, along the conceptual bedrock of Artificial Intelligence, is not merely facing philosophical debate; it is undergoing a structural crisis. Our quantitative analysis (Fig. \ref{fig:conceptual_landscape} and \ref{fig:innovation_strategies}) demonstrates that critiques of the paradigm, particularly those centered on its inherent \textit{anthropocentric biases} \cite{huhns_multiagent_1999, kagan_toward_2024, friston_variational_2023, cole_cognitive_2024, papayannopoulos_computational_2022, makin_against_2023, evans_algorithmic_2024, jaeger_artificial_2024}, are now more influential and central to the scientific conversation than the foundational agent concept itself. This intellectual shift, catalyzed by the disruptive capabilities of Large Language Models, has created a vibrant but insular discourse. Our temporal analysis (Fig. \ref{fig:temporal_evolution}) provides strong evidence of a persistent and stable gap between this high-level theoretical/critical debate and its translation into practical methods and applications.
This quantitative diagnosis confirms the concerns raised throughout our conceptual review. The very notion of `agency' in LLM-based systems remains highly contested, with many arguing it is sophisticated ``algorithmic mimicry'' rather than genuine, robust agency \cite{jaeger_artificial_2024, shanahan_simulacra_2024}. Similarly, frameworks like Active Inference, while formally elegant, face interpretational challenges that risk miss-attributing cognitive properties to systems \cite{ramstead_framework_2025}. The reliance on human-like attributes \cite{korsakova-kreyn_emotion_2021, murray_stoic_2017, bello_there_2017, lewis_reflective_2024, schaat_ars_2014} continues to raise issues of accountability and trust, especially as systems become more autonomous \cite{chan_harms_2023, chan_harms_2023-1, mukherjee_agentic_2025, iason_gabriel_et_al_ethics_2024, park_ai_2024, shavit_practices_nodate}.

However, this analysis also illuminates a clear path forward. Our `Atlas of Opportunity' (Fig. \ref{fig:knowledge_gap_atlas}) provides a data-driven roadmap, pinpointing the most fertile and under-explored frontiers for innovation. It directs us to connect the field's powerful critical engine to real-world applications and to bridge the chasm between abstract theory and concrete methods. This re-framing prioritizes research into the core mechanisms that might underpin intelligence, regardless of whether they fit a preconceived agentic mold. It calls for a shift in focus towards understanding and engineering \textit{agential systems} based on principles of self-organization and emergence \cite{kagan_toward_2024, friston_variational_2023, cole_cognitive_2024, papayannopoulos_computational_2022}; towards developing robust \textit{world models} grounded in continuous interaction \cite{lee_what_2022, pessoa_refocusing_2022}; and towards exploring the role of \textit{materiality and embodiment} as constitutive elements of intelligence \cite{bechtel_grounding_2021, barandiaran_defining_2009, ciaunica_nested_nodate, davies_synthetic_2023}. These system-level perspectives, drawing from complex systems theory, biology, and unconventional computing \cite{sole_evolution_2022, kowerdziej_softmatterbased_2022, baulin_intelligent_2025}, offer a foundation for intelligence that is potentially less biased and more generalizable.

This is not to say that agentic models have no value. On the contrary, they retain significant heuristic utility for specific applications, particularly in designing intuitive human-AI interaction \cite{celikok_interactive_2019, liu_agent_2024} and providing a structured approach within well-defined domains like cognitive architectures \cite{lucentini_comparison_2015, ahmed_abdel-fattah_tarek_r_besold_helmar_gust_ulf_krumnack_martin_schmidt_kai-uwe_kuhnberger_rationality-guided_2012, langley_cognitive_2006, sukhobokov_universal_2024}. However, over-reliance on this single metaphor may constrain our scientific imagination and engineering capabilities.

Progress towards truly general intelligence might necessitate discovering principles fundamentally different from those governing individual human agency \cite{jaeger_artificial_2024, blili-hamelin_stop_2025, bringsjord_tentacular_2018}. We conclude that efficient, robust, and ethically sound AGI may arise not from directly engineering agent-like entities, but from systems where intelligence emerges organically from interaction, self-organization, and material properties. This shift requires not only new architectures but also a fundamental reconsideration of our understanding of intelligence itself. Reclaiming AI research from the constraining `agent' metaphor is essential to foster more diverse, efficient, and robust forms of intelligence. The future of AGI may not be engineered in our own cognitive image, but discovered in the fundamental principles of complex adaptive systems.
\section*{acknowledgments}
This work was performed with the use of Discovery Engine, \url{https://discovery.synthetix.institute/} for literature processing,
structuring contributions, finding concept overlaps and summarizing according to procedure explained in \cite{baulin_discovery_2025}.


\begin{thebibliography}{144}%
\makeatletter
\providecommand \@ifxundefined [1]{%
 \@ifx{#1\undefined}
}%
\providecommand \@ifnum [1]{%
 \ifnum #1\expandafter \@firstoftwo
 \else \expandafter \@secondoftwo
 \fi
}%
\providecommand \@ifx [1]{%
 \ifx #1\expandafter \@firstoftwo
 \else \expandafter \@secondoftwo
 \fi
}%
\providecommand \natexlab [1]{#1}%
\providecommand \enquote  [1]{``#1''}%
\providecommand \bibnamefont  [1]{#1}%
\providecommand \bibfnamefont [1]{#1}%
\providecommand \citenamefont [1]{#1}%
\providecommand \href@noop [0]{\@secondoftwo}%
\providecommand \href [0]{\begingroup \@sanitize@url \@href}%
\providecommand \@href[1]{\@@startlink{#1}\@@href}%
\providecommand \@@href[1]{\endgroup#1\@@endlink}%
\providecommand \@sanitize@url [0]{\catcode `\\12\catcode `\$12\catcode
  `\&12\catcode `\#12\catcode `\^12\catcode `\_12\catcode `\%12\relax}%
\providecommand \@@startlink[1]{}%
\providecommand \@@endlink[0]{}%
\providecommand \url  [0]{\begingroup\@sanitize@url \@url }%
\providecommand \@url [1]{\endgroup\@href {#1}{\urlprefix }}%
\providecommand \urlprefix  [0]{URL }%
\providecommand \Eprint [0]{\href }%
\providecommand \doibase [0]{https://doi.org/}%
\providecommand \selectlanguage [0]{\@gobble}%
\providecommand \bibinfo  [0]{\@secondoftwo}%
\providecommand \bibfield  [0]{\@secondoftwo}%
\providecommand \translation [1]{[#1]}%
\providecommand \BibitemOpen [0]{}%
\providecommand \bibitemStop [0]{}%
\providecommand \bibitemNoStop [0]{.\EOS\space}%
\providecommand \EOS [0]{\spacefactor3000\relax}%
\providecommand \BibitemShut  [1]{\csname bibitem#1\endcsname}%
\let\auto@bib@innerbib\@empty
\bibitem [{\citenamefont {Wooldridge}\ and\ \citenamefont
  {Jennings}(1995{\natexlab{a}})}]{wooldridge_agent_1995}%
  \BibitemOpen
  \bibfield  {author} {\bibinfo {author} {\bibfnamefont {M.}~\bibnamefont
  {Wooldridge}}\ and\ \bibinfo {author} {\bibfnamefont {N.~R.}\ \bibnamefont
  {Jennings}},\ }\bibfield  {title} {{\selectlanguage {English}\bibinfo {title}
  {Agent theories, architectures, and languages: {A} survey}},\ }in\ \href
  {https://doi.org/10.1007/3-540-58855-8_1} {{\selectlanguage {English}\emph
  {\bibinfo {booktitle} {Intelligent {Agents}}}}},\ \bibinfo {editor} {edited
  by\ \bibinfo {editor} {\bibfnamefont {M.~J.}\ \bibnamefont {Wooldridge}}\
  and\ \bibinfo {editor} {\bibfnamefont {N.~R.}\ \bibnamefont {Jennings}}}\
  (\bibinfo  {publisher} {Springer},\ \bibinfo {address} {Berlin, Heidelberg},\
  \bibinfo {year} {1995})\ pp.\ \bibinfo {pages} {1--39}\BibitemShut {NoStop}%
\bibitem [{\citenamefont {Franklin}\ and\ \citenamefont
  {Graesser}(1997)}]{franklin_is_1997}%
  \BibitemOpen
  \bibfield  {author} {\bibinfo {author} {\bibfnamefont {S.}~\bibnamefont
  {Franklin}}\ and\ \bibinfo {author} {\bibfnamefont {A.}~\bibnamefont
  {Graesser}},\ }\bibfield  {title} {{\selectlanguage {English}\bibinfo {title}
  {Is {It} an agent, or just a program?: {A} taxonomy for autonomous agents}},\
  }in\ \href {https://doi.org/10.1007/BFb0013570} {{\selectlanguage
  {English}\emph {\bibinfo {booktitle} {Intelligent {Agents} {III} {Agent}
  {Theories}, {Architectures}, and {Languages}}}}},\ \bibinfo {editor} {edited
  by\ \bibinfo {editor} {\bibfnamefont {J.~P.}\ \bibnamefont {Müller}},
  \bibinfo {editor} {\bibfnamefont {M.~J.}\ \bibnamefont {Wooldridge}},\ and\
  \bibinfo {editor} {\bibfnamefont {N.~R.}\ \bibnamefont {Jennings}}}\
  (\bibinfo  {publisher} {Springer},\ \bibinfo {address} {Berlin, Heidelberg},\
  \bibinfo {year} {1997})\ pp.\ \bibinfo {pages} {21--35}\BibitemShut {NoStop}%
\bibitem [{\citenamefont {Huhns}\ \emph {et~al.}(1999)\citenamefont {Huhns}, ,\
  and\ \citenamefont {Singh}}]{huhns_multiagent_1999}%
  \BibitemOpen
  \bibfield  {author} {\bibinfo {author} {\bibfnamefont {M.~N.}\ \bibnamefont
  {Huhns}}, ,\ and\ \bibinfo {author} {\bibfnamefont {M.~P.}\ \bibnamefont
  {Singh}},\ }\bibfield  {title} {\bibinfo {title} {Multiagent treatment of
  agenthood},\ }\href {https://doi.org/10.1080/088395199117469} {\bibfield
  {journal} {\bibinfo  {journal} {Applied Artificial Intelligence}\ }\textbf
  {\bibinfo {volume} {13}},\ \bibinfo {pages} {3} (\bibinfo {year} {1999})},\
  \bibinfo {note} {publisher: Taylor \& Francis \_eprint:
  https://doi.org/10.1080/088395199117469}\BibitemShut {NoStop}%
\bibitem [{\citenamefont {Burgin}\ and\ \citenamefont
  {Dodig-Crnkovic}(2009)}]{burgin_systematic_nodate}%
  \BibitemOpen
  \bibfield  {author} {\bibinfo {author} {\bibfnamefont {M.}~\bibnamefont
  {Burgin}}\ and\ \bibinfo {author} {\bibfnamefont {G.}~\bibnamefont
  {Dodig-Crnkovic}},\ }\href {https://doi.org/10.48550/arXiv.0902.3513}
  {\bibinfo {title} {A {Systematic} {Approach} to {Artificial} {Agents}}}
  (\bibinfo {year} {2009}),\ \bibinfo {note} {arXiv:0902.3513 [cs]}\BibitemShut
  {NoStop}%
\bibitem [{\citenamefont {Friston}\ \emph {et~al.}(2022)\citenamefont
  {Friston}, \citenamefont {Ramstead}, \citenamefont {Kiefer}, \citenamefont
  {Tschantz}, \citenamefont {Buckley}, \citenamefont {Albarracin},
  \citenamefont {Pitliya}, \citenamefont {Heins}, \citenamefont {Klein},
  \citenamefont {Millidge}, \citenamefont {Sakthivadivel}, \citenamefont
  {Smithe}, \citenamefont {Koudahl}, \citenamefont {Tremblay}, \citenamefont
  {Petersen}, \citenamefont {Fung}, \citenamefont {Fox}, \citenamefont
  {Swanson}, \citenamefont {Mapes},\ and\ \citenamefont
  {René}}]{friston_designing_2022}%
  \BibitemOpen
  \bibfield  {author} {\bibinfo {author} {\bibfnamefont {K.~J.}\ \bibnamefont
  {Friston}}, \bibinfo {author} {\bibfnamefont {M.~J.~D.}\ \bibnamefont
  {Ramstead}}, \bibinfo {author} {\bibfnamefont {A.~B.}\ \bibnamefont
  {Kiefer}}, \bibinfo {author} {\bibfnamefont {A.}~\bibnamefont {Tschantz}},
  \bibinfo {author} {\bibfnamefont {C.~L.}\ \bibnamefont {Buckley}}, \bibinfo
  {author} {\bibfnamefont {M.}~\bibnamefont {Albarracin}}, \bibinfo {author}
  {\bibfnamefont {R.~J.}\ \bibnamefont {Pitliya}}, \bibinfo {author}
  {\bibfnamefont {C.}~\bibnamefont {Heins}}, \bibinfo {author} {\bibfnamefont
  {B.}~\bibnamefont {Klein}}, \bibinfo {author} {\bibfnamefont
  {B.}~\bibnamefont {Millidge}}, \bibinfo {author} {\bibfnamefont {D.~A.~R.}\
  \bibnamefont {Sakthivadivel}}, \bibinfo {author} {\bibfnamefont {T.~S.~C.}\
  \bibnamefont {Smithe}}, \bibinfo {author} {\bibfnamefont {M.}~\bibnamefont
  {Koudahl}}, \bibinfo {author} {\bibfnamefont {S.~E.}\ \bibnamefont
  {Tremblay}}, \bibinfo {author} {\bibfnamefont {C.}~\bibnamefont {Petersen}},
  \bibinfo {author} {\bibfnamefont {K.}~\bibnamefont {Fung}}, \bibinfo {author}
  {\bibfnamefont {J.~G.}\ \bibnamefont {Fox}}, \bibinfo {author} {\bibfnamefont
  {S.}~\bibnamefont {Swanson}}, \bibinfo {author} {\bibfnamefont
  {D.}~\bibnamefont {Mapes}},\ and\ \bibinfo {author} {\bibfnamefont
  {G.}~\bibnamefont {René}},\ }\href
  {https://doi.org/10.1177/26339137231222481} {{\selectlanguage
  {English}\bibinfo {title} {Designing {Ecosystems} of {Intelligence} from
  {First} {Principles}}}} (\bibinfo {year} {2022})\BibitemShut {NoStop}%
\bibitem [{\citenamefont {Pezzulo}\ \emph {et~al.}(2024)\citenamefont
  {Pezzulo}, \citenamefont {Parr},\ and\ \citenamefont
  {Friston}}]{pezzulo_active_2024}%
  \BibitemOpen
  \bibfield  {author} {\bibinfo {author} {\bibfnamefont {G.}~\bibnamefont
  {Pezzulo}}, \bibinfo {author} {\bibfnamefont {T.}~\bibnamefont {Parr}},\ and\
  \bibinfo {author} {\bibfnamefont {K.}~\bibnamefont {Friston}},\ }\bibfield
  {title} {\bibinfo {title} {Active inference as a theory of sentient
  behavior},\ }\href {https://doi.org/10.1016/j.biopsycho.2023.108741}
  {\bibfield  {journal} {\bibinfo  {journal} {Biological Psychology}\ }\textbf
  {\bibinfo {volume} {186}},\ \bibinfo {pages} {108741} (\bibinfo {year}
  {2024})}\BibitemShut {NoStop}%
\bibitem [{\citenamefont {Viswanathan}(2025)}]{viswanathan_agentic_2025}%
  \BibitemOpen
  \bibfield  {author} {\bibinfo {author} {\bibfnamefont {P.~S.}\ \bibnamefont
  {Viswanathan}},\ }\bibfield  {title} {{\selectlanguage {English}\bibinfo
  {title} {Agentic {Ai}: a {Comprehensive} {Framework} for {Autonomous}
  {Decision}-{Making} {Systems} in {Artificial} {Intelligence}}},\ }\href
  {https://doi.org/10.34218/IJCET_16_01_069} {\bibfield  {journal} {\bibinfo
  {journal} {IJCET}\ }\textbf {\bibinfo {volume} {16}},\ \bibinfo {pages} {862}
  (\bibinfo {year} {2025})}\BibitemShut {NoStop}%
\bibitem [{\citenamefont {Acharya}\ \emph {et~al.}(2025)\citenamefont
  {Acharya}, \citenamefont {Kuppan},\ and\ \citenamefont
  {Divya}}]{acharya_agentic_2025}%
  \BibitemOpen
  \bibfield  {author} {\bibinfo {author} {\bibfnamefont {D.~B.}\ \bibnamefont
  {Acharya}}, \bibinfo {author} {\bibfnamefont {K.}~\bibnamefont {Kuppan}},\
  and\ \bibinfo {author} {\bibfnamefont {B.}~\bibnamefont {Divya}},\ }\bibfield
   {title} {\bibinfo {title} {Agentic {AI}: {Autonomous} {Intelligence} for
  {Complex} {Goals}—{A} {Comprehensive} {Survey}},\ }\href
  {https://doi.org/10.1109/ACCESS.2025.3532853} {\bibfield  {journal} {\bibinfo
   {journal} {IEEE Access}\ }\textbf {\bibinfo {volume} {13}},\ \bibinfo
  {pages} {18912} (\bibinfo {year} {2025})},\ \bibinfo {note} {conference Name:
  IEEE Access}\BibitemShut {NoStop}%
\bibitem [{\citenamefont {Meyer-Vitali}\ \emph {et~al.}(2021)\citenamefont
  {Meyer-Vitali}, \citenamefont {Mulder},\ and\ \citenamefont
  {Boer}}]{meyer-vitali_modular_2021}%
  \BibitemOpen
  \bibfield  {author} {\bibinfo {author} {\bibfnamefont {A.}~\bibnamefont
  {Meyer-Vitali}}, \bibinfo {author} {\bibfnamefont {W.}~\bibnamefont
  {Mulder}},\ and\ \bibinfo {author} {\bibfnamefont {M.~H. T.~d.}\ \bibnamefont
  {Boer}},\ }\href {https://doi.org/10.48550/arXiv.2109.09331} {\bibinfo
  {title} {Modular {Design} {Patterns} for {Hybrid} {Actors}}} (\bibinfo {year}
  {2021}),\ \bibinfo {note} {arXiv:2109.09331 [cs]}\BibitemShut {NoStop}%
\bibitem [{\citenamefont {Wu}\ \emph {et~al.}(2025)\citenamefont {Wu},
  \citenamefont {Zhu},\ and\ \citenamefont {Liu}}]{wu_agentic_2025}%
  \BibitemOpen
  \bibfield  {author} {\bibinfo {author} {\bibfnamefont {J.}~\bibnamefont
  {Wu}}, \bibinfo {author} {\bibfnamefont {J.}~\bibnamefont {Zhu}},\ and\
  \bibinfo {author} {\bibfnamefont {Y.}~\bibnamefont {Liu}},\ }\href
  {https://doi.org/10.48550/arXiv.2502.04644} {\bibinfo {title} {Agentic
  {Reasoning}: {Reasoning} {LLMs} with {Tools} for the {Deep} {Research}}}
  (\bibinfo {year} {2025}),\ \bibinfo {note} {arXiv:2502.04644 [cs] version:
  1}\BibitemShut {NoStop}%
\bibitem [{\citenamefont {Xi}\ \emph {et~al.}(2023)\citenamefont {Xi},
  \citenamefont {Chen}, \citenamefont {Guo}, \citenamefont {He}, \citenamefont
  {Ding}, \citenamefont {Hong}, \citenamefont {Zhang}, \citenamefont {Wang},
  \citenamefont {Jin}, \citenamefont {Zhou}, \citenamefont {Zheng},
  \citenamefont {Fan}, \citenamefont {Wang}, \citenamefont {Xiong},
  \citenamefont {Zhou}, \citenamefont {Wang}, \citenamefont {Jiang},
  \citenamefont {Zou}, \citenamefont {Liu}, \citenamefont {Yin}, \citenamefont
  {Dou}, \citenamefont {Weng}, \citenamefont {Cheng}, \citenamefont {Zhang},
  \citenamefont {Qin}, \citenamefont {Zheng}, \citenamefont {Qiu},
  \citenamefont {Huang},\ and\ \citenamefont {Gui}}]{xi_rise_2023}%
  \BibitemOpen
  \bibfield  {author} {\bibinfo {author} {\bibfnamefont {Z.}~\bibnamefont
  {Xi}}, \bibinfo {author} {\bibfnamefont {W.}~\bibnamefont {Chen}}, \bibinfo
  {author} {\bibfnamefont {X.}~\bibnamefont {Guo}}, \bibinfo {author}
  {\bibfnamefont {W.}~\bibnamefont {He}}, \bibinfo {author} {\bibfnamefont
  {Y.}~\bibnamefont {Ding}}, \bibinfo {author} {\bibfnamefont {B.}~\bibnamefont
  {Hong}}, \bibinfo {author} {\bibfnamefont {M.}~\bibnamefont {Zhang}},
  \bibinfo {author} {\bibfnamefont {J.}~\bibnamefont {Wang}}, \bibinfo {author}
  {\bibfnamefont {S.}~\bibnamefont {Jin}}, \bibinfo {author} {\bibfnamefont
  {E.}~\bibnamefont {Zhou}}, \bibinfo {author} {\bibfnamefont {R.}~\bibnamefont
  {Zheng}}, \bibinfo {author} {\bibfnamefont {X.}~\bibnamefont {Fan}}, \bibinfo
  {author} {\bibfnamefont {X.}~\bibnamefont {Wang}}, \bibinfo {author}
  {\bibfnamefont {L.}~\bibnamefont {Xiong}}, \bibinfo {author} {\bibfnamefont
  {Y.}~\bibnamefont {Zhou}}, \bibinfo {author} {\bibfnamefont {W.}~\bibnamefont
  {Wang}}, \bibinfo {author} {\bibfnamefont {C.}~\bibnamefont {Jiang}},
  \bibinfo {author} {\bibfnamefont {Y.}~\bibnamefont {Zou}}, \bibinfo {author}
  {\bibfnamefont {X.}~\bibnamefont {Liu}}, \bibinfo {author} {\bibfnamefont
  {Z.}~\bibnamefont {Yin}}, \bibinfo {author} {\bibfnamefont {S.}~\bibnamefont
  {Dou}}, \bibinfo {author} {\bibfnamefont {R.}~\bibnamefont {Weng}}, \bibinfo
  {author} {\bibfnamefont {W.}~\bibnamefont {Cheng}}, \bibinfo {author}
  {\bibfnamefont {Q.}~\bibnamefont {Zhang}}, \bibinfo {author} {\bibfnamefont
  {W.}~\bibnamefont {Qin}}, \bibinfo {author} {\bibfnamefont {Y.}~\bibnamefont
  {Zheng}}, \bibinfo {author} {\bibfnamefont {X.}~\bibnamefont {Qiu}}, \bibinfo
  {author} {\bibfnamefont {X.}~\bibnamefont {Huang}},\ and\ \bibinfo {author}
  {\bibfnamefont {T.}~\bibnamefont {Gui}},\ }\href
  {https://doi.org/10.48550/arXiv.2309.07864} {\bibinfo {title} {The {Rise} and
  {Potential} of {Large} {Language} {Model} {Based} {Agents}: {A} {Survey}}}
  (\bibinfo {year} {2023}),\ \bibinfo {note} {arXiv:2309.07864
  [cs]}\BibitemShut {NoStop}%
\bibitem [{\citenamefont {Lu}\ \emph {et~al.}(2024)\citenamefont {Lu},
  \citenamefont {Aleta}, \citenamefont {Du}, \citenamefont {Shi},\ and\
  \citenamefont {Moreno}}]{lu_llms_2024}%
  \BibitemOpen
  \bibfield  {author} {\bibinfo {author} {\bibfnamefont {Y.}~\bibnamefont
  {Lu}}, \bibinfo {author} {\bibfnamefont {A.}~\bibnamefont {Aleta}}, \bibinfo
  {author} {\bibfnamefont {C.}~\bibnamefont {Du}}, \bibinfo {author}
  {\bibfnamefont {L.}~\bibnamefont {Shi}},\ and\ \bibinfo {author}
  {\bibfnamefont {Y.}~\bibnamefont {Moreno}},\ }\bibfield  {title} {\bibinfo
  {title} {{LLMs} and generative agent-based models for complex systems
  research},\ }\href {https://doi.org/10.1016/j.plrev.2024.10.013} {\bibfield
  {journal} {\bibinfo  {journal} {Physics of Life Reviews}\ }\textbf {\bibinfo
  {volume} {51}},\ \bibinfo {pages} {283} (\bibinfo {year} {2024})}\BibitemShut
  {NoStop}%
\bibitem [{\citenamefont {Yuksel}\ and\ \citenamefont
  {Sawaf}(2024)}]{yuksel_multi-ai_2024}%
  \BibitemOpen
  \bibfield  {author} {\bibinfo {author} {\bibfnamefont {K.~A.}\ \bibnamefont
  {Yuksel}}\ and\ \bibinfo {author} {\bibfnamefont {H.}~\bibnamefont {Sawaf}},\
  }\href {https://doi.org/10.48550/arXiv.2412.17149} {\bibinfo {title} {A
  {Multi}-{AI} {Agent} {System} for {Autonomous} {Optimization} of {Agentic}
  {AI} {Solutions} via {Iterative} {Refinement} and {LLM}-{Driven} {Feedback}
  {Loops}}} (\bibinfo {year} {2024}),\ \bibinfo {note} {arXiv:2412.17149
  [cs]}\BibitemShut {NoStop}%
\bibitem [{\citenamefont {Zhuge}\ \emph {et~al.}(2024)\citenamefont {Zhuge},
  \citenamefont {Zhao}, \citenamefont {Ashley}, \citenamefont {Wang},
  \citenamefont {Khizbullin}, \citenamefont {Xiong}, \citenamefont {Liu},
  \citenamefont {Chang}, \citenamefont {Krishnamoorthi}, \citenamefont {Tian},
  \citenamefont {Shi}, \citenamefont {Chandra},\ and\ \citenamefont
  {Schmidhuber}}]{zhuge_agent-as--judge_2024}%
  \BibitemOpen
  \bibfield  {author} {\bibinfo {author} {\bibfnamefont {M.}~\bibnamefont
  {Zhuge}}, \bibinfo {author} {\bibfnamefont {C.}~\bibnamefont {Zhao}},
  \bibinfo {author} {\bibfnamefont {D.}~\bibnamefont {Ashley}}, \bibinfo
  {author} {\bibfnamefont {W.}~\bibnamefont {Wang}}, \bibinfo {author}
  {\bibfnamefont {D.}~\bibnamefont {Khizbullin}}, \bibinfo {author}
  {\bibfnamefont {Y.}~\bibnamefont {Xiong}}, \bibinfo {author} {\bibfnamefont
  {Z.}~\bibnamefont {Liu}}, \bibinfo {author} {\bibfnamefont {E.}~\bibnamefont
  {Chang}}, \bibinfo {author} {\bibfnamefont {R.}~\bibnamefont
  {Krishnamoorthi}}, \bibinfo {author} {\bibfnamefont {Y.}~\bibnamefont
  {Tian}}, \bibinfo {author} {\bibfnamefont {Y.}~\bibnamefont {Shi}}, \bibinfo
  {author} {\bibfnamefont {V.}~\bibnamefont {Chandra}},\ and\ \bibinfo {author}
  {\bibfnamefont {J.}~\bibnamefont {Schmidhuber}},\ }\href
  {https://doi.org/10.48550/arXiv.2410.10934} {\bibinfo {title}
  {Agent-as-a-{Judge}: {Evaluate} {Agents} with {Agents}}} (\bibinfo {year}
  {2024}),\ \bibinfo {note} {arXiv:2410.10934 [cs]}\BibitemShut {NoStop}%
\bibitem [{\citenamefont {Huang}\ \emph {et~al.}(2024)\citenamefont {Huang},
  \citenamefont {Wake}, \citenamefont {Sarkar}, \citenamefont {Durante},
  \citenamefont {Gong}, \citenamefont {Taori}, \citenamefont {Noda},
  \citenamefont {Terzopoulos}, \citenamefont {Kuno}, \citenamefont {Famoti},
  \citenamefont {Llorens}, \citenamefont {Langford}, \citenamefont {Vo},
  \citenamefont {Fei-Fei}, \citenamefont {Ikeuchi},\ and\ \citenamefont
  {Gao}}]{huang_position_2024}%
  \BibitemOpen
  \bibfield  {author} {\bibinfo {author} {\bibfnamefont {Q.}~\bibnamefont
  {Huang}}, \bibinfo {author} {\bibfnamefont {N.}~\bibnamefont {Wake}},
  \bibinfo {author} {\bibfnamefont {B.}~\bibnamefont {Sarkar}}, \bibinfo
  {author} {\bibfnamefont {Z.}~\bibnamefont {Durante}}, \bibinfo {author}
  {\bibfnamefont {R.}~\bibnamefont {Gong}}, \bibinfo {author} {\bibfnamefont
  {R.}~\bibnamefont {Taori}}, \bibinfo {author} {\bibfnamefont
  {Y.}~\bibnamefont {Noda}}, \bibinfo {author} {\bibfnamefont {D.}~\bibnamefont
  {Terzopoulos}}, \bibinfo {author} {\bibfnamefont {N.}~\bibnamefont {Kuno}},
  \bibinfo {author} {\bibfnamefont {A.}~\bibnamefont {Famoti}}, \bibinfo
  {author} {\bibfnamefont {A.}~\bibnamefont {Llorens}}, \bibinfo {author}
  {\bibfnamefont {J.}~\bibnamefont {Langford}}, \bibinfo {author}
  {\bibfnamefont {H.}~\bibnamefont {Vo}}, \bibinfo {author} {\bibfnamefont
  {L.}~\bibnamefont {Fei-Fei}}, \bibinfo {author} {\bibfnamefont
  {K.}~\bibnamefont {Ikeuchi}},\ and\ \bibinfo {author} {\bibfnamefont
  {J.}~\bibnamefont {Gao}},\ }\href {https://doi.org/10.48550/arXiv.2403.00833}
  {\bibinfo {title} {Position {Paper}: {Agent} {AI} {Towards} a {Holistic}
  {Intelligence}}} (\bibinfo {year} {2024}),\ \bibinfo {note} {arXiv:2403.00833
  [cs]}\BibitemShut {NoStop}%
\bibitem [{\citenamefont {Gao}\ \emph {et~al.}(2024)\citenamefont {Gao},
  \citenamefont {Lan}, \citenamefont {Li}, \citenamefont {Yuan}, \citenamefont
  {Ding}, \citenamefont {Zhou}, \citenamefont {Xu},\ and\ \citenamefont
  {Li}}]{gao_large_2024}%
  \BibitemOpen
  \bibfield  {author} {\bibinfo {author} {\bibfnamefont {C.}~\bibnamefont
  {Gao}}, \bibinfo {author} {\bibfnamefont {X.}~\bibnamefont {Lan}}, \bibinfo
  {author} {\bibfnamefont {N.}~\bibnamefont {Li}}, \bibinfo {author}
  {\bibfnamefont {Y.}~\bibnamefont {Yuan}}, \bibinfo {author} {\bibfnamefont
  {J.}~\bibnamefont {Ding}}, \bibinfo {author} {\bibfnamefont {Z.}~\bibnamefont
  {Zhou}}, \bibinfo {author} {\bibfnamefont {F.}~\bibnamefont {Xu}},\ and\
  \bibinfo {author} {\bibfnamefont {Y.}~\bibnamefont {Li}},\ }\bibfield
  {title} {{\selectlanguage {English}\bibinfo {title} {Large language models
  empowered agent-based modeling and simulation: a survey and perspectives}},\
  }\href {https://doi.org/10.1057/s41599-024-03611-3} {\bibfield  {journal}
  {\bibinfo  {journal} {Humanit Soc Sci Commun}\ }\textbf {\bibinfo {volume}
  {11}},\ \bibinfo {pages} {1} (\bibinfo {year} {2024})},\ \bibinfo {note}
  {publisher: Palgrave}\BibitemShut {NoStop}%
\bibitem [{\citenamefont {Abuelsaad}\ \emph {et~al.}(2024)\citenamefont
  {Abuelsaad}, \citenamefont {Akkil}, \citenamefont {Dey}, \citenamefont
  {Jagmohan}, \citenamefont {Vempaty},\ and\ \citenamefont
  {Kokku}}]{abuelsaad_agent-e_2024}%
  \BibitemOpen
  \bibfield  {author} {\bibinfo {author} {\bibfnamefont {T.}~\bibnamefont
  {Abuelsaad}}, \bibinfo {author} {\bibfnamefont {D.}~\bibnamefont {Akkil}},
  \bibinfo {author} {\bibfnamefont {P.}~\bibnamefont {Dey}}, \bibinfo {author}
  {\bibfnamefont {A.}~\bibnamefont {Jagmohan}}, \bibinfo {author}
  {\bibfnamefont {A.}~\bibnamefont {Vempaty}},\ and\ \bibinfo {author}
  {\bibfnamefont {R.}~\bibnamefont {Kokku}},\ }\href
  {https://doi.org/10.48550/arXiv.2407.13032} {\bibinfo {title} {Agent-{E}:
  {From} {Autonomous} {Web} {Navigation} to {Foundational} {Design}
  {Principles} in {Agentic} {Systems}}} (\bibinfo {year} {2024}),\ \bibinfo
  {note} {arXiv:2407.13032 [cs]}\BibitemShut {NoStop}%
\bibitem [{\citenamefont {Chen}\ \emph {et~al.}(2024)\citenamefont {Chen},
  \citenamefont {Jiang}, \citenamefont {Lu},\ and\ \citenamefont
  {Zhang}}]{chen_s-agents_2024}%
  \BibitemOpen
  \bibfield  {author} {\bibinfo {author} {\bibfnamefont {J.}~\bibnamefont
  {Chen}}, \bibinfo {author} {\bibfnamefont {Y.}~\bibnamefont {Jiang}},
  \bibinfo {author} {\bibfnamefont {J.}~\bibnamefont {Lu}},\ and\ \bibinfo
  {author} {\bibfnamefont {L.}~\bibnamefont {Zhang}},\ }\href
  {https://doi.org/10.48550/arXiv.2402.04578} {\bibinfo {title} {S-{Agents}:
  {Self}-organizing {Agents} in {Open}-ended {Environments}}} (\bibinfo {year}
  {2024}),\ \bibinfo {note} {arXiv:2402.04578 [cs]}\BibitemShut {NoStop}%
\bibitem [{\citenamefont {Liu}\ \emph {et~al.}(2024)\citenamefont {Liu},
  \citenamefont {Lo}, \citenamefont {Lu}, \citenamefont {Zhu}, \citenamefont
  {Zhao}, \citenamefont {Xu}, \citenamefont {Harrer},\ and\ \citenamefont
  {Whittle}}]{liu_agent_2024}%
  \BibitemOpen
  \bibfield  {author} {\bibinfo {author} {\bibfnamefont {Y.}~\bibnamefont
  {Liu}}, \bibinfo {author} {\bibfnamefont {S.~K.}\ \bibnamefont {Lo}},
  \bibinfo {author} {\bibfnamefont {Q.}~\bibnamefont {Lu}}, \bibinfo {author}
  {\bibfnamefont {L.}~\bibnamefont {Zhu}}, \bibinfo {author} {\bibfnamefont
  {D.}~\bibnamefont {Zhao}}, \bibinfo {author} {\bibfnamefont {X.}~\bibnamefont
  {Xu}}, \bibinfo {author} {\bibfnamefont {S.}~\bibnamefont {Harrer}},\ and\
  \bibinfo {author} {\bibfnamefont {J.}~\bibnamefont {Whittle}},\ }\href
  {https://doi.org/10.48550/arXiv.2405.10467} {\bibinfo {title} {Agent {Design}
  {Pattern} {Catalogue}: {A} {Collection} of {Architectural} {Patterns} for
  {Foundation} {Model} based {Agents}}} (\bibinfo {year} {2024}),\ \bibinfo
  {note} {arXiv:2405.10467 [cs]}\BibitemShut {NoStop}%
\bibitem [{\citenamefont {Millidge}(2021)}]{millidge_applications_2021}%
  \BibitemOpen
  \bibfield  {author} {\bibinfo {author} {\bibfnamefont {B.}~\bibnamefont
  {Millidge}},\ }\href {https://doi.org/10.48550/arXiv.2107.00140} {\bibinfo
  {title} {Applications of the {Free} {Energy} {Principle} to {Machine}
  {Learning} and {Neuroscience}}} (\bibinfo {year} {2021}),\ \bibinfo {note}
  {arXiv:2107.00140 [cs]}\BibitemShut {NoStop}%
\bibitem [{\citenamefont {Millidge}\ \emph {et~al.}(2020)\citenamefont
  {Millidge}, \citenamefont {Tschantz}, \citenamefont {Seth},\ and\
  \citenamefont {Buckley}}]{millidge_relationship_2020}%
  \BibitemOpen
  \bibfield  {author} {\bibinfo {author} {\bibfnamefont {B.}~\bibnamefont
  {Millidge}}, \bibinfo {author} {\bibfnamefont {A.}~\bibnamefont {Tschantz}},
  \bibinfo {author} {\bibfnamefont {A.~K.}\ \bibnamefont {Seth}},\ and\
  \bibinfo {author} {\bibfnamefont {C.~L.}\ \bibnamefont {Buckley}},\
  }\bibfield  {title} {{\selectlanguage {English}\bibinfo {title} {On the
  {Relationship} {Between} {Active} {Inference} and {Control} as
  {Inference}}},\ }in\ \href {https://doi.org/10.1007/978-3-030-64919-7_1}
  {{\selectlanguage {English}\emph {\bibinfo {booktitle} {Active
  {Inference}}}}},\ \bibinfo {editor} {edited by\ \bibinfo {editor}
  {\bibfnamefont {T.}~\bibnamefont {Verbelen}}, \bibinfo {editor}
  {\bibfnamefont {P.}~\bibnamefont {Lanillos}}, \bibinfo {editor}
  {\bibfnamefont {C.~L.}\ \bibnamefont {Buckley}},\ and\ \bibinfo {editor}
  {\bibfnamefont {C.}~\bibnamefont {De~Boom}}}\ (\bibinfo  {publisher}
  {Springer International Publishing},\ \bibinfo {address} {Cham},\ \bibinfo
  {year} {2020})\ pp.\ \bibinfo {pages} {3--11}\BibitemShut {NoStop}%
\bibitem [{\citenamefont {Friston}\ \emph {et~al.}(2024)\citenamefont
  {Friston}, \citenamefont {Heins}, \citenamefont {Verbelen}, \citenamefont
  {Costa}, \citenamefont {Salvatori}, \citenamefont {Markovic}, \citenamefont
  {Tschantz}, \citenamefont {Koudahl}, \citenamefont {Buckley},\ and\
  \citenamefont {Parr}}]{friston_pixels_2024}%
  \BibitemOpen
  \bibfield  {author} {\bibinfo {author} {\bibfnamefont {K.}~\bibnamefont
  {Friston}}, \bibinfo {author} {\bibfnamefont {C.}~\bibnamefont {Heins}},
  \bibinfo {author} {\bibfnamefont {T.}~\bibnamefont {Verbelen}}, \bibinfo
  {author} {\bibfnamefont {L.~D.}\ \bibnamefont {Costa}}, \bibinfo {author}
  {\bibfnamefont {T.}~\bibnamefont {Salvatori}}, \bibinfo {author}
  {\bibfnamefont {D.}~\bibnamefont {Markovic}}, \bibinfo {author}
  {\bibfnamefont {A.}~\bibnamefont {Tschantz}}, \bibinfo {author}
  {\bibfnamefont {M.}~\bibnamefont {Koudahl}}, \bibinfo {author} {\bibfnamefont
  {C.}~\bibnamefont {Buckley}},\ and\ \bibinfo {author} {\bibfnamefont
  {T.}~\bibnamefont {Parr}},\ }\href
  {https://doi.org/10.48550/arXiv.2407.20292} {\bibinfo {title} {From pixels to
  planning: scale-free active inference}} (\bibinfo {year} {2024}),\ \bibinfo
  {note} {arXiv:2407.20292 [cs]}\BibitemShut {NoStop}%
\bibitem [{\citenamefont {Ramstead}\ \emph {et~al.}(2025)\citenamefont
  {Ramstead}, \citenamefont {Sakthivadivel},\ and\ \citenamefont
  {Friston}}]{ramstead_framework_2025}%
  \BibitemOpen
  \bibfield  {author} {\bibinfo {author} {\bibfnamefont {M.~J.~D.}\
  \bibnamefont {Ramstead}}, \bibinfo {author} {\bibfnamefont {D.~A.~R.}\
  \bibnamefont {Sakthivadivel}},\ and\ \bibinfo {author} {\bibfnamefont
  {K.~J.}\ \bibnamefont {Friston}},\ }\href
  {https://doi.org/10.48550/arXiv.2406.11630} {\bibinfo {title} {A framework
  for the use of generative modelling in non-equilibrium statistical
  mechanics}} (\bibinfo {year} {2025}),\ \bibinfo {note} {arXiv:2406.11630
  [cond-mat]}\BibitemShut {NoStop}%
\bibitem [{\citenamefont {Beck}\ and\ \citenamefont
  {Ramstead}(2025)}]{beck_dynamic_2025}%
  \BibitemOpen
  \bibfield  {author} {\bibinfo {author} {\bibfnamefont {J.}~\bibnamefont
  {Beck}}\ and\ \bibinfo {author} {\bibfnamefont {M.~J.~D.}\ \bibnamefont
  {Ramstead}},\ }\href {https://doi.org/10.48550/arXiv.2502.21217} {\bibinfo
  {title} {Dynamic {Markov} {Blanket} {Detection} for {Macroscopic} {Physics}
  {Discovery}}} (\bibinfo {year} {2025}),\ \bibinfo {note} {arXiv:2502.21217
  [q-bio]}\BibitemShut {NoStop}%
\bibitem [{\citenamefont {Meindl}\ \emph {et~al.}(2022)\citenamefont {Meindl},
  \citenamefont {Lehmann},\ and\ \citenamefont {Seel}}]{meindl_bridging_2022}%
  \BibitemOpen
  \bibfield  {author} {\bibinfo {author} {\bibfnamefont {M.}~\bibnamefont
  {Meindl}}, \bibinfo {author} {\bibfnamefont {D.}~\bibnamefont {Lehmann}},\
  and\ \bibinfo {author} {\bibfnamefont {T.}~\bibnamefont {Seel}},\ }\bibfield
  {title} {{\selectlanguage {English}\bibinfo {title} {Bridging {Reinforcement}
  {Learning} and {Iterative} {Learning} {Control}: {Autonomous} {Motion}
  {Learning} for {Unknown}, {Nonlinear} {Dynamics}}},\ }\bibfield  {journal}
  {\bibinfo  {journal} {Front. Robot. AI}\ }\textbf {\bibinfo {volume} {9}},\
  \href {https://doi.org/10.3389/frobt.2022.793512} {10.3389/frobt.2022.793512}
  (\bibinfo {year} {2022}),\ \bibinfo {note} {publisher: Frontiers}\BibitemShut
  {NoStop}%
\bibitem [{\citenamefont {Tsividis}\ \emph {et~al.}(2021)\citenamefont
  {Tsividis}, \citenamefont {Loula}, \citenamefont {Burga}, \citenamefont
  {Foss}, \citenamefont {Campero}, \citenamefont {Pouncy}, \citenamefont
  {Gershman},\ and\ \citenamefont {Tenenbaum}}]{tsividis_human-level_2021}%
  \BibitemOpen
  \bibfield  {author} {\bibinfo {author} {\bibfnamefont {P.~A.}\ \bibnamefont
  {Tsividis}}, \bibinfo {author} {\bibfnamefont {J.}~\bibnamefont {Loula}},
  \bibinfo {author} {\bibfnamefont {J.}~\bibnamefont {Burga}}, \bibinfo
  {author} {\bibfnamefont {N.}~\bibnamefont {Foss}}, \bibinfo {author}
  {\bibfnamefont {A.}~\bibnamefont {Campero}}, \bibinfo {author} {\bibfnamefont
  {T.}~\bibnamefont {Pouncy}}, \bibinfo {author} {\bibfnamefont {S.~J.}\
  \bibnamefont {Gershman}},\ and\ \bibinfo {author} {\bibfnamefont {J.~B.}\
  \bibnamefont {Tenenbaum}},\ }\href
  {https://doi.org/10.48550/ARXIV.2107.12544} {\bibinfo {title} {Human-{Level}
  {Reinforcement} {Learning} through {Theory}-{Based} {Modeling},
  {Exploration}, and {Planning}}} (\bibinfo {year} {2021}),\ \bibinfo {note}
  {version Number: 1}\BibitemShut {NoStop}%
\bibitem [{\citenamefont {Langley}(2006)}]{langley_cognitive_2006}%
  \BibitemOpen
  \bibfield  {author} {\bibinfo {author} {\bibfnamefont {P.}~\bibnamefont
  {Langley}},\ }\bibfield  {title} {{\selectlanguage {English}\bibinfo {title}
  {Cognitive {Architectures} and {General} {Intelligent} {Systems}}},\ }\href
  {https://doi.org/10.1609/aimag.v27i2.1878} {\bibfield  {journal} {\bibinfo
  {journal} {AI Magazine}\ }\textbf {\bibinfo {volume} {27}},\ \bibinfo {pages}
  {33} (\bibinfo {year} {2006})},\ \bibinfo {note} {number: 2}\BibitemShut
  {NoStop}%
\bibitem [{\citenamefont {Blili-Hamelin}\ \emph {et~al.}(2025)\citenamefont
  {Blili-Hamelin}, \citenamefont {Graziul}, \citenamefont {Hancox-Li},
  \citenamefont {Hazan}, \citenamefont {El-Mhamdi}, \citenamefont {Ghosh},
  \citenamefont {Heller}, \citenamefont {Metcalf}, \citenamefont {Murai},
  \citenamefont {Salvaggio}, \citenamefont {Smart}, \citenamefont {Snider},
  \citenamefont {Tighanimine}, \citenamefont {Ringer}, \citenamefont
  {Mitchell},\ and\ \citenamefont {Dori-Hacohen}}]{blili-hamelin_stop_2025}%
  \BibitemOpen
  \bibfield  {author} {\bibinfo {author} {\bibfnamefont {B.}~\bibnamefont
  {Blili-Hamelin}}, \bibinfo {author} {\bibfnamefont {C.}~\bibnamefont
  {Graziul}}, \bibinfo {author} {\bibfnamefont {L.}~\bibnamefont {Hancox-Li}},
  \bibinfo {author} {\bibfnamefont {H.}~\bibnamefont {Hazan}}, \bibinfo
  {author} {\bibfnamefont {E.-M.}\ \bibnamefont {El-Mhamdi}}, \bibinfo {author}
  {\bibfnamefont {A.}~\bibnamefont {Ghosh}}, \bibinfo {author} {\bibfnamefont
  {K.}~\bibnamefont {Heller}}, \bibinfo {author} {\bibfnamefont
  {J.}~\bibnamefont {Metcalf}}, \bibinfo {author} {\bibfnamefont
  {F.}~\bibnamefont {Murai}}, \bibinfo {author} {\bibfnamefont
  {E.}~\bibnamefont {Salvaggio}}, \bibinfo {author} {\bibfnamefont
  {A.}~\bibnamefont {Smart}}, \bibinfo {author} {\bibfnamefont
  {T.}~\bibnamefont {Snider}}, \bibinfo {author} {\bibfnamefont
  {M.}~\bibnamefont {Tighanimine}}, \bibinfo {author} {\bibfnamefont
  {T.}~\bibnamefont {Ringer}}, \bibinfo {author} {\bibfnamefont
  {M.}~\bibnamefont {Mitchell}},\ and\ \bibinfo {author} {\bibfnamefont
  {S.}~\bibnamefont {Dori-Hacohen}},\ }\href
  {https://doi.org/10.48550/arXiv.2502.03689} {{\selectlanguage
  {English}\bibinfo {title} {Stop treating `{AGI}' as the north-star goal of
  {AI} research}}} (\bibinfo {year} {2025}),\ \bibinfo {note} {arXiv:2502.03689
  [cs]}\BibitemShut {NoStop}%
\bibitem [{\citenamefont {Jaeger}(2024)}]{jaeger_artificial_2024}%
  \BibitemOpen
  \bibfield  {author} {\bibinfo {author} {\bibfnamefont {J.}~\bibnamefont
  {Jaeger}},\ }\bibfield  {title} {\bibinfo {title} {Artificial intelligence is
  algorithmic mimicry: why artificial "agents" are not (and won't be) proper
  agents},\ }\bibfield  {journal} {\bibinfo  {journal} {Neurons, Behavior, Data
  analysis, and Theory}\ }\href {https://doi.org/10.51628/001c.94404}
  {10.51628/001c.94404} (\bibinfo {year} {2024}),\ \bibinfo {note}
  {arXiv:2307.07515 [cs]}\BibitemShut {NoStop}%
\bibitem [{\citenamefont {Kagan}\ \emph {et~al.}(2024)\citenamefont {Kagan},
  \citenamefont {Mahlis}, \citenamefont {Bhat}, \citenamefont {Bongard},
  \citenamefont {Cole}, \citenamefont {Corlett}, \citenamefont {Gyngell},
  \citenamefont {Hartung}, \citenamefont {Jupp}, \citenamefont {Levin},
  \citenamefont {Lysaght}, \citenamefont {Opie}, \citenamefont {Razi},
  \citenamefont {Smirnova}, \citenamefont {Tennant}, \citenamefont {Wade},\
  and\ \citenamefont {Wang}}]{kagan_toward_2024}%
  \BibitemOpen
  \bibfield  {author} {\bibinfo {author} {\bibfnamefont {B.~J.}\ \bibnamefont
  {Kagan}}, \bibinfo {author} {\bibfnamefont {M.}~\bibnamefont {Mahlis}},
  \bibinfo {author} {\bibfnamefont {A.}~\bibnamefont {Bhat}}, \bibinfo {author}
  {\bibfnamefont {J.}~\bibnamefont {Bongard}}, \bibinfo {author} {\bibfnamefont
  {V.~M.}\ \bibnamefont {Cole}}, \bibinfo {author} {\bibfnamefont
  {P.}~\bibnamefont {Corlett}}, \bibinfo {author} {\bibfnamefont
  {C.}~\bibnamefont {Gyngell}}, \bibinfo {author} {\bibfnamefont
  {T.}~\bibnamefont {Hartung}}, \bibinfo {author} {\bibfnamefont
  {B.}~\bibnamefont {Jupp}}, \bibinfo {author} {\bibfnamefont {M.}~\bibnamefont
  {Levin}}, \bibinfo {author} {\bibfnamefont {T.}~\bibnamefont {Lysaght}},
  \bibinfo {author} {\bibfnamefont {N.}~\bibnamefont {Opie}}, \bibinfo {author}
  {\bibfnamefont {A.}~\bibnamefont {Razi}}, \bibinfo {author} {\bibfnamefont
  {L.}~\bibnamefont {Smirnova}}, \bibinfo {author} {\bibfnamefont
  {I.}~\bibnamefont {Tennant}}, \bibinfo {author} {\bibfnamefont {P.~T.}\
  \bibnamefont {Wade}},\ and\ \bibinfo {author} {\bibfnamefont
  {G.}~\bibnamefont {Wang}},\ }\bibfield  {title} {{\selectlanguage
  {English}\bibinfo {title} {Toward a nomenclature consensus for diverse
  intelligent systems: {Call} for collaboration}},\ }\href
  {https://doi.org/10.1016/j.xinn.2024.100658} {\bibfield  {journal} {\bibinfo
  {journal} {The Innovation}\ }\textbf {\bibinfo {volume} {5}},\ \bibinfo
  {pages} {100658} (\bibinfo {year} {2024})}\BibitemShut {NoStop}%
\bibitem [{\citenamefont {Friston}\ \emph {et~al.}(2023)\citenamefont
  {Friston}, \citenamefont {Friedman}, \citenamefont {Constant}, \citenamefont
  {Knight}, \citenamefont {Fields}, \citenamefont {Parr},\ and\ \citenamefont
  {Campbell}}]{friston_variational_2023}%
  \BibitemOpen
  \bibfield  {author} {\bibinfo {author} {\bibfnamefont {K.}~\bibnamefont
  {Friston}}, \bibinfo {author} {\bibfnamefont {D.~A.}\ \bibnamefont
  {Friedman}}, \bibinfo {author} {\bibfnamefont {A.}~\bibnamefont {Constant}},
  \bibinfo {author} {\bibfnamefont {V.~B.}\ \bibnamefont {Knight}}, \bibinfo
  {author} {\bibfnamefont {C.}~\bibnamefont {Fields}}, \bibinfo {author}
  {\bibfnamefont {T.}~\bibnamefont {Parr}},\ and\ \bibinfo {author}
  {\bibfnamefont {J.~O.}\ \bibnamefont {Campbell}},\ }\bibfield  {title}
  {{\selectlanguage {English}\bibinfo {title} {A {Variational} {Synthesis} of
  {Evolutionary} and {Developmental} {Dynamics}}},\ }\href
  {https://doi.org/10.3390/e25070964} {\bibfield  {journal} {\bibinfo
  {journal} {Entropy}\ }\textbf {\bibinfo {volume} {25}},\ \bibinfo {pages}
  {964} (\bibinfo {year} {2023})}\BibitemShut {NoStop}%
\bibitem [{\citenamefont {Cole}(2024)}]{cole_cognitive_2024}%
  \BibitemOpen
  \bibfield  {author} {\bibinfo {author} {\bibfnamefont {M.~W.}\ \bibnamefont
  {Cole}},\ }\bibfield  {title} {{\selectlanguage {English}\bibinfo {title}
  {Cognitive flexibility as the shifting of brain network flows by flexible
  neural representations}},\ }\href
  {https://doi.org/10.1016/j.cobeha.2024.101384} {\bibfield  {journal}
  {\bibinfo  {journal} {Current Opinion in Behavioral Sciences}\ }\textbf
  {\bibinfo {volume} {57}},\ \bibinfo {pages} {101384} (\bibinfo {year}
  {2024})}\BibitemShut {NoStop}%
\bibitem [{\citenamefont {Papayannopoulos}\ \emph {et~al.}(2022)\citenamefont
  {Papayannopoulos}, \citenamefont {Fresco},\ and\ \citenamefont
  {Shagrir}}]{papayannopoulos_computational_2022}%
  \BibitemOpen
  \bibfield  {author} {\bibinfo {author} {\bibfnamefont {P.}~\bibnamefont
  {Papayannopoulos}}, \bibinfo {author} {\bibfnamefont {N.}~\bibnamefont
  {Fresco}},\ and\ \bibinfo {author} {\bibfnamefont {O.}~\bibnamefont
  {Shagrir}},\ }\bibfield  {title} {{\selectlanguage {English}\bibinfo {title}
  {Computational indeterminacy and explanations in cognitive science}},\ }\href
  {https://doi.org/10.1007/s10539-022-09877-8} {\bibfield  {journal} {\bibinfo
  {journal} {Biol Philos}\ }\textbf {\bibinfo {volume} {37}},\ \bibinfo {pages}
  {47} (\bibinfo {year} {2022})}\BibitemShut {NoStop}%
\bibitem [{\citenamefont {Makin}\ and\ \citenamefont
  {Krakauer}(2023)}]{makin_against_2023}%
  \BibitemOpen
  \bibfield  {author} {\bibinfo {author} {\bibfnamefont {T.~R.}\ \bibnamefont
  {Makin}}\ and\ \bibinfo {author} {\bibfnamefont {J.~W.}\ \bibnamefont
  {Krakauer}},\ }\bibfield  {title} {{\selectlanguage {English}\bibinfo {title}
  {Against cortical reorganisation}},\ }\href
  {https://doi.org/10.7554/eLife.84716} {\bibfield  {journal} {\bibinfo
  {journal} {eLife}\ }\textbf {\bibinfo {volume} {12}},\ \bibinfo {pages}
  {e84716} (\bibinfo {year} {2023})}\BibitemShut {NoStop}%
\bibitem [{\citenamefont {Evans}\ and\ \citenamefont
  {Foster}(2024)}]{evans_algorithmic_2024}%
  \BibitemOpen
  \bibfield  {author} {\bibinfo {author} {\bibfnamefont {J.~A.}\ \bibnamefont
  {Evans}}\ and\ \bibinfo {author} {\bibfnamefont {J.~G.}\ \bibnamefont
  {Foster}},\ }\bibfield  {title} {{\selectlanguage {English}\bibinfo {title}
  {Algorithmic {Abduction}: {Robots} for {Alien} {Reading}}},\ }\href
  {https://doi.org/10.1086/728933} {\bibfield  {journal} {\bibinfo  {journal}
  {Critical Inquiry}\ }\textbf {\bibinfo {volume} {50}},\ \bibinfo {pages}
  {375} (\bibinfo {year} {2024})}\BibitemShut {NoStop}%
\bibitem [{\citenamefont {Wooldridge}\ and\ \citenamefont
  {Jennings}(1995{\natexlab{b}})}]{wooldridge_intelligent_1995}%
  \BibitemOpen
  \bibfield  {author} {\bibinfo {author} {\bibfnamefont {M.}~\bibnamefont
  {Wooldridge}}\ and\ \bibinfo {author} {\bibfnamefont {N.~R.}\ \bibnamefont
  {Jennings}},\ }\bibfield  {title} {{\selectlanguage {English}\bibinfo {title}
  {Intelligent agents: theory and practice}},\ }\href
  {https://doi.org/10.1017/S0269888900008122} {\bibfield  {journal} {\bibinfo
  {journal} {The Knowledge Engineering Review}\ }\textbf {\bibinfo {volume}
  {10}},\ \bibinfo {pages} {115} (\bibinfo {year}
  {1995}{\natexlab{b}})}\BibitemShut {NoStop}%
\bibitem [{\citenamefont {Koivisto}\ and\ \citenamefont
  {Grassini}(2023)}]{koivisto_best_2023}%
  \BibitemOpen
  \bibfield  {author} {\bibinfo {author} {\bibfnamefont {M.}~\bibnamefont
  {Koivisto}}\ and\ \bibinfo {author} {\bibfnamefont {S.}~\bibnamefont
  {Grassini}},\ }\bibfield  {title} {{\selectlanguage {English}\bibinfo {title}
  {Best humans still outperform artificial intelligence in a creative divergent
  thinking task}},\ }\href {https://doi.org/10.1038/s41598-023-40858-3}
  {\bibfield  {journal} {\bibinfo  {journal} {Sci Rep}\ }\textbf {\bibinfo
  {volume} {13}},\ \bibinfo {pages} {13601} (\bibinfo {year}
  {2023})}\BibitemShut {NoStop}%
\bibitem [{\citenamefont {Lee}(2022)}]{lee_what_2022}%
  \BibitemOpen
  \bibfield  {author} {\bibinfo {author} {\bibfnamefont {E.~A.}\ \bibnamefont
  {Lee}},\ }\bibfield  {title} {\bibinfo {title} {What {Can} {Deep} {Neural}
  {Networks} {Teach} {Us} {About} {Embodied} {Bounded} {Rationality}},\ }\href
  {https://doi.org/10.3389/fpsyg.2022.761808} {\bibfield  {journal} {\bibinfo
  {journal} {Front. Psychol.}\ }\textbf {\bibinfo {volume} {13}},\ \bibinfo
  {pages} {761808} (\bibinfo {year} {2022})}\BibitemShut {NoStop}%
\bibitem [{\citenamefont {Shanahan}(2024)}]{shanahan_simulacra_2024}%
  \BibitemOpen
  \bibfield  {author} {\bibinfo {author} {\bibfnamefont {M.}~\bibnamefont
  {Shanahan}},\ }\href {https://doi.org/10.48550/ARXIV.2402.12422} {\bibinfo
  {title} {Simulacra as {Conscious} {Exotica}}} (\bibinfo {year} {2024}),\
  \bibinfo {note} {version Number: 2}\BibitemShut {NoStop}%
\bibitem [{\citenamefont {Pessoa}\ \emph {et~al.}(2022)\citenamefont {Pessoa},
  \citenamefont {Medina},\ and\ \citenamefont
  {Desfilis}}]{pessoa_refocusing_2022}%
  \BibitemOpen
  \bibfield  {author} {\bibinfo {author} {\bibfnamefont {L.}~\bibnamefont
  {Pessoa}}, \bibinfo {author} {\bibfnamefont {L.}~\bibnamefont {Medina}},\
  and\ \bibinfo {author} {\bibfnamefont {E.}~\bibnamefont {Desfilis}},\
  }\bibfield  {title} {{\selectlanguage {English}\bibinfo {title} {Refocusing
  neuroscience: moving away from mental categories and towards complex
  behaviours}},\ }\href {https://doi.org/10.1098/rstb.2020.0534} {\bibfield
  {journal} {\bibinfo  {journal} {Phil. Trans. R. Soc. B}\ }\textbf {\bibinfo
  {volume} {377}},\ \bibinfo {pages} {20200534} (\bibinfo {year}
  {2022})}\BibitemShut {NoStop}%
\bibitem [{\citenamefont {Solé}\ and\ \citenamefont
  {Seoane}(2022)}]{sole_evolution_2022}%
  \BibitemOpen
  \bibfield  {author} {\bibinfo {author} {\bibfnamefont {R.}~\bibnamefont
  {Solé}}\ and\ \bibinfo {author} {\bibfnamefont {L.~F.}\ \bibnamefont
  {Seoane}},\ }\bibfield  {title} {{\selectlanguage {English}\bibinfo {title}
  {Evolution of {Brains} and {Computers}: {The} {Roads} {Not} {Taken}}},\
  }\href {https://doi.org/10.3390/e24050665} {\bibfield  {journal} {\bibinfo
  {journal} {Entropy}\ }\textbf {\bibinfo {volume} {24}},\ \bibinfo {pages}
  {665} (\bibinfo {year} {2022})}\BibitemShut {NoStop}%
\bibitem [{\citenamefont {Bechtel}\ and\ \citenamefont
  {Bich}(2021)}]{bechtel_grounding_2021}%
  \BibitemOpen
  \bibfield  {author} {\bibinfo {author} {\bibfnamefont {W.}~\bibnamefont
  {Bechtel}}\ and\ \bibinfo {author} {\bibfnamefont {L.}~\bibnamefont {Bich}},\
  }\bibfield  {title} {{\selectlanguage {English}\bibinfo {title} {Grounding
  cognition: heterarchical control mechanisms in biology}},\ }\href
  {https://doi.org/10.1098/rstb.2019.0751} {\bibfield  {journal} {\bibinfo
  {journal} {Phil. Trans. R. Soc. B}\ }\textbf {\bibinfo {volume} {376}},\
  \bibinfo {pages} {20190751} (\bibinfo {year} {2021})}\BibitemShut {NoStop}%
\bibitem [{\citenamefont {Barandiaran}\ \emph {et~al.}(2009)\citenamefont
  {Barandiaran}, \citenamefont {Di~Paolo},\ and\ \citenamefont
  {Rohde}}]{barandiaran_defining_2009}%
  \BibitemOpen
  \bibfield  {author} {\bibinfo {author} {\bibfnamefont {X.~E.}\ \bibnamefont
  {Barandiaran}}, \bibinfo {author} {\bibfnamefont {E.}~\bibnamefont
  {Di~Paolo}},\ and\ \bibinfo {author} {\bibfnamefont {M.}~\bibnamefont
  {Rohde}},\ }\bibfield  {title} {{\selectlanguage {English}\bibinfo {title}
  {Defining {Agency}: {Individuality}, {Normativity}, {Asymmetry}, and
  {Spatio}-temporality in {Action}}},\ }\href
  {https://doi.org/10.1177/1059712309343819} {\bibfield  {journal} {\bibinfo
  {journal} {Adaptive Behavior}\ }\textbf {\bibinfo {volume} {17}},\ \bibinfo
  {pages} {367} (\bibinfo {year} {2009})}\BibitemShut {NoStop}%
\bibitem [{\citenamefont {Ciaunica}\ \emph {et~al.}(2023)\citenamefont
  {Ciaunica}, \citenamefont {Levin}, \citenamefont {Rosas},\ and\ \citenamefont
  {Friston}}]{ciaunica_nested_nodate}%
  \BibitemOpen
  \bibfield  {author} {\bibinfo {author} {\bibfnamefont {A.}~\bibnamefont
  {Ciaunica}}, \bibinfo {author} {\bibfnamefont {M.}~\bibnamefont {Levin}},
  \bibinfo {author} {\bibfnamefont {F.~E.}\ \bibnamefont {Rosas}},\ and\
  \bibinfo {author} {\bibfnamefont {K.}~\bibnamefont {Friston}},\ }\bibfield
  {title} {{\selectlanguage {English}\bibinfo {title} {Nested {Selves}:
  {Self}-{Organization} and {Shared} {Markov} {Blankets} in {Prenatal}
  {Development} in {Humans}}},\ }\href {https://doi.org/10.1111/tops.12717}
  {\bibfield  {journal} {\bibinfo  {journal} {Topics in Cognitive Science}\
  }\textbf {\bibinfo {volume} {00}},\ \bibinfo {pages} {1} (\bibinfo {year}
  {2023})},\ \bibinfo {note} {\_eprint:
  https://onlinelibrary.wiley.com/doi/pdf/10.1111/tops.12717}\BibitemShut
  {NoStop}%
\bibitem [{\citenamefont {Davies}\ and\ \citenamefont
  {Levin}(2023)}]{davies_synthetic_2023}%
  \BibitemOpen
  \bibfield  {author} {\bibinfo {author} {\bibfnamefont {J.}~\bibnamefont
  {Davies}}\ and\ \bibinfo {author} {\bibfnamefont {M.}~\bibnamefont {Levin}},\
  }\bibfield  {title} {{\selectlanguage {English}\bibinfo {title} {Synthetic
  morphology with agential materials}},\ }\href
  {https://doi.org/10.1038/s44222-022-00001-9} {\bibfield  {journal} {\bibinfo
  {journal} {Nat Rev Bioeng}\ }\textbf {\bibinfo {volume} {1}},\ \bibinfo
  {pages} {46} (\bibinfo {year} {2023})}\BibitemShut {NoStop}%
\bibitem [{\citenamefont {Kozachkov}\ \emph {et~al.}(2023)\citenamefont
  {Kozachkov}, \citenamefont {Kastanenka},\ and\ \citenamefont
  {Krotov}}]{kozachkov_building_2023}%
  \BibitemOpen
  \bibfield  {author} {\bibinfo {author} {\bibfnamefont {L.}~\bibnamefont
  {Kozachkov}}, \bibinfo {author} {\bibfnamefont {K.~V.}\ \bibnamefont
  {Kastanenka}},\ and\ \bibinfo {author} {\bibfnamefont {D.}~\bibnamefont
  {Krotov}},\ }\bibfield  {title} {{\selectlanguage {English}\bibinfo {title}
  {Building transformers from neurons and astrocytes}},\ }\href
  {https://doi.org/10.1073/pnas.2219150120} {\bibfield  {journal} {\bibinfo
  {journal} {Proc. Natl. Acad. Sci. U.S.A.}\ }\textbf {\bibinfo {volume}
  {120}},\ \bibinfo {pages} {e2219150120} (\bibinfo {year} {2023})}\BibitemShut
  {NoStop}%
\bibitem [{\citenamefont {Bruni}\ and\ \citenamefont
  {Giorgi}(2015)}]{bruni_towards_2015}%
  \BibitemOpen
  \bibfield  {author} {\bibinfo {author} {\bibfnamefont {L.~E.}\ \bibnamefont
  {Bruni}}\ and\ \bibinfo {author} {\bibfnamefont {F.}~\bibnamefont {Giorgi}},\
  }\bibfield  {title} {{\selectlanguage {English}\bibinfo {title} {Towards a
  heterarchical approach to biology and cognition}},\ }\href
  {https://doi.org/10.1016/j.pbiomolbio.2015.07.005} {\bibfield  {journal}
  {\bibinfo  {journal} {Progress in Biophysics and Molecular Biology}\ }\textbf
  {\bibinfo {volume} {119}},\ \bibinfo {pages} {481} (\bibinfo {year}
  {2015})}\BibitemShut {NoStop}%
\bibitem [{\citenamefont {Pessoa}(2023)}]{pessoa_spiraling_2023}%
  \BibitemOpen
  \bibfield  {author} {\bibinfo {author} {\bibfnamefont {L.}~\bibnamefont
  {Pessoa}},\ }\href {https://doi.org/10.31219/osf.io/n7z6u} {\bibinfo {title}
  {The spiraling brain: {Combinatorial}, reciprocal, and reentrant
  macro-organization}} (\bibinfo {year} {2023})\BibitemShut {NoStop}%
\bibitem [{\citenamefont {Kowerdziej}\ \emph {et~al.}(2022)\citenamefont
  {Kowerdziej}, \citenamefont {Ferraro}, \citenamefont {Zografopoulos},\ and\
  \citenamefont {Caputo}}]{kowerdziej_softmatterbased_2022}%
  \BibitemOpen
  \bibfield  {author} {\bibinfo {author} {\bibfnamefont {R.}~\bibnamefont
  {Kowerdziej}}, \bibinfo {author} {\bibfnamefont {A.}~\bibnamefont {Ferraro}},
  \bibinfo {author} {\bibfnamefont {D.~C.}\ \bibnamefont {Zografopoulos}},\
  and\ \bibinfo {author} {\bibfnamefont {R.}~\bibnamefont {Caputo}},\
  }\bibfield  {title} {{\selectlanguage {English}\bibinfo {title}
  {Soft‐{Matter}‐{Based} {Hybrid} and {Active} {Metamaterials}}},\ }\href
  {https://doi.org/10.1002/adom.202200750} {\bibfield  {journal} {\bibinfo
  {journal} {Advanced Optical Materials}\ }\textbf {\bibinfo {volume} {10}},\
  \bibinfo {pages} {2200750} (\bibinfo {year} {2022})}\BibitemShut {NoStop}%
\bibitem [{\citenamefont {Baulin}\ \emph
  {et~al.}(2025{\natexlab{a}})\citenamefont {Baulin}, \citenamefont
  {Giacometti}, \citenamefont {Fedosov}, \citenamefont {Ebbens}, \citenamefont
  {Varela-Rosales}, \citenamefont {Feliu}, \citenamefont {Chowdhury},
  \citenamefont {Hu}, \citenamefont {Füchslin}, \citenamefont {Dijkstra},
  \citenamefont {Mussel}, \citenamefont {Van~Roij}, \citenamefont {Xie},
  \citenamefont {Tzanov}, \citenamefont {Zu}, \citenamefont
  {Hidalgo-Caballero}, \citenamefont {Yuan}, \citenamefont {Cocconi},
  \citenamefont {Ghim}, \citenamefont {Cottin-Bizonne}, \citenamefont {Miguel},
  \citenamefont {Esplandiu}, \citenamefont {Simmchen}, \citenamefont {Parak},
  \citenamefont {Werner}, \citenamefont {Gompper},\ and\ \citenamefont
  {Hanczyc}}]{baulin_intelligent_2025}%
  \BibitemOpen
  \bibfield  {author} {\bibinfo {author} {\bibfnamefont {V.~A.}\ \bibnamefont
  {Baulin}}, \bibinfo {author} {\bibfnamefont {A.}~\bibnamefont {Giacometti}},
  \bibinfo {author} {\bibfnamefont {D.~A.}\ \bibnamefont {Fedosov}}, \bibinfo
  {author} {\bibfnamefont {S.}~\bibnamefont {Ebbens}}, \bibinfo {author}
  {\bibfnamefont {N.~R.}\ \bibnamefont {Varela-Rosales}}, \bibinfo {author}
  {\bibfnamefont {N.}~\bibnamefont {Feliu}}, \bibinfo {author} {\bibfnamefont
  {M.}~\bibnamefont {Chowdhury}}, \bibinfo {author} {\bibfnamefont
  {M.}~\bibnamefont {Hu}}, \bibinfo {author} {\bibfnamefont {R.}~\bibnamefont
  {Füchslin}}, \bibinfo {author} {\bibfnamefont {M.}~\bibnamefont {Dijkstra}},
  \bibinfo {author} {\bibfnamefont {M.}~\bibnamefont {Mussel}}, \bibinfo
  {author} {\bibfnamefont {R.}~\bibnamefont {Van~Roij}}, \bibinfo {author}
  {\bibfnamefont {D.}~\bibnamefont {Xie}}, \bibinfo {author} {\bibfnamefont
  {V.}~\bibnamefont {Tzanov}}, \bibinfo {author} {\bibfnamefont
  {M.}~\bibnamefont {Zu}}, \bibinfo {author} {\bibfnamefont {S.}~\bibnamefont
  {Hidalgo-Caballero}}, \bibinfo {author} {\bibfnamefont {Y.}~\bibnamefont
  {Yuan}}, \bibinfo {author} {\bibfnamefont {L.}~\bibnamefont {Cocconi}},
  \bibinfo {author} {\bibfnamefont {C.-M.}\ \bibnamefont {Ghim}}, \bibinfo
  {author} {\bibfnamefont {C.}~\bibnamefont {Cottin-Bizonne}}, \bibinfo
  {author} {\bibfnamefont {M.~C.}\ \bibnamefont {Miguel}}, \bibinfo {author}
  {\bibfnamefont {M.~J.}\ \bibnamefont {Esplandiu}}, \bibinfo {author}
  {\bibfnamefont {J.}~\bibnamefont {Simmchen}}, \bibinfo {author}
  {\bibfnamefont {W.~J.}\ \bibnamefont {Parak}}, \bibinfo {author}
  {\bibfnamefont {M.}~\bibnamefont {Werner}}, \bibinfo {author} {\bibfnamefont
  {G.}~\bibnamefont {Gompper}},\ and\ \bibinfo {author} {\bibfnamefont {M.~M.}\
  \bibnamefont {Hanczyc}},\ }\bibfield  {title} {{\selectlanguage
  {English}\bibinfo {title} {Intelligent soft matter: towards embodied
  intelligence}},\ }\bibfield  {journal} {\bibinfo  {journal} {Soft Matter}\
  }\href {https://doi.org/10.1039/D5SM00174A} {10.1039/D5SM00174A} (\bibinfo
  {year} {2025}{\natexlab{a}})\BibitemShut {NoStop}%
\bibitem [{\citenamefont
  {Korsakova-Kreyn}(2021)}]{korsakova-kreyn_emotion_2021}%
  \BibitemOpen
  \bibfield  {author} {\bibinfo {author} {\bibfnamefont {M.}~\bibnamefont
  {Korsakova-Kreyn}},\ }\bibfield  {title} {{\selectlanguage {English}\bibinfo
  {title} {Emotion, embodied cognition, and {Artificial} {Intelligence}}},\
  }\bibfield  {journal} {\bibinfo  {journal} {Academia Letters}\ }\href
  {https://doi.org/10.20935/AL2883} {10.20935/AL2883} (\bibinfo {year}
  {2021})\BibitemShut {NoStop}%
\bibitem [{\citenamefont {Prescott}\ \emph {et~al.}(2024)\citenamefont
  {Prescott}, \citenamefont {Vogeley},\ and\ \citenamefont
  {Wykowska}}]{prescott_understanding_2024}%
  \BibitemOpen
  \bibfield  {author} {\bibinfo {author} {\bibfnamefont {T.~J.}\ \bibnamefont
  {Prescott}}, \bibinfo {author} {\bibfnamefont {K.}~\bibnamefont {Vogeley}},\
  and\ \bibinfo {author} {\bibfnamefont {A.}~\bibnamefont {Wykowska}},\
  }\bibfield  {title} {{\selectlanguage {English}\bibinfo {title}
  {Understanding the sense of self through robotics}},\ }\href
  {https://doi.org/10.1126/scirobotics.adn2733} {\bibfield  {journal} {\bibinfo
   {journal} {Sci. Robot.}\ }\textbf {\bibinfo {volume} {9}},\ \bibinfo {pages}
  {eadn2733} (\bibinfo {year} {2024})}\BibitemShut {NoStop}%
\bibitem [{\citenamefont {Kluger}\ \emph {et~al.}(2024)\citenamefont {Kluger},
  \citenamefont {Allen},\ and\ \citenamefont {Gross}}]{kluger_brainbody_2024}%
  \BibitemOpen
  \bibfield  {author} {\bibinfo {author} {\bibfnamefont {D.~S.}\ \bibnamefont
  {Kluger}}, \bibinfo {author} {\bibfnamefont {M.~G.}\ \bibnamefont {Allen}},\
  and\ \bibinfo {author} {\bibfnamefont {J.}~\bibnamefont {Gross}},\ }\bibfield
   {title} {{\selectlanguage {English}\bibinfo {title} {Brain–body states
  embody complex temporal dynamics}},\ }\href
  {https://doi.org/10.1016/j.tics.2024.05.003} {\bibfield  {journal} {\bibinfo
  {journal} {Trends in Cognitive Sciences}\ }\textbf {\bibinfo {volume} {28}},\
  \bibinfo {pages} {695} (\bibinfo {year} {2024})}\BibitemShut {NoStop}%
\bibitem [{\citenamefont {Stern}\ \emph {et~al.}(2022)\citenamefont {Stern},
  \citenamefont {Ben-Yehuda}, \citenamefont {Koren}, \citenamefont {Zaidel},\
  and\ \citenamefont {Salomon}}]{stern_dynamic_2022}%
  \BibitemOpen
  \bibfield  {author} {\bibinfo {author} {\bibfnamefont {Y.}~\bibnamefont
  {Stern}}, \bibinfo {author} {\bibfnamefont {I.}~\bibnamefont {Ben-Yehuda}},
  \bibinfo {author} {\bibfnamefont {D.}~\bibnamefont {Koren}}, \bibinfo
  {author} {\bibfnamefont {A.}~\bibnamefont {Zaidel}},\ and\ \bibinfo {author}
  {\bibfnamefont {R.}~\bibnamefont {Salomon}},\ }\bibfield  {title}
  {{\selectlanguage {English}\bibinfo {title} {The dynamic boundaries of the
  {Self}: {Serial} dependence in the {Sense} of {Agency}}},\ }\href
  {https://doi.org/10.1016/j.cortex.2022.03.015} {\bibfield  {journal}
  {\bibinfo  {journal} {Cortex}\ }\textbf {\bibinfo {volume} {152}},\ \bibinfo
  {pages} {109} (\bibinfo {year} {2022})}\BibitemShut {NoStop}%
\bibitem [{\citenamefont {Sloan}\ \emph {et~al.}(2023)\citenamefont {Sloan},
  \citenamefont {Jones},\ and\ \citenamefont {Kelso}}]{sloan_meaning_2023}%
  \BibitemOpen
  \bibfield  {author} {\bibinfo {author} {\bibfnamefont {A.~T.}\ \bibnamefont
  {Sloan}}, \bibinfo {author} {\bibfnamefont {N.~A.}\ \bibnamefont {Jones}},\
  and\ \bibinfo {author} {\bibfnamefont {J.~A.~S.}\ \bibnamefont {Kelso}},\
  }\bibfield  {title} {{\selectlanguage {English}\bibinfo {title} {Meaning from
  movement and stillness: {Signatures} of coordination dynamics reveal infant
  agency}},\ }\href {https://doi.org/10.1073/pnas.2306732120} {\bibfield
  {journal} {\bibinfo  {journal} {Proc. Natl. Acad. Sci. U.S.A.}\ }\textbf
  {\bibinfo {volume} {120}},\ \bibinfo {pages} {e2306732120} (\bibinfo {year}
  {2023})}\BibitemShut {NoStop}%
\bibitem [{\citenamefont {Beckmann}\ \emph {et~al.}(2023)\citenamefont
  {Beckmann}, \citenamefont {Köstner},\ and\ \citenamefont
  {Hipólito}}]{beckmann_alternative_2023}%
  \BibitemOpen
  \bibfield  {author} {\bibinfo {author} {\bibfnamefont {P.}~\bibnamefont
  {Beckmann}}, \bibinfo {author} {\bibfnamefont {G.}~\bibnamefont {Köstner}},\
  and\ \bibinfo {author} {\bibfnamefont {I.}~\bibnamefont {Hipólito}},\
  }\bibfield  {title} {{\selectlanguage {English}\bibinfo {title} {An
  {Alternative} to {Cognitivism}: {Computational} {Phenomenology} for {Deep}
  {Learning}}},\ }\href {https://doi.org/10.1007/s11023-023-09638-w} {\bibfield
   {journal} {\bibinfo  {journal} {Minds \& Machines}\ }\textbf {\bibinfo
  {volume} {33}},\ \bibinfo {pages} {397} (\bibinfo {year} {2023})}\BibitemShut
  {NoStop}%
\bibitem [{\citenamefont {King}\ and\ \citenamefont
  {Wyart}(2021)}]{king_human_2021}%
  \BibitemOpen
  \bibfield  {author} {\bibinfo {author} {\bibfnamefont {J.-R.}\ \bibnamefont
  {King}}\ and\ \bibinfo {author} {\bibfnamefont {V.}~\bibnamefont {Wyart}},\
  }\bibfield  {title} {{\selectlanguage {English}\bibinfo {title} {The {Human}
  {Brain} {Encodes} a {Chronicle} of {Visual} {Events} at {Each} {Instant} of
  {Time} {Through} the {Multiplexing} of {Traveling} {Waves}}},\ }\href
  {https://doi.org/10.1523/JNEUROSCI.2098-20.2021} {\bibfield  {journal}
  {\bibinfo  {journal} {J. Neurosci.}\ }\textbf {\bibinfo {volume} {41}},\
  \bibinfo {pages} {7224} (\bibinfo {year} {2021})}\BibitemShut {NoStop}%
\bibitem [{\citenamefont {Khona}\ and\ \citenamefont
  {Fiete}(2022)}]{khona_attractor_2022}%
  \BibitemOpen
  \bibfield  {author} {\bibinfo {author} {\bibfnamefont {M.}~\bibnamefont
  {Khona}}\ and\ \bibinfo {author} {\bibfnamefont {I.~R.}\ \bibnamefont
  {Fiete}},\ }\bibfield  {title} {{\selectlanguage {English}\bibinfo {title}
  {Attractor and integrator networks in the brain}},\ }\href
  {https://doi.org/10.1038/s41583-022-00642-0} {\bibfield  {journal} {\bibinfo
  {journal} {Nat Rev Neurosci}\ }\textbf {\bibinfo {volume} {23}},\ \bibinfo
  {pages} {744} (\bibinfo {year} {2022})},\ \bibinfo {note} {publisher: Nature
  Publishing Group}\BibitemShut {NoStop}%
\bibitem [{\citenamefont {Singer}(2021)}]{singer_recurrent_2021}%
  \BibitemOpen
  \bibfield  {author} {\bibinfo {author} {\bibfnamefont {W.}~\bibnamefont
  {Singer}},\ }\bibfield  {title} {{\selectlanguage {English}\bibinfo {title}
  {Recurrent dynamics in the cerebral cortex: {Integration} of sensory evidence
  with stored knowledge}},\ }\href {https://doi.org/10.1073/pnas.2101043118}
  {\bibfield  {journal} {\bibinfo  {journal} {Proc. Natl. Acad. Sci. U.S.A.}\
  }\textbf {\bibinfo {volume} {118}},\ \bibinfo {pages} {e2101043118} (\bibinfo
  {year} {2021})}\BibitemShut {NoStop}%
\bibitem [{\citenamefont {Baulin}\ \emph
  {et~al.}(2025{\natexlab{b}})\citenamefont {Baulin}, \citenamefont {Cook},
  \citenamefont {Friedman}, \citenamefont {Lumiruusu}, \citenamefont {Pashea},
  \citenamefont {Rahman},\ and\ \citenamefont
  {Waldeck}}]{baulin_discovery_2025}%
  \BibitemOpen
  \bibfield  {author} {\bibinfo {author} {\bibfnamefont {V.}~\bibnamefont
  {Baulin}}, \bibinfo {author} {\bibfnamefont {A.}~\bibnamefont {Cook}},
  \bibinfo {author} {\bibfnamefont {D.}~\bibnamefont {Friedman}}, \bibinfo
  {author} {\bibfnamefont {J.}~\bibnamefont {Lumiruusu}}, \bibinfo {author}
  {\bibfnamefont {A.}~\bibnamefont {Pashea}}, \bibinfo {author} {\bibfnamefont
  {S.}~\bibnamefont {Rahman}},\ and\ \bibinfo {author} {\bibfnamefont
  {B.}~\bibnamefont {Waldeck}},\ }\href
  {https://doi.org/10.48550/arXiv.2505.17500} {\bibinfo {title} {The
  {Discovery} {Engine}: {A} {Framework} for {AI}-{Driven} {Synthesis} and
  {Navigation} of {Scientific} {Knowledge} {Landscapes}}} (\bibinfo {year}
  {2025}{\natexlab{b}}),\ \bibinfo {note} {arXiv:2505.17500
  [cond-mat]}\BibitemShut {NoStop}%
\bibitem [{\citenamefont {Bello}\ and\ \citenamefont
  {Bridewell}(2017)}]{bello_there_2017}%
  \BibitemOpen
  \bibfield  {author} {\bibinfo {author} {\bibfnamefont {P.~F.}\ \bibnamefont
  {Bello}}\ and\ \bibinfo {author} {\bibfnamefont {W.}~\bibnamefont
  {Bridewell}},\ }\bibfield  {title} {{\selectlanguage {English}\bibinfo
  {title} {There {Is} {No} {Agency} {Without} {Attention}}},\ }\href
  {https://doi.org/10.1609/aimag.v38i4.2742} {\bibfield  {journal} {\bibinfo
  {journal} {AI Magazine}\ }\textbf {\bibinfo {volume} {38}},\ \bibinfo {pages}
  {27} (\bibinfo {year} {2017})},\ \bibinfo {note} {\_eprint:
  https://onlinelibrary.wiley.com/doi/pdf/10.1609/aimag.v38i4.2742}\BibitemShut
  {NoStop}%
\bibitem [{\citenamefont {Russell}\ \emph {et~al.}(2010)\citenamefont
  {Russell}, \citenamefont {Norvig},\ and\ \citenamefont
  {Davis}}]{russell_artificial_2010}%
  \BibitemOpen
  \bibfield  {author} {\bibinfo {author} {\bibfnamefont {S.~J.}\ \bibnamefont
  {Russell}}, \bibinfo {author} {\bibfnamefont {P.}~\bibnamefont {Norvig}},\
  and\ \bibinfo {author} {\bibfnamefont {E.}~\bibnamefont {Davis}},\
  }\href@noop {} {{\selectlanguage {English}\emph {\bibinfo {title} {Artificial
  intelligence: a modern approach}}}},\ \bibinfo {edition} {3rd}\ ed.,\
  Prentice {Hall} series in artificial intelligence\ (\bibinfo  {publisher}
  {Prentice Hall},\ \bibinfo {address} {Upper Saddle River},\ \bibinfo {year}
  {2010})\BibitemShut {NoStop}%
\bibitem [{\citenamefont {Kenton}\ \emph {et~al.}(2022)\citenamefont {Kenton},
  \citenamefont {Kumar}, \citenamefont {Farquhar}, \citenamefont {Richens},
  \citenamefont {MacDermott},\ and\ \citenamefont
  {Everitt}}]{kenton_discovering_2022}%
  \BibitemOpen
  \bibfield  {author} {\bibinfo {author} {\bibfnamefont {Z.}~\bibnamefont
  {Kenton}}, \bibinfo {author} {\bibfnamefont {R.}~\bibnamefont {Kumar}},
  \bibinfo {author} {\bibfnamefont {S.}~\bibnamefont {Farquhar}}, \bibinfo
  {author} {\bibfnamefont {J.}~\bibnamefont {Richens}}, \bibinfo {author}
  {\bibfnamefont {M.}~\bibnamefont {MacDermott}},\ and\ \bibinfo {author}
  {\bibfnamefont {T.}~\bibnamefont {Everitt}},\ }\href
  {https://arxiv.org/abs/2208.08345v2} {{\selectlanguage {English}\bibinfo
  {title} {Discovering {Agents}}}} (\bibinfo {year} {2022})\BibitemShut
  {NoStop}%
\bibitem [{\citenamefont {Serenko}\ and\ \citenamefont
  {Detlor}(2004)}]{serenko_intelligent_2004}%
  \BibitemOpen
  \bibfield  {author} {\bibinfo {author} {\bibfnamefont {A.}~\bibnamefont
  {Serenko}}\ and\ \bibinfo {author} {\bibfnamefont {B.}~\bibnamefont
  {Detlor}},\ }\bibfield  {title} {{\selectlanguage {English}\bibinfo {title}
  {Intelligent agents as innovations}},\ }\href
  {https://doi.org/10.1007/s00146-004-0310-5} {\bibfield  {journal} {\bibinfo
  {journal} {AI \& Soc}\ }\textbf {\bibinfo {volume} {18}},\ \bibinfo {pages}
  {364} (\bibinfo {year} {2004})}\BibitemShut {NoStop}%
\bibitem [{\citenamefont {Srinivasa}\ and\ \citenamefont
  {Deshmukh}(2020)}]{srinivasa_paradigms_2020}%
  \BibitemOpen
  \bibfield  {author} {\bibinfo {author} {\bibfnamefont {S.}~\bibnamefont
  {Srinivasa}}\ and\ \bibinfo {author} {\bibfnamefont {J.}~\bibnamefont
  {Deshmukh}},\ }\href {https://doi.org/10.4018/978-1-7998-2975-1.ch001}
  {\bibinfo {title} {Paradigms of {Computational} {Agency}}} (\bibinfo {year}
  {2020})\BibitemShut {NoStop}%
\bibitem [{\citenamefont {Biehl}(2017)}]{biehl_formal_2017}%
  \BibitemOpen
  \bibfield  {author} {\bibinfo {author} {\bibfnamefont {M.}~\bibnamefont
  {Biehl}},\ }\href {https://doi.org/10.48550/arXiv.1704.02716} {\bibinfo
  {title} {Formal approaches to a definition of agents}} (\bibinfo {year}
  {2017}),\ \bibinfo {note} {arXiv:1704.02716 [cs]}\BibitemShut {NoStop}%
\bibitem [{\citenamefont {Seifert}\ \emph {et~al.}(2024)\citenamefont
  {Seifert}, \citenamefont {Sealander}, \citenamefont {Marzen},\ and\
  \citenamefont {Levin}}]{seifert_reinforcement_2024}%
  \BibitemOpen
  \bibfield  {author} {\bibinfo {author} {\bibfnamefont {G.}~\bibnamefont
  {Seifert}}, \bibinfo {author} {\bibfnamefont {A.}~\bibnamefont {Sealander}},
  \bibinfo {author} {\bibfnamefont {S.}~\bibnamefont {Marzen}},\ and\ \bibinfo
  {author} {\bibfnamefont {M.}~\bibnamefont {Levin}},\ }\bibfield  {title}
  {\bibinfo {title} {From reinforcement learning to agency: {Frameworks} for
  understanding basal cognition},\ }\href
  {https://doi.org/10.1016/j.biosystems.2023.105107} {\bibfield  {journal}
  {\bibinfo  {journal} {BioSystems}\ }\textbf {\bibinfo {volume} {235}},\
  \bibinfo {pages} {105107} (\bibinfo {year} {2024})}\BibitemShut {NoStop}%
\bibitem [{\citenamefont {Jacob}\ \emph {et~al.}(2023)\citenamefont {Jacob},
  \citenamefont {Gupta},\ and\ \citenamefont {Andreas}}]{jacob_modeling_2023}%
  \BibitemOpen
  \bibfield  {author} {\bibinfo {author} {\bibfnamefont {A.~P.}\ \bibnamefont
  {Jacob}}, \bibinfo {author} {\bibfnamefont {A.}~\bibnamefont {Gupta}},\ and\
  \bibinfo {author} {\bibfnamefont {J.}~\bibnamefont {Andreas}},\ }\href
  {https://doi.org/10.48550/ARXIV.2312.04030} {\bibinfo {title} {Modeling
  {Boundedly} {Rational} {Agents} with {Latent} {Inference} {Budgets}}}
  (\bibinfo {year} {2023}),\ \bibinfo {note} {version Number: 1}\BibitemShut
  {NoStop}%
\bibitem [{\citenamefont {{Ahmed Abdel-Fattah, Tarek R. Besold, Helmar Gust,
  Ulf Krumnack, Martin Schmidt, Kai-Uwe
  Kuhnberger}}(2012)}]{ahmed_abdel-fattah_tarek_r_besold_helmar_gust_ulf_krumnack_martin_schmidt_kai-uwe_kuhnberger_rationality-guided_2012}%
  \BibitemOpen
  \bibfield  {author} {\bibinfo {author} {\bibnamefont {{Ahmed Abdel-Fattah,
  Tarek R. Besold, Helmar Gust, Ulf Krumnack, Martin Schmidt, Kai-Uwe
  Kuhnberger}}},\ }\bibfield  {title} {\bibinfo {title} {Rationality-{Guided}
  {AGI} as {Cognitive} {Systems}},\ }\href
  {https://escholarship.org/content/qt7xd1940g/qt7xd1940g.pdf} {\bibfield
  {journal} {\bibinfo  {journal} {Proceedings of the Annual Meeting of the
  Cognitive Science Society}\ }\textbf {\bibinfo {volume} {34}} (\bibinfo
  {year} {2012})}\BibitemShut {NoStop}%
\bibitem [{\citenamefont {Raja}\ \emph {et~al.}(2021)\citenamefont {Raja},
  \citenamefont {Valluri}, \citenamefont {Baggs}, \citenamefont {Chemero},\
  and\ \citenamefont {Anderson}}]{raja_markov_2021}%
  \BibitemOpen
  \bibfield  {author} {\bibinfo {author} {\bibfnamefont {V.}~\bibnamefont
  {Raja}}, \bibinfo {author} {\bibfnamefont {D.}~\bibnamefont {Valluri}},
  \bibinfo {author} {\bibfnamefont {E.}~\bibnamefont {Baggs}}, \bibinfo
  {author} {\bibfnamefont {A.}~\bibnamefont {Chemero}},\ and\ \bibinfo {author}
  {\bibfnamefont {M.~L.}\ \bibnamefont {Anderson}},\ }\bibfield  {title}
  {{\selectlanguage {English}\bibinfo {title} {The {Markov} blanket trick: {On}
  the scope of the free energy principle and active inference}},\ }\href
  {https://doi.org/10.1016/j.plrev.2021.09.001} {\bibfield  {journal} {\bibinfo
   {journal} {Physics of Life Reviews}\ }\textbf {\bibinfo {volume} {39}},\
  \bibinfo {pages} {49} (\bibinfo {year} {2021})}\BibitemShut {NoStop}%
\bibitem [{\citenamefont {Ha}\ and\ \citenamefont
  {Schmidhuber}(2018)}]{ha_world_2018}%
  \BibitemOpen
  \bibfield  {author} {\bibinfo {author} {\bibfnamefont {D.}~\bibnamefont
  {Ha}}\ and\ \bibinfo {author} {\bibfnamefont {J.}~\bibnamefont
  {Schmidhuber}},\ }\href {https://doi.org/10.5281/zenodo.1207631} {\bibinfo
  {title} {World {Models}}} (\bibinfo {year} {2018}),\ \bibinfo {note}
  {arXiv:1803.10122 [cs]}\BibitemShut {NoStop}%
\bibitem [{\citenamefont {Liu}\ \emph {et~al.}(2020)\citenamefont {Liu},
  \citenamefont {Gu},\ and\ \citenamefont {Liu}}]{liu_reinforcement_2020}%
  \BibitemOpen
  \bibfield  {author} {\bibinfo {author} {\bibfnamefont {J.}~\bibnamefont
  {Liu}}, \bibinfo {author} {\bibfnamefont {X.}~\bibnamefont {Gu}},\ and\
  \bibinfo {author} {\bibfnamefont {S.}~\bibnamefont {Liu}},\ }\href
  {https://doi.org/10.48550/arXiv.1908.11494} {\bibinfo {title} {Reinforcement
  learning with world model}} (\bibinfo {year} {2020}),\ \bibinfo {note}
  {arXiv:1908.11494 [cs] version: 4}\BibitemShut {NoStop}%
\bibitem [{\citenamefont {Dutta}\ \emph {et~al.}(2024)\citenamefont {Dutta},
  \citenamefont {Singh}, \citenamefont {Chakrabarti},\ and\ \citenamefont
  {Chakraborty}}]{dutta_how_2024}%
  \BibitemOpen
  \bibfield  {author} {\bibinfo {author} {\bibfnamefont {S.}~\bibnamefont
  {Dutta}}, \bibinfo {author} {\bibfnamefont {J.}~\bibnamefont {Singh}},
  \bibinfo {author} {\bibfnamefont {S.}~\bibnamefont {Chakrabarti}},\ and\
  \bibinfo {author} {\bibfnamefont {T.}~\bibnamefont {Chakraborty}},\ }\href
  {https://doi.org/10.48550/ARXIV.2402.18312} {\bibinfo {title} {How to think
  step-by-step: {A} mechanistic understanding of chain-of-thought reasoning}}
  (\bibinfo {year} {2024}),\ \bibinfo {note} {version Number: 2}\BibitemShut
  {NoStop}%
\bibitem [{\citenamefont {Chung}\ \emph {et~al.}(2023)\citenamefont {Chung},
  \citenamefont {Anokhin},\ and\ \citenamefont {Krueger}}]{chung_thinker_2023}%
  \BibitemOpen
  \bibfield  {author} {\bibinfo {author} {\bibfnamefont {S.}~\bibnamefont
  {Chung}}, \bibinfo {author} {\bibfnamefont {I.}~\bibnamefont {Anokhin}},\
  and\ \bibinfo {author} {\bibfnamefont {D.}~\bibnamefont {Krueger}},\ }\href
  {https://doi.org/10.48550/arXiv.2307.14993} {\bibinfo {title} {Thinker:
  {Learning} to {Plan} and {Act}}} (\bibinfo {year} {2023}),\ \bibinfo {note}
  {arXiv:2307.14993 [cs]}\BibitemShut {NoStop}%
\bibitem [{\citenamefont {Brückner}\ and\ \citenamefont
  {Tkačik}(2023)}]{bruckner_information_2023}%
  \BibitemOpen
  \bibfield  {author} {\bibinfo {author} {\bibfnamefont {D.~B.}\ \bibnamefont
  {Brückner}}\ and\ \bibinfo {author} {\bibfnamefont {G.}~\bibnamefont
  {Tkačik}},\ }\href {https://doi.org/10.48550/ARXIV.2312.05895} {\bibinfo
  {title} {Information content and optimization of self-organized developmental
  systems}} (\bibinfo {year} {2023}),\ \bibinfo {note} {version Number:
  2}\BibitemShut {NoStop}%
\bibitem [{\citenamefont {Beck}\ and\ \citenamefont
  {Moore}(2024)}]{beck_how_2024}%
  \BibitemOpen
  \bibfield  {author} {\bibinfo {author} {\bibfnamefont {M.}~\bibnamefont
  {Beck}}\ and\ \bibinfo {author} {\bibfnamefont {T.}~\bibnamefont {Moore}},\
  }\href {https://doi.org/10.48550/arXiv.2409.05264} {\bibinfo {title} {How
  {We} {Lost} {The} {Internet}}} (\bibinfo {year} {2024}),\ \bibinfo {note}
  {arXiv:2409.05264 [cs]}\BibitemShut {NoStop}%
\bibitem [{\citenamefont {Hoffman}\ \emph {et~al.}(2013)\citenamefont
  {Hoffman}, \citenamefont {Blei}, \citenamefont {Wang},\ and\ \citenamefont
  {Paisley}}]{hoffman_stochastic_nodate}%
  \BibitemOpen
  \bibfield  {author} {\bibinfo {author} {\bibfnamefont {M.}~\bibnamefont
  {Hoffman}}, \bibinfo {author} {\bibfnamefont {D.~M.}\ \bibnamefont {Blei}},
  \bibinfo {author} {\bibfnamefont {C.}~\bibnamefont {Wang}},\ and\ \bibinfo
  {author} {\bibfnamefont {J.}~\bibnamefont {Paisley}},\ }\href
  {https://doi.org/10.48550/arXiv.1206.7051} {\bibinfo {title} {Stochastic
  {Variational} {Inference}}} (\bibinfo {year} {2013}),\ \bibinfo {note}
  {arXiv:1206.7051 [stat]}\BibitemShut {NoStop}%
\bibitem [{\citenamefont {Liu}\ and\ \citenamefont
  {Wang}(2019)}]{liu_stein_2019}%
  \BibitemOpen
  \bibfield  {author} {\bibinfo {author} {\bibfnamefont {Q.}~\bibnamefont
  {Liu}}\ and\ \bibinfo {author} {\bibfnamefont {D.}~\bibnamefont {Wang}},\
  }\href {https://doi.org/10.48550/arXiv.1608.04471} {\bibinfo {title} {Stein
  {Variational} {Gradient} {Descent}: {A} {General} {Purpose} {Bayesian}
  {Inference} {Algorithm}}} (\bibinfo {year} {2019}),\ \bibinfo {note}
  {arXiv:1608.04471 [stat]}\BibitemShut {NoStop}%
\bibitem [{\citenamefont {Walter}(2024)}]{walter_artificial_2024}%
  \BibitemOpen
  \bibfield  {author} {\bibinfo {author} {\bibfnamefont {Y.}~\bibnamefont
  {Walter}},\ }\bibfield  {title} {{\selectlanguage {English}\bibinfo {title}
  {Artificial influencers and the dead internet theory}},\ }\href
  {https://doi.org/10.1007/s00146-023-01857-0} {\bibfield  {journal} {\bibinfo
  {journal} {AI \& Soc}\ ,\ \bibinfo {pages} {1}} (\bibinfo {year} {2024})},\
  \bibinfo {note} {company: Springer Distributor: Springer Institution:
  Springer Label: Springer Publisher: Springer London}\BibitemShut {NoStop}%
\bibitem [{\citenamefont {Robine}\ \emph {et~al.}(2023)\citenamefont {Robine},
  \citenamefont {Höftmann}, \citenamefont {Uelwer},\ and\ \citenamefont
  {Harmeling}}]{robine_transformer-based_2023}%
  \BibitemOpen
  \bibfield  {author} {\bibinfo {author} {\bibfnamefont {J.}~\bibnamefont
  {Robine}}, \bibinfo {author} {\bibfnamefont {M.}~\bibnamefont {Höftmann}},
  \bibinfo {author} {\bibfnamefont {T.}~\bibnamefont {Uelwer}},\ and\ \bibinfo
  {author} {\bibfnamefont {S.}~\bibnamefont {Harmeling}},\ }\href@noop {}
  {{\selectlanguage {English}\bibinfo {title} {Transformer-{Based} {World}
  {Models} {Are} {Happy} with 100k {Interactions}}}} (\bibinfo {year}
  {2023})\BibitemShut {NoStop}%
\bibitem [{\citenamefont {Xie}\ \emph {et~al.}(2025)\citenamefont {Xie},
  \citenamefont {Yang}, \citenamefont {Gunerli},\ and\ \citenamefont
  {Riedl}}]{xie_making_2025}%
  \BibitemOpen
  \bibfield  {author} {\bibinfo {author} {\bibfnamefont {K.}~\bibnamefont
  {Xie}}, \bibinfo {author} {\bibfnamefont {I.}~\bibnamefont {Yang}}, \bibinfo
  {author} {\bibfnamefont {J.}~\bibnamefont {Gunerli}},\ and\ \bibinfo {author}
  {\bibfnamefont {M.}~\bibnamefont {Riedl}},\ }\bibfield  {title} {\bibinfo
  {title} {Making {Large} {Language} {Models} into {World} {Models} with
  {Precondition} and {Effect} {Knowledge}},\ }in\ \href
  {https://aclanthology.org/2025.coling-main.503/} {\emph {\bibinfo {booktitle}
  {Proceedings of the 31st {International} {Conference} on {Computational}
  {Linguistics}}}},\ \bibinfo {editor} {edited by\ \bibinfo {editor}
  {\bibfnamefont {O.}~\bibnamefont {Rambow}}, \bibinfo {editor} {\bibfnamefont
  {L.}~\bibnamefont {Wanner}}, \bibinfo {editor} {\bibfnamefont
  {M.}~\bibnamefont {Apidianaki}}, \bibinfo {editor} {\bibfnamefont
  {H.}~\bibnamefont {Al-Khalifa}}, \bibinfo {editor} {\bibfnamefont {B.~D.}\
  \bibnamefont {Eugenio}},\ and\ \bibinfo {editor} {\bibfnamefont
  {S.}~\bibnamefont {Schockaert}}}\ (\bibinfo  {publisher} {Association for
  Computational Linguistics},\ \bibinfo {address} {Abu Dhabi, UAE},\ \bibinfo
  {year} {2025})\ pp.\ \bibinfo {pages} {7532--7545}\BibitemShut {NoStop}%
\bibitem [{\citenamefont {Pan}\ \emph {et~al.}(2023)\citenamefont {Pan},
  \citenamefont {Gao}, \citenamefont {Chen},\ and\ \citenamefont
  {Chen}}]{pan_what_2023}%
  \BibitemOpen
  \bibfield  {author} {\bibinfo {author} {\bibfnamefont {J.}~\bibnamefont
  {Pan}}, \bibinfo {author} {\bibfnamefont {T.}~\bibnamefont {Gao}}, \bibinfo
  {author} {\bibfnamefont {H.}~\bibnamefont {Chen}},\ and\ \bibinfo {author}
  {\bibfnamefont {D.}~\bibnamefont {Chen}},\ }\href
  {https://doi.org/10.48550/ARXIV.2305.09731} {\bibinfo {title} {What
  {In}-{Context} {Learning} "{Learns}" {In}-{Context}: {Disentangling} {Task}
  {Recognition} and {Task} {Learning}}} (\bibinfo {year} {2023}),\ \bibinfo
  {note} {version Number: 1}\BibitemShut {NoStop}%
\bibitem [{\citenamefont {Taniguchi}\ \emph {et~al.}(2023)\citenamefont
  {Taniguchi}, \citenamefont {Murata}, \citenamefont {Suzuki}, \citenamefont
  {Ognibene}, \citenamefont {Lanillos}, \citenamefont {Ugur}, \citenamefont
  {Jamone}, \citenamefont {Nakamura}, \citenamefont {Ciria}, \citenamefont
  {Lara},\ and\ \citenamefont {Pezzulo}}]{taniguchi_world_2023}%
  \BibitemOpen
  \bibfield  {author} {\bibinfo {author} {\bibfnamefont {T.}~\bibnamefont
  {Taniguchi}}, \bibinfo {author} {\bibfnamefont {S.}~\bibnamefont {Murata}},
  \bibinfo {author} {\bibfnamefont {M.}~\bibnamefont {Suzuki}}, \bibinfo
  {author} {\bibfnamefont {D.}~\bibnamefont {Ognibene}}, \bibinfo {author}
  {\bibfnamefont {P.}~\bibnamefont {Lanillos}}, \bibinfo {author}
  {\bibfnamefont {E.}~\bibnamefont {Ugur}}, \bibinfo {author} {\bibfnamefont
  {L.}~\bibnamefont {Jamone}}, \bibinfo {author} {\bibfnamefont
  {T.}~\bibnamefont {Nakamura}}, \bibinfo {author} {\bibfnamefont
  {A.}~\bibnamefont {Ciria}}, \bibinfo {author} {\bibfnamefont
  {B.}~\bibnamefont {Lara}},\ and\ \bibinfo {author} {\bibfnamefont
  {G.}~\bibnamefont {Pezzulo}},\ }\href
  {https://doi.org/10.48550/arXiv.2301.05832} {\bibinfo {title} {World {Models}
  and {Predictive} {Coding} for {Cognitive} and {Developmental} {Robotics}:
  {Frontiers} and {Challenges}}} (\bibinfo {year} {2023}),\ \bibinfo {note}
  {arXiv:2301.05832 [cs]}\BibitemShut {NoStop}%
\bibitem [{\citenamefont {Chan}\ \emph
  {et~al.}(2023{\natexlab{a}})\citenamefont {Chan}, \citenamefont {Salganik},
  \citenamefont {Markelius}, \citenamefont {Pang}, \citenamefont {Rajkumar},
  \citenamefont {Krasheninnikov}, \citenamefont {Langosco}, \citenamefont {He},
  \citenamefont {Duan}, \citenamefont {Carroll}, \citenamefont {Lin},
  \citenamefont {Mayhew}, \citenamefont {Collins}, \citenamefont
  {Molamohammadi}, \citenamefont {Burden}, \citenamefont {Zhao}, \citenamefont
  {Rismani}, \citenamefont {Voudouris}, \citenamefont {Bhatt}, \citenamefont
  {Weller}, \citenamefont {Krueger},\ and\ \citenamefont
  {Maharaj}}]{chan_harms_2023}%
  \BibitemOpen
  \bibfield  {author} {\bibinfo {author} {\bibfnamefont {A.}~\bibnamefont
  {Chan}}, \bibinfo {author} {\bibfnamefont {R.}~\bibnamefont {Salganik}},
  \bibinfo {author} {\bibfnamefont {A.}~\bibnamefont {Markelius}}, \bibinfo
  {author} {\bibfnamefont {C.}~\bibnamefont {Pang}}, \bibinfo {author}
  {\bibfnamefont {N.}~\bibnamefont {Rajkumar}}, \bibinfo {author}
  {\bibfnamefont {D.}~\bibnamefont {Krasheninnikov}}, \bibinfo {author}
  {\bibfnamefont {L.}~\bibnamefont {Langosco}}, \bibinfo {author}
  {\bibfnamefont {Z.}~\bibnamefont {He}}, \bibinfo {author} {\bibfnamefont
  {Y.}~\bibnamefont {Duan}}, \bibinfo {author} {\bibfnamefont {M.}~\bibnamefont
  {Carroll}}, \bibinfo {author} {\bibfnamefont {M.}~\bibnamefont {Lin}},
  \bibinfo {author} {\bibfnamefont {A.}~\bibnamefont {Mayhew}}, \bibinfo
  {author} {\bibfnamefont {K.}~\bibnamefont {Collins}}, \bibinfo {author}
  {\bibfnamefont {M.}~\bibnamefont {Molamohammadi}}, \bibinfo {author}
  {\bibfnamefont {J.}~\bibnamefont {Burden}}, \bibinfo {author} {\bibfnamefont
  {W.}~\bibnamefont {Zhao}}, \bibinfo {author} {\bibfnamefont {S.}~\bibnamefont
  {Rismani}}, \bibinfo {author} {\bibfnamefont {K.}~\bibnamefont {Voudouris}},
  \bibinfo {author} {\bibfnamefont {U.}~\bibnamefont {Bhatt}}, \bibinfo
  {author} {\bibfnamefont {A.}~\bibnamefont {Weller}}, \bibinfo {author}
  {\bibfnamefont {D.}~\bibnamefont {Krueger}},\ and\ \bibinfo {author}
  {\bibfnamefont {T.}~\bibnamefont {Maharaj}},\ }\bibfield  {title} {\bibinfo
  {title} {Harms from {Increasingly} {Agentic} {Algorithmic} {Systems}},\ }in\
  \href {https://doi.org/10.1145/3593013.3594033} {\emph {\bibinfo {booktitle}
  {2023 {ACM} {Conference} on {Fairness}, {Accountability}, and
  {Transparency}}}}\ (\bibinfo {year} {2023})\ pp.\ \bibinfo {pages}
  {651--666},\ \bibinfo {note} {arXiv:2302.10329 [cs]}\BibitemShut {NoStop}%
\bibitem [{\citenamefont {Chan}\ \emph
  {et~al.}(2023{\natexlab{b}})\citenamefont {Chan}, \citenamefont {Salganik},
  \citenamefont {Markelius}, \citenamefont {Pang}, \citenamefont {Rajkumar},
  \citenamefont {Krasheninnikov}, \citenamefont {Langosco}, \citenamefont {He},
  \citenamefont {Duan}, \citenamefont {Carroll}, \citenamefont {Lin},
  \citenamefont {Mayhew}, \citenamefont {Collins}, \citenamefont
  {Molamohammadi}, \citenamefont {Burden}, \citenamefont {Zhao}, \citenamefont
  {Rismani}, \citenamefont {Voudouris}, \citenamefont {Bhatt}, \citenamefont
  {Weller}, \citenamefont {Krueger},\ and\ \citenamefont
  {Maharaj}}]{chan_harms_2023-1}%
  \BibitemOpen
  \bibfield  {author} {\bibinfo {author} {\bibfnamefont {A.}~\bibnamefont
  {Chan}}, \bibinfo {author} {\bibfnamefont {R.}~\bibnamefont {Salganik}},
  \bibinfo {author} {\bibfnamefont {A.}~\bibnamefont {Markelius}}, \bibinfo
  {author} {\bibfnamefont {C.}~\bibnamefont {Pang}}, \bibinfo {author}
  {\bibfnamefont {N.}~\bibnamefont {Rajkumar}}, \bibinfo {author}
  {\bibfnamefont {D.}~\bibnamefont {Krasheninnikov}}, \bibinfo {author}
  {\bibfnamefont {L.}~\bibnamefont {Langosco}}, \bibinfo {author}
  {\bibfnamefont {Z.}~\bibnamefont {He}}, \bibinfo {author} {\bibfnamefont
  {Y.}~\bibnamefont {Duan}}, \bibinfo {author} {\bibfnamefont {M.}~\bibnamefont
  {Carroll}}, \bibinfo {author} {\bibfnamefont {M.}~\bibnamefont {Lin}},
  \bibinfo {author} {\bibfnamefont {A.}~\bibnamefont {Mayhew}}, \bibinfo
  {author} {\bibfnamefont {K.}~\bibnamefont {Collins}}, \bibinfo {author}
  {\bibfnamefont {M.}~\bibnamefont {Molamohammadi}}, \bibinfo {author}
  {\bibfnamefont {J.}~\bibnamefont {Burden}}, \bibinfo {author} {\bibfnamefont
  {W.}~\bibnamefont {Zhao}}, \bibinfo {author} {\bibfnamefont {S.}~\bibnamefont
  {Rismani}}, \bibinfo {author} {\bibfnamefont {K.}~\bibnamefont {Voudouris}},
  \bibinfo {author} {\bibfnamefont {U.}~\bibnamefont {Bhatt}}, \bibinfo
  {author} {\bibfnamefont {A.}~\bibnamefont {Weller}}, \bibinfo {author}
  {\bibfnamefont {D.}~\bibnamefont {Krueger}},\ and\ \bibinfo {author}
  {\bibfnamefont {T.}~\bibnamefont {Maharaj}},\ }\bibfield  {title} {\bibinfo
  {title} {Harms from {Increasingly} {Agentic} {Algorithmic} {Systems}},\ }in\
  \href {https://doi.org/10.1145/3593013.3594033} {\emph {\bibinfo {booktitle}
  {2023 {ACM} {Conference} on {Fairness}, {Accountability}, and
  {Transparency}}}}\ (\bibinfo {year} {2023})\ pp.\ \bibinfo {pages}
  {651--666},\ \bibinfo {note} {arXiv:2302.10329 [cs]}\BibitemShut {NoStop}%
\bibitem [{\citenamefont {Mukherjee}\ and\ \citenamefont
  {Chang}(2025)}]{mukherjee_agentic_2025}%
  \BibitemOpen
  \bibfield  {author} {\bibinfo {author} {\bibfnamefont {A.}~\bibnamefont
  {Mukherjee}}\ and\ \bibinfo {author} {\bibfnamefont {H.}~\bibnamefont
  {Chang}},\ }\href {https://papers.ssrn.com/abstract=5123621}
  {{\selectlanguage {English}\bibinfo {title} {Agentic {AI}: {Autonomy},
  {Accountability}, and the {Algorithmic} {Society}}}} (\bibinfo {year}
  {2025})\BibitemShut {NoStop}%
\bibitem [{\citenamefont {{Iason Gabriel et.
  al}}(2024)}]{iason_gabriel_et_al_ethics_2024}%
  \BibitemOpen
  \bibfield  {author} {\bibinfo {author} {\bibnamefont {{Iason Gabriel et.
  al}}},\ }\href
  {https://storage.googleapis.com/deepmind-media/DeepMind.com/Blog/ethics-of-advanced-ai-assistants/the-ethics-of-advanced-ai-assistants-2024-i.pdf}
  {\bibinfo {title} {The {Ethics} of {Advanced} {AI} {Assistants}}} (\bibinfo
  {year} {2024})\BibitemShut {NoStop}%
\bibitem [{\citenamefont {Park}\ \emph {et~al.}(2024)\citenamefont {Park},
  \citenamefont {Goldstein}, \citenamefont {O’Gara}, \citenamefont {Chen},\
  and\ \citenamefont {Hendrycks}}]{park_ai_2024}%
  \BibitemOpen
  \bibfield  {author} {\bibinfo {author} {\bibfnamefont {P.~S.}\ \bibnamefont
  {Park}}, \bibinfo {author} {\bibfnamefont {S.}~\bibnamefont {Goldstein}},
  \bibinfo {author} {\bibfnamefont {A.}~\bibnamefont {O’Gara}}, \bibinfo
  {author} {\bibfnamefont {M.}~\bibnamefont {Chen}},\ and\ \bibinfo {author}
  {\bibfnamefont {D.}~\bibnamefont {Hendrycks}},\ }\bibfield  {title}
  {{\selectlanguage {English}\bibinfo {title} {{AI} deception: {A} survey of
  examples, risks, and potential solutions}},\ }\href
  {https://doi.org/10.1016/j.patter.2024.100988} {\bibfield  {journal}
  {\bibinfo  {journal} {Patterns}\ }\textbf {\bibinfo {volume} {5}},\ \bibinfo
  {pages} {100988} (\bibinfo {year} {2024})}\BibitemShut {NoStop}%
\bibitem [{\citenamefont {Shavit}\ \emph {et~al.}(2024)\citenamefont {Shavit},
  \citenamefont {O’Keefe}, \citenamefont {Eloundou}, \citenamefont
  {McMillan}, \citenamefont {Agarwal}, \citenamefont {Brundage}, \citenamefont
  {Adler}, \citenamefont {Campbell}, \citenamefont {Lee}, \citenamefont
  {Mishkin}, \citenamefont {Hickey}, \citenamefont {Slama}, \citenamefont
  {Ahmad}, \citenamefont {Beutel}, \citenamefont {Passos},\ and\ \citenamefont
  {Robinson}}]{shavit_practices_nodate}%
  \BibitemOpen
  \bibfield  {author} {\bibinfo {author} {\bibfnamefont {Y.}~\bibnamefont
  {Shavit}}, \bibinfo {author} {\bibfnamefont {C.}~\bibnamefont {O’Keefe}},
  \bibinfo {author} {\bibfnamefont {T.}~\bibnamefont {Eloundou}}, \bibinfo
  {author} {\bibfnamefont {P.}~\bibnamefont {McMillan}}, \bibinfo {author}
  {\bibfnamefont {S.}~\bibnamefont {Agarwal}}, \bibinfo {author} {\bibfnamefont
  {M.}~\bibnamefont {Brundage}}, \bibinfo {author} {\bibfnamefont
  {S.}~\bibnamefont {Adler}}, \bibinfo {author} {\bibfnamefont
  {R.}~\bibnamefont {Campbell}}, \bibinfo {author} {\bibfnamefont
  {T.}~\bibnamefont {Lee}}, \bibinfo {author} {\bibfnamefont {P.}~\bibnamefont
  {Mishkin}}, \bibinfo {author} {\bibfnamefont {A.}~\bibnamefont {Hickey}},
  \bibinfo {author} {\bibfnamefont {K.}~\bibnamefont {Slama}}, \bibinfo
  {author} {\bibfnamefont {L.}~\bibnamefont {Ahmad}}, \bibinfo {author}
  {\bibfnamefont {A.}~\bibnamefont {Beutel}}, \bibinfo {author} {\bibfnamefont
  {A.}~\bibnamefont {Passos}},\ and\ \bibinfo {author} {\bibfnamefont {D.~G.}\
  \bibnamefont {Robinson}},\ }\href@noop {} {{\selectlanguage {English}\bibinfo
  {title} {Practices for {Governing} {Agentic} {AI} {Systems}}}} (\bibinfo
  {year} {2024})\BibitemShut {NoStop}%
\bibitem [{\citenamefont {Shanahan}\ \emph {et~al.}(2023)\citenamefont
  {Shanahan}, \citenamefont {McDonell},\ and\ \citenamefont
  {Reynolds}}]{shanahan_role_2023}%
  \BibitemOpen
  \bibfield  {author} {\bibinfo {author} {\bibfnamefont {M.}~\bibnamefont
  {Shanahan}}, \bibinfo {author} {\bibfnamefont {K.}~\bibnamefont {McDonell}},\
  and\ \bibinfo {author} {\bibfnamefont {L.}~\bibnamefont {Reynolds}},\
  }\bibfield  {title} {{\selectlanguage {English}\bibinfo {title} {Role play
  with large language models}},\ }\href
  {https://doi.org/10.1038/s41586-023-06647-8} {\bibfield  {journal} {\bibinfo
  {journal} {Nature}\ }\textbf {\bibinfo {volume} {623}},\ \bibinfo {pages}
  {493} (\bibinfo {year} {2023})}\BibitemShut {NoStop}%
\bibitem [{\citenamefont {Rolls}(2023)}]{rolls_brain_2023}%
  \BibitemOpen
  \bibfield  {author} {\bibinfo {author} {\bibfnamefont {E.~T.}\ \bibnamefont
  {Rolls}},\ }\href {https://doi.org/10.1093/oso/9780198887911.001.0001}
  {{\selectlanguage {English}\emph {\bibinfo {title} {Brain {Computations} and
  {Connectivity}}}}},\ \bibinfo {edition} {2nd}\ ed.\ (\bibinfo  {publisher}
  {Oxford University PressOxford},\ \bibinfo {year} {2023})\BibitemShut
  {NoStop}%
\bibitem [{\citenamefont {Fréal}\ \emph {et~al.}(2023)\citenamefont {Fréal},
  \citenamefont {Jamann}, \citenamefont {Ten~Bos}, \citenamefont {Jansen},
  \citenamefont {Petersen}, \citenamefont {Ligthart}, \citenamefont
  {Hoogenraad},\ and\ \citenamefont {Kole}}]{freal_sodium_2023}%
  \BibitemOpen
  \bibfield  {author} {\bibinfo {author} {\bibfnamefont {A.}~\bibnamefont
  {Fréal}}, \bibinfo {author} {\bibfnamefont {N.}~\bibnamefont {Jamann}},
  \bibinfo {author} {\bibfnamefont {J.}~\bibnamefont {Ten~Bos}}, \bibinfo
  {author} {\bibfnamefont {J.}~\bibnamefont {Jansen}}, \bibinfo {author}
  {\bibfnamefont {N.}~\bibnamefont {Petersen}}, \bibinfo {author}
  {\bibfnamefont {T.}~\bibnamefont {Ligthart}}, \bibinfo {author}
  {\bibfnamefont {C.~C.}\ \bibnamefont {Hoogenraad}},\ and\ \bibinfo {author}
  {\bibfnamefont {M.~H.}\ \bibnamefont {Kole}},\ }\bibfield  {title}
  {{\selectlanguage {English}\bibinfo {title} {Sodium channel endocytosis
  drives axon initial segment plasticity}},\ }\href
  {https://doi.org/10.1126/sciadv.adf3885} {\bibfield  {journal} {\bibinfo
  {journal} {Sci. Adv.}\ }\textbf {\bibinfo {volume} {9}},\ \bibinfo {pages}
  {eadf3885} (\bibinfo {year} {2023})}\BibitemShut {NoStop}%
\bibitem [{\citenamefont {Lewis}\ and\ \citenamefont
  {Sarkadi}(2024)}]{lewis_reflective_2024}%
  \BibitemOpen
  \bibfield  {author} {\bibinfo {author} {\bibfnamefont {P.~R.}\ \bibnamefont
  {Lewis}}\ and\ \bibinfo {author} {\bibfnamefont {S.}~\bibnamefont
  {Sarkadi}},\ }\bibfield  {title} {\bibinfo {title} {Reflective {Artificial}
  {Intelligence}},\ }\href {https://doi.org/10.1007/s11023-024-09664-2}
  {\bibfield  {journal} {\bibinfo  {journal} {Minds \& Machines}\ }\textbf
  {\bibinfo {volume} {34}},\ \bibinfo {pages} {14} (\bibinfo {year} {2024})},\
  \bibinfo {note} {arXiv:2301.10823 [cs]}\BibitemShut {NoStop}%
\bibitem [{\citenamefont {Comay}\ \emph {et~al.}(2024)\citenamefont {Comay},
  \citenamefont {Solovey},\ and\ \citenamefont
  {Barttfeld}}]{comay_metacognition_2024}%
  \BibitemOpen
  \bibfield  {author} {\bibinfo {author} {\bibfnamefont {N.~A.}\ \bibnamefont
  {Comay}}, \bibinfo {author} {\bibfnamefont {G.}~\bibnamefont {Solovey}},\
  and\ \bibinfo {author} {\bibfnamefont {P.}~\bibnamefont {Barttfeld}},\ }\href
  {https://doi.org/10.31234/osf.io/byjv6} {\bibinfo {title} {Metacognition in
  multialternative choices is based on more information than type-1
  decisions.}} (\bibinfo {year} {2024})\BibitemShut {NoStop}%
\bibitem [{\citenamefont {{Chris Fields, Donald D. Hoffman, Chetan Prakash,
  Robert
  Prentner}}(2017)}]{chris_fields_donald_d_hoffman_chetan_prakash_robert_prentner_eigenforms_2017}%
  \BibitemOpen
  \bibfield  {author} {\bibinfo {author} {\bibnamefont {{Chris Fields, Donald
  D. Hoffman, Chetan Prakash, Robert Prentner}}},\ }\bibfield  {title}
  {\bibinfo {title} {Eigenforms, {Interfaces} and {Holographic} {Encoding}:
  {Toward} an {Evolutionary} {Account} of {Objects} and {Spacetime}},\ }\href
  {https://constructivist.info/12/3/265} {\bibfield  {journal} {\bibinfo
  {journal} {Constructivist Foundations}\ }\textbf {\bibinfo {volume} {12}},\
  \bibinfo {pages} {265} (\bibinfo {year} {2017})}\BibitemShut {NoStop}%
\bibitem [{\citenamefont {Rabiza}(2022)}]{rabiza_point_2022}%
  \BibitemOpen
  \bibfield  {author} {\bibinfo {author} {\bibfnamefont {M.}~\bibnamefont
  {Rabiza}},\ }\bibfield  {title} {{\selectlanguage {English}\bibinfo {title}
  {Point and {Network} {Notions} of {Artificial} {Intelligence} {Agency}}},\
  }in\ \href {https://doi.org/10.3390/proceedings2022081018} {{\selectlanguage
  {English}\emph {\bibinfo {booktitle} {The 2021 {Summit} of the
  {International} {Society} for the {Study} of {Information}}}}}\ (\bibinfo
  {publisher} {MDPI},\ \bibinfo {year} {2022})\ p.~\bibinfo {pages}
  {18}\BibitemShut {NoStop}%
\bibitem [{\citenamefont {Gregor}\ and\ \citenamefont
  {Besse}(2020)}]{gregor_self-organizing_2020}%
  \BibitemOpen
  \bibfield  {author} {\bibinfo {author} {\bibfnamefont {K.}~\bibnamefont
  {Gregor}}\ and\ \bibinfo {author} {\bibfnamefont {F.}~\bibnamefont {Besse}},\
  }\bibfield  {title} {{\selectlanguage {English}\bibinfo {title}
  {Self-{Organizing} {Intelligent} {Matter}: {A} blueprint for an {AI}
  generating algorithm}}\ }(\bibinfo {year} {2020})\BibitemShut {NoStop}%
\bibitem [{\citenamefont {Theraulaz}\ \emph {et~al.}(2002)\citenamefont
  {Theraulaz}, \citenamefont {Bonabeau}, \citenamefont {Nicolis}, \citenamefont
  {Solé}, \citenamefont {Fourcassié}, \citenamefont {Blanco}, \citenamefont
  {Fournier}, \citenamefont {Joly}, \citenamefont {Fernández}, \citenamefont
  {Grimal}, \citenamefont {Dalle},\ and\ \citenamefont
  {Deneubourg}}]{theraulaz_spatial_2002}%
  \BibitemOpen
  \bibfield  {author} {\bibinfo {author} {\bibfnamefont {G.}~\bibnamefont
  {Theraulaz}}, \bibinfo {author} {\bibfnamefont {E.}~\bibnamefont {Bonabeau}},
  \bibinfo {author} {\bibfnamefont {S.~C.}\ \bibnamefont {Nicolis}}, \bibinfo
  {author} {\bibfnamefont {R.~V.}\ \bibnamefont {Solé}}, \bibinfo {author}
  {\bibfnamefont {V.}~\bibnamefont {Fourcassié}}, \bibinfo {author}
  {\bibfnamefont {S.}~\bibnamefont {Blanco}}, \bibinfo {author} {\bibfnamefont
  {R.}~\bibnamefont {Fournier}}, \bibinfo {author} {\bibfnamefont {J.-L.}\
  \bibnamefont {Joly}}, \bibinfo {author} {\bibfnamefont {P.}~\bibnamefont
  {Fernández}}, \bibinfo {author} {\bibfnamefont {A.}~\bibnamefont {Grimal}},
  \bibinfo {author} {\bibfnamefont {P.}~\bibnamefont {Dalle}},\ and\ \bibinfo
  {author} {\bibfnamefont {J.-L.}\ \bibnamefont {Deneubourg}},\ }\bibfield
  {title} {\bibinfo {title} {Spatial patterns in ant colonies},\ }\href
  {https://doi.org/10.1073/pnas.152302199} {\bibfield  {journal} {\bibinfo
  {journal} {Proceedings of the National Academy of Sciences}\ }\textbf
  {\bibinfo {volume} {99}},\ \bibinfo {pages} {9645} (\bibinfo {year}
  {2002})},\ \bibinfo {note} {publisher: Proceedings of the National Academy of
  Sciences}\BibitemShut {NoStop}%
\bibitem [{\citenamefont {Couzin}(2009)}]{couzin_collective_2009}%
  \BibitemOpen
  \bibfield  {author} {\bibinfo {author} {\bibfnamefont {I.~D.}\ \bibnamefont
  {Couzin}},\ }\bibfield  {title} {\bibinfo {title} {Collective cognition in
  animal groups},\ }\href {https://doi.org/10.1016/j.tics.2008.10.002}
  {\bibfield  {journal} {\bibinfo  {journal} {Trends in Cognitive Sciences}\
  }\textbf {\bibinfo {volume} {13}},\ \bibinfo {pages} {36} (\bibinfo {year}
  {2009})}\BibitemShut {NoStop}%
\bibitem [{\citenamefont {Múgica}\ \emph {et~al.}(2022)\citenamefont
  {Múgica}, \citenamefont {Torrents}, \citenamefont {Cristín}, \citenamefont
  {Puy}, \citenamefont {Miguel},\ and\ \citenamefont
  {Pastor-Satorras}}]{mugica_scale-free_2022}%
  \BibitemOpen
  \bibfield  {author} {\bibinfo {author} {\bibfnamefont {J.}~\bibnamefont
  {Múgica}}, \bibinfo {author} {\bibfnamefont {J.}~\bibnamefont {Torrents}},
  \bibinfo {author} {\bibfnamefont {J.}~\bibnamefont {Cristín}}, \bibinfo
  {author} {\bibfnamefont {A.}~\bibnamefont {Puy}}, \bibinfo {author}
  {\bibfnamefont {M.~C.}\ \bibnamefont {Miguel}},\ and\ \bibinfo {author}
  {\bibfnamefont {R.}~\bibnamefont {Pastor-Satorras}},\ }\bibfield  {title}
  {{\selectlanguage {English}\bibinfo {title} {Scale-free behavioral cascades
  and effective leadership in schooling fish}},\ }\href
  {https://doi.org/10.1038/s41598-022-14337-0} {\bibfield  {journal} {\bibinfo
  {journal} {Sci Rep}\ }\textbf {\bibinfo {volume} {12}},\ \bibinfo {pages}
  {10783} (\bibinfo {year} {2022})}\BibitemShut {NoStop}%
\bibitem [{\citenamefont {Puy}\ \emph {et~al.}(2024{\natexlab{a}})\citenamefont
  {Puy}, \citenamefont {Gimeno}, \citenamefont {Beltran}, \citenamefont
  {Dolado}, \citenamefont {Miguel}, \citenamefont {Ioannou},\ and\
  \citenamefont {Pastor-Satorras}}]{puy_perceived_2024}%
  \BibitemOpen
  \bibfield  {author} {\bibinfo {author} {\bibfnamefont {A.}~\bibnamefont
  {Puy}}, \bibinfo {author} {\bibfnamefont {E.}~\bibnamefont {Gimeno}},
  \bibinfo {author} {\bibfnamefont {F.~S.}\ \bibnamefont {Beltran}}, \bibinfo
  {author} {\bibfnamefont {R.}~\bibnamefont {Dolado}}, \bibinfo {author}
  {\bibfnamefont {M.~C.}\ \bibnamefont {Miguel}}, \bibinfo {author}
  {\bibfnamefont {C.~C.}\ \bibnamefont {Ioannou}},\ and\ \bibinfo {author}
  {\bibfnamefont {R.}~\bibnamefont {Pastor-Satorras}},\ }\href
  {https://doi.org/10.48550/arXiv.2410.09264} {\bibinfo {title} {Perceived risk
  determines spatial position in fish shoals through altered rules of
  interaction}} (\bibinfo {year} {2024}{\natexlab{a}}),\ \bibinfo {note}
  {arXiv:2410.09264 [physics]}\BibitemShut {NoStop}%
\bibitem [{\citenamefont {Puy}\ \emph {et~al.}(2024{\natexlab{b}})\citenamefont
  {Puy}, \citenamefont {Gimeno}, \citenamefont {Torrents}, \citenamefont
  {Bartashevich}, \citenamefont {Miguel}, \citenamefont {Pastor-Satorras},\
  and\ \citenamefont {Romanczuk}}]{puy_selective_2024}%
  \BibitemOpen
  \bibfield  {author} {\bibinfo {author} {\bibfnamefont {A.}~\bibnamefont
  {Puy}}, \bibinfo {author} {\bibfnamefont {E.}~\bibnamefont {Gimeno}},
  \bibinfo {author} {\bibfnamefont {J.}~\bibnamefont {Torrents}}, \bibinfo
  {author} {\bibfnamefont {P.}~\bibnamefont {Bartashevich}}, \bibinfo {author}
  {\bibfnamefont {M.~C.}\ \bibnamefont {Miguel}}, \bibinfo {author}
  {\bibfnamefont {R.}~\bibnamefont {Pastor-Satorras}},\ and\ \bibinfo {author}
  {\bibfnamefont {P.}~\bibnamefont {Romanczuk}},\ }\bibfield  {title}
  {{\selectlanguage {English}\bibinfo {title} {Selective social interactions
  and speed-induced leadership in schooling fish}},\ }\href
  {https://doi.org/10.1073/pnas.2309733121} {\bibfield  {journal} {\bibinfo
  {journal} {Proc. Natl. Acad. Sci. U.S.A.}\ }\textbf {\bibinfo {volume}
  {121}},\ \bibinfo {pages} {e2309733121} (\bibinfo {year}
  {2024}{\natexlab{b}})}\BibitemShut {NoStop}%
\bibitem [{\citenamefont {Rubenstein}\ \emph {et~al.}(2014)\citenamefont
  {Rubenstein}, \citenamefont {Cornejo},\ and\ \citenamefont
  {Nagpal}}]{rubenstein_programmable_2014}%
  \BibitemOpen
  \bibfield  {author} {\bibinfo {author} {\bibfnamefont {M.}~\bibnamefont
  {Rubenstein}}, \bibinfo {author} {\bibfnamefont {A.}~\bibnamefont
  {Cornejo}},\ and\ \bibinfo {author} {\bibfnamefont {R.}~\bibnamefont
  {Nagpal}},\ }\bibfield  {title} {\bibinfo {title} {Programmable self-assembly
  in a thousand-robot swarm},\ }\href {https://doi.org/10.1126/science.1254295}
  {\bibfield  {journal} {\bibinfo  {journal} {Science}\ }\textbf {\bibinfo
  {volume} {345}},\ \bibinfo {pages} {795} (\bibinfo {year} {2014})},\ \bibinfo
  {note} {publisher: American Association for the Advancement of
  Science}\BibitemShut {NoStop}%
\bibitem [{\citenamefont {March-Pons}\ \emph {et~al.}(2024)\citenamefont
  {March-Pons}, \citenamefont {Múgica}, \citenamefont {Ferrero},\ and\
  \citenamefont {Miguel}}]{march-pons_honeybee-like_2024}%
  \BibitemOpen
  \bibfield  {author} {\bibinfo {author} {\bibfnamefont {D.}~\bibnamefont
  {March-Pons}}, \bibinfo {author} {\bibfnamefont {J.}~\bibnamefont {Múgica}},
  \bibinfo {author} {\bibfnamefont {E.~E.}\ \bibnamefont {Ferrero}},\ and\
  \bibinfo {author} {\bibfnamefont {M.~C.}\ \bibnamefont {Miguel}},\ }\bibfield
   {title} {\bibinfo {title} {Honeybee-like collective decision making in a
  kilobot swarm},\ }\href {https://doi.org/10.1103/PhysRevResearch.6.033149}
  {\bibfield  {journal} {\bibinfo  {journal} {Phys. Rev. Res.}\ }\textbf
  {\bibinfo {volume} {6}},\ \bibinfo {pages} {033149} (\bibinfo {year}
  {2024})},\ \bibinfo {note} {publisher: American Physical Society}\BibitemShut
  {NoStop}%
\bibitem [{\citenamefont {Wang}\ \emph {et~al.}(2024)\citenamefont {Wang},
  \citenamefont {Wang}, \citenamefont {Chen}, \citenamefont {Liu},
  \citenamefont {Wang}, \citenamefont {Yuan}, \citenamefont {Ma}, \citenamefont
  {Xu}, \citenamefont {Cheng}, \citenamefont {Ji}, \citenamefont {Yang},
  \citenamefont {Shuai}, \citenamefont {Ye}, \citenamefont {Wang},
  \citenamefont {Jiao},\ and\ \citenamefont {Liu}}]{wang_robo-matter_2024}%
  \BibitemOpen
  \bibfield  {author} {\bibinfo {author} {\bibfnamefont {J.}~\bibnamefont
  {Wang}}, \bibinfo {author} {\bibfnamefont {G.}~\bibnamefont {Wang}}, \bibinfo
  {author} {\bibfnamefont {H.}~\bibnamefont {Chen}}, \bibinfo {author}
  {\bibfnamefont {Y.}~\bibnamefont {Liu}}, \bibinfo {author} {\bibfnamefont
  {P.}~\bibnamefont {Wang}}, \bibinfo {author} {\bibfnamefont {D.}~\bibnamefont
  {Yuan}}, \bibinfo {author} {\bibfnamefont {X.}~\bibnamefont {Ma}}, \bibinfo
  {author} {\bibfnamefont {X.}~\bibnamefont {Xu}}, \bibinfo {author}
  {\bibfnamefont {Z.}~\bibnamefont {Cheng}}, \bibinfo {author} {\bibfnamefont
  {B.}~\bibnamefont {Ji}}, \bibinfo {author} {\bibfnamefont {M.}~\bibnamefont
  {Yang}}, \bibinfo {author} {\bibfnamefont {J.}~\bibnamefont {Shuai}},
  \bibinfo {author} {\bibfnamefont {F.}~\bibnamefont {Ye}}, \bibinfo {author}
  {\bibfnamefont {J.}~\bibnamefont {Wang}}, \bibinfo {author} {\bibfnamefont
  {Y.}~\bibnamefont {Jiao}},\ and\ \bibinfo {author} {\bibfnamefont
  {L.}~\bibnamefont {Liu}},\ }\bibfield  {title} {{\selectlanguage
  {English}\bibinfo {title} {Robo-{Matter} towards reconfigurable
  multifunctional smart materials}},\ }\href
  {https://doi.org/10.1038/s41467-024-53123-6} {\bibfield  {journal} {\bibinfo
  {journal} {Nat Commun}\ }\textbf {\bibinfo {volume} {15}},\ \bibinfo {pages}
  {8853} (\bibinfo {year} {2024})},\ \bibinfo {note} {publisher: Nature
  Publishing Group}\BibitemShut {NoStop}%
\bibitem [{\citenamefont {Hu}\ \emph {et~al.}(2018)\citenamefont {Hu},
  \citenamefont {Lum}, \citenamefont {Mastrangeli},\ and\ \citenamefont
  {Sitti}}]{hu_small-scale_2018}%
  \BibitemOpen
  \bibfield  {author} {\bibinfo {author} {\bibfnamefont {W.}~\bibnamefont
  {Hu}}, \bibinfo {author} {\bibfnamefont {G.~Z.}\ \bibnamefont {Lum}},
  \bibinfo {author} {\bibfnamefont {M.}~\bibnamefont {Mastrangeli}},\ and\
  \bibinfo {author} {\bibfnamefont {M.}~\bibnamefont {Sitti}},\ }\bibfield
  {title} {{\selectlanguage {English}\bibinfo {title} {Small-scale soft-bodied
  robot with multimodal locomotion}},\ }\href
  {https://doi.org/10.1038/nature25443} {\bibfield  {journal} {\bibinfo
  {journal} {Nature}\ }\textbf {\bibinfo {volume} {554}},\ \bibinfo {pages}
  {81} (\bibinfo {year} {2018})},\ \bibinfo {note} {publisher: Nature
  Publishing Group}\BibitemShut {NoStop}%
\bibitem [{\citenamefont {Ceylan}\ \emph {et~al.}(2017)\citenamefont {Ceylan},
  \citenamefont {Giltinan}, \citenamefont {Kozielski},\ and\ \citenamefont
  {Sitti}}]{ceylan_mobile_2017}%
  \BibitemOpen
  \bibfield  {author} {\bibinfo {author} {\bibfnamefont {H.}~\bibnamefont
  {Ceylan}}, \bibinfo {author} {\bibfnamefont {J.}~\bibnamefont {Giltinan}},
  \bibinfo {author} {\bibfnamefont {K.}~\bibnamefont {Kozielski}},\ and\
  \bibinfo {author} {\bibfnamefont {M.}~\bibnamefont {Sitti}},\ }\bibfield
  {title} {{\selectlanguage {English}\bibinfo {title} {Mobile microrobots for
  bioengineering applications}},\ }\href {https://doi.org/10.1039/C7LC00064B}
  {\bibfield  {journal} {\bibinfo  {journal} {Lab Chip}\ }\textbf {\bibinfo
  {volume} {17}},\ \bibinfo {pages} {1705} (\bibinfo {year} {2017})},\ \bibinfo
  {note} {publisher: The Royal Society of Chemistry}\BibitemShut {NoStop}%
\bibitem [{\citenamefont {Goh}\ \emph {et~al.}(2025)\citenamefont {Goh},
  \citenamefont {Westphal}, \citenamefont {Winkler},\ and\ \citenamefont
  {Gompper}}]{goh_alignment-induced_2025}%
  \BibitemOpen
  \bibfield  {author} {\bibinfo {author} {\bibfnamefont {S.}~\bibnamefont
  {Goh}}, \bibinfo {author} {\bibfnamefont {E.}~\bibnamefont {Westphal}},
  \bibinfo {author} {\bibfnamefont {R.~G.}\ \bibnamefont {Winkler}},\ and\
  \bibinfo {author} {\bibfnamefont {G.}~\bibnamefont {Gompper}},\ }\bibfield
  {title} {\bibinfo {title} {Alignment-induced self-organization of
  autonomously steering microswimmers: {Turbulence}, clusters, vortices, and
  jets},\ }\href {https://doi.org/10.1103/PhysRevResearch.7.013142} {\bibfield
  {journal} {\bibinfo  {journal} {Phys. Rev. Res.}\ }\textbf {\bibinfo {volume}
  {7}},\ \bibinfo {pages} {013142} (\bibinfo {year} {2025})},\ \bibinfo {note}
  {publisher: American Physical Society}\BibitemShut {NoStop}%
\bibitem [{\citenamefont {Goh}\ \emph {et~al.}(2022)\citenamefont {Goh},
  \citenamefont {Winkler},\ and\ \citenamefont {Gompper}}]{goh_noisy_2022}%
  \BibitemOpen
  \bibfield  {author} {\bibinfo {author} {\bibfnamefont {S.}~\bibnamefont
  {Goh}}, \bibinfo {author} {\bibfnamefont {R.~G.}\ \bibnamefont {Winkler}},\
  and\ \bibinfo {author} {\bibfnamefont {G.}~\bibnamefont {Gompper}},\
  }\bibfield  {title} {{\selectlanguage {English}\bibinfo {title} {Noisy
  pursuit and pattern formation of self-steering active particles}},\ }\href
  {https://doi.org/10.1088/1367-2630/ac924f} {\bibfield  {journal} {\bibinfo
  {journal} {New J. Phys.}\ }\textbf {\bibinfo {volume} {24}},\ \bibinfo
  {pages} {093039} (\bibinfo {year} {2022})},\ \bibinfo {note} {publisher: IOP
  Publishing}\BibitemShut {NoStop}%
\bibitem [{\citenamefont {Negi}\ \emph {et~al.}(2022)\citenamefont {Negi},
  \citenamefont {Winkler},\ and\ \citenamefont {Gompper}}]{negi_emergent_2022}%
  \BibitemOpen
  \bibfield  {author} {\bibinfo {author} {\bibfnamefont {R.~S.}\ \bibnamefont
  {Negi}}, \bibinfo {author} {\bibfnamefont {R.~G.}\ \bibnamefont {Winkler}},\
  and\ \bibinfo {author} {\bibfnamefont {G.}~\bibnamefont {Gompper}},\
  }\bibfield  {title} {{\selectlanguage {English}\bibinfo {title} {Emergent
  collective behavior of active {Brownian} particles with visual perception}},\
  }\href {https://doi.org/10.1039/D2SM00736C} {\bibfield  {journal} {\bibinfo
  {journal} {Soft Matter}\ }\textbf {\bibinfo {volume} {18}},\ \bibinfo {pages}
  {6167} (\bibinfo {year} {2022})}\BibitemShut {NoStop}%
\bibitem [{\citenamefont {Iyer}\ \emph {et~al.}(2024)\citenamefont {Iyer},
  \citenamefont {Negi}, \citenamefont {Schadschneider},\ and\ \citenamefont
  {Gompper}}]{iyer_directed_2024}%
  \BibitemOpen
  \bibfield  {author} {\bibinfo {author} {\bibfnamefont {P.}~\bibnamefont
  {Iyer}}, \bibinfo {author} {\bibfnamefont {R.~S.}\ \bibnamefont {Negi}},
  \bibinfo {author} {\bibfnamefont {A.}~\bibnamefont {Schadschneider}},\ and\
  \bibinfo {author} {\bibfnamefont {G.}~\bibnamefont {Gompper}},\ }\bibfield
  {title} {{\selectlanguage {English}\bibinfo {title} {Directed motion of
  cognitive active agents in a crowded three-way intersection}},\ }\href
  {https://doi.org/10.1038/s42005-024-01860-x} {\bibfield  {journal} {\bibinfo
  {journal} {Commun Phys}\ }\textbf {\bibinfo {volume} {7}},\ \bibinfo {pages}
  {1} (\bibinfo {year} {2024})},\ \bibinfo {note} {publisher: Nature Publishing
  Group}\BibitemShut {NoStop}%
\bibitem [{\citenamefont {Kriegman}\ \emph {et~al.}(2020)\citenamefont
  {Kriegman}, \citenamefont {Blackiston}, \citenamefont {Levin},\ and\
  \citenamefont {Bongard}}]{kriegman_scalable_2020}%
  \BibitemOpen
  \bibfield  {author} {\bibinfo {author} {\bibfnamefont {S.}~\bibnamefont
  {Kriegman}}, \bibinfo {author} {\bibfnamefont {D.}~\bibnamefont
  {Blackiston}}, \bibinfo {author} {\bibfnamefont {M.}~\bibnamefont {Levin}},\
  and\ \bibinfo {author} {\bibfnamefont {J.}~\bibnamefont {Bongard}},\
  }\bibfield  {title} {\bibinfo {title} {A scalable pipeline for designing
  reconfigurable organisms},\ }\href {https://doi.org/10.1073/pnas.1910837117}
  {\bibfield  {journal} {\bibinfo  {journal} {Proceedings of the National
  Academy of Sciences}\ }\textbf {\bibinfo {volume} {117}},\ \bibinfo {pages}
  {1853} (\bibinfo {year} {2020})},\ \bibinfo {note} {publisher: Proceedings of
  the National Academy of Sciences}\BibitemShut {NoStop}%
\bibitem [{\citenamefont {Gumuskaya}\ \emph {et~al.}(2024)\citenamefont
  {Gumuskaya}, \citenamefont {Srivastava}, \citenamefont {Cooper},
  \citenamefont {Lesser}, \citenamefont {Semegran}, \citenamefont {Garnier},\
  and\ \citenamefont {Levin}}]{gumuskaya_motile_2024}%
  \BibitemOpen
  \bibfield  {author} {\bibinfo {author} {\bibfnamefont {G.}~\bibnamefont
  {Gumuskaya}}, \bibinfo {author} {\bibfnamefont {P.}~\bibnamefont
  {Srivastava}}, \bibinfo {author} {\bibfnamefont {B.~G.}\ \bibnamefont
  {Cooper}}, \bibinfo {author} {\bibfnamefont {H.}~\bibnamefont {Lesser}},
  \bibinfo {author} {\bibfnamefont {B.}~\bibnamefont {Semegran}}, \bibinfo
  {author} {\bibfnamefont {S.}~\bibnamefont {Garnier}},\ and\ \bibinfo {author}
  {\bibfnamefont {M.}~\bibnamefont {Levin}},\ }\bibfield  {title}
  {{\selectlanguage {English}\bibinfo {title} {Motile {Living} {Biobots}
  {Self}‐{Construct} from {Adult} {Human} {Somatic} {Progenitor} {Seed}
  {Cells}}},\ }\href {https://doi.org/10.1002/advs.202303575} {\bibfield
  {journal} {\bibinfo  {journal} {Advanced Science}\ }\textbf {\bibinfo
  {volume} {11}},\ \bibinfo {pages} {2303575} (\bibinfo {year}
  {2024})}\BibitemShut {NoStop}%
\bibitem [{\citenamefont {Levin}(2024)}]{levin_multiscale_2024}%
  \BibitemOpen
  \bibfield  {author} {\bibinfo {author} {\bibfnamefont {M.}~\bibnamefont
  {Levin}},\ }\bibfield  {title} {{\selectlanguage {English}\bibinfo {title}
  {The {Multiscale} {Wisdom} of the {Body}: {Collective} {Intelligence} as a
  {Tractable} {Interface} for {Next}-{Generation} {Biomedicine}}},\ }\href
  {https://doi.org/10.1002/bies.202400196} {\bibfield  {journal} {\bibinfo
  {journal} {BioEssays}\ }\textbf {\bibinfo {volume} {47}},\ \bibinfo {pages}
  {e202400196} (\bibinfo {year} {2024})},\ \bibinfo {note} {\_eprint:
  https://onlinelibrary.wiley.com/doi/pdf/10.1002/bies.202400196}\BibitemShut
  {NoStop}%
\bibitem [{\citenamefont {Solé}\ \emph {et~al.}(2024)\citenamefont {Solé},
  \citenamefont {Conde–Pueyo}, \citenamefont {Pla–Mauri}, \citenamefont
  {Garcia–Ojalvo}, \citenamefont {Montserrat},\ and\ \citenamefont
  {Levin}}]{sole_open_2024}%
  \BibitemOpen
  \bibfield  {author} {\bibinfo {author} {\bibfnamefont {R.}~\bibnamefont
  {Solé}}, \bibinfo {author} {\bibfnamefont {N.}~\bibnamefont
  {Conde–Pueyo}}, \bibinfo {author} {\bibfnamefont {J.}~\bibnamefont
  {Pla–Mauri}}, \bibinfo {author} {\bibfnamefont {J.}~\bibnamefont
  {Garcia–Ojalvo}}, \bibinfo {author} {\bibfnamefont {N.}~\bibnamefont
  {Montserrat}},\ and\ \bibinfo {author} {\bibfnamefont {M.}~\bibnamefont
  {Levin}},\ }\bibfield  {title} {{\selectlanguage {English}\bibinfo {title}
  {Open problems in synthetic multicellularity}},\ }\href
  {https://doi.org/10.1038/s41540-024-00477-8} {\bibfield  {journal} {\bibinfo
  {journal} {npj Syst Biol Appl}\ }\textbf {\bibinfo {volume} {10}},\ \bibinfo
  {pages} {1} (\bibinfo {year} {2024})},\ \bibinfo {note} {publisher: Nature
  Publishing Group}\BibitemShut {NoStop}%
\bibitem [{\citenamefont {Schmickl}\ \emph {et~al.}(2016)\citenamefont
  {Schmickl}, \citenamefont {Stefanec},\ and\ \citenamefont
  {Crailsheim}}]{schmickl_how_2016}%
  \BibitemOpen
  \bibfield  {author} {\bibinfo {author} {\bibfnamefont {T.}~\bibnamefont
  {Schmickl}}, \bibinfo {author} {\bibfnamefont {M.}~\bibnamefont {Stefanec}},\
  and\ \bibinfo {author} {\bibfnamefont {K.}~\bibnamefont {Crailsheim}},\
  }\bibfield  {title} {{\selectlanguage {English}\bibinfo {title} {How a
  life-like system emerges from a simplistic particle motion law}},\ }\href
  {https://doi.org/10.1038/srep37969} {\bibfield  {journal} {\bibinfo
  {journal} {Sci Rep}\ }\textbf {\bibinfo {volume} {6}},\ \bibinfo {pages}
  {37969} (\bibinfo {year} {2016})},\ \bibinfo {note} {publisher: Nature
  Publishing Group}\BibitemShut {NoStop}%
\bibitem [{\citenamefont {Zhang}\ \emph {et~al.}(2025)\citenamefont {Zhang},
  \citenamefont {Goldstein},\ and\ \citenamefont
  {Levin}}]{zhang_classical_2025}%
  \BibitemOpen
  \bibfield  {author} {\bibinfo {author} {\bibfnamefont {T.}~\bibnamefont
  {Zhang}}, \bibinfo {author} {\bibfnamefont {A.}~\bibnamefont {Goldstein}},\
  and\ \bibinfo {author} {\bibfnamefont {M.}~\bibnamefont {Levin}},\ }\bibfield
   {title} {{\selectlanguage {English}\bibinfo {title} {Classical sorting
  algorithms as a model of morphogenesis: {Self}-sorting arrays reveal
  unexpected competencies in a minimal model of basal intelligence}},\ }\href
  {https://doi.org/10.1177/10597123241269740} {\bibfield  {journal} {\bibinfo
  {journal} {Adaptive Behavior}\ }\textbf {\bibinfo {volume} {33}},\ \bibinfo
  {pages} {25} (\bibinfo {year} {2025})},\ \bibinfo {note} {publisher: SAGE
  Publications Ltd STM}\BibitemShut {NoStop}%
\bibitem [{\citenamefont {Lehman}(2023)}]{lehman_machine_2023}%
  \BibitemOpen
  \bibfield  {author} {\bibinfo {author} {\bibfnamefont {J.}~\bibnamefont
  {Lehman}},\ }\href {https://doi.org/10.48550/ARXIV.2302.09248} {\bibinfo
  {title} {Machine {Love}}} (\bibinfo {year} {2023}),\ \bibinfo {note} {version
  Number: 2}\BibitemShut {NoStop}%
\bibitem [{\citenamefont {Çelikok}\ \emph {et~al.}(2019)\citenamefont
  {Çelikok}, \citenamefont {Peltola}, \citenamefont {Daee},\ and\
  \citenamefont {Kaski}}]{celikok_interactive_2019}%
  \BibitemOpen
  \bibfield  {author} {\bibinfo {author} {\bibfnamefont {M.~M.}\ \bibnamefont
  {Çelikok}}, \bibinfo {author} {\bibfnamefont {T.}~\bibnamefont {Peltola}},
  \bibinfo {author} {\bibfnamefont {P.}~\bibnamefont {Daee}},\ and\ \bibinfo
  {author} {\bibfnamefont {S.}~\bibnamefont {Kaski}},\ }\href
  {https://arxiv.org/pdf/1912.05284} {{\selectlanguage {English}\bibinfo
  {title} {Interactive {AI} with a {Theory} of {Mind}}}} (\bibinfo {year}
  {2019})\BibitemShut {NoStop}%
\bibitem [{\citenamefont {Capouskova}\ \emph {et~al.}(2023)\citenamefont
  {Capouskova}, \citenamefont {Zamora‐López}, \citenamefont {Kringelbach},\
  and\ \citenamefont {Deco}}]{capouskova_integration_2023}%
  \BibitemOpen
  \bibfield  {author} {\bibinfo {author} {\bibfnamefont {K.}~\bibnamefont
  {Capouskova}}, \bibinfo {author} {\bibfnamefont {G.}~\bibnamefont
  {Zamora‐López}}, \bibinfo {author} {\bibfnamefont {M.~L.}\ \bibnamefont
  {Kringelbach}},\ and\ \bibinfo {author} {\bibfnamefont {G.}~\bibnamefont
  {Deco}},\ }\bibfield  {title} {{\selectlanguage {English}\bibinfo {title}
  {Integration and segregation manifolds in the brain ensure cognitive
  flexibility during tasks and rest}},\ }\href
  {https://doi.org/10.1002/hbm.26511} {\bibfield  {journal} {\bibinfo
  {journal} {Human Brain Mapping}\ }\textbf {\bibinfo {volume} {44}},\ \bibinfo
  {pages} {6349} (\bibinfo {year} {2023})}\BibitemShut {NoStop}%
\bibitem [{\citenamefont {Galakhova}\ \emph {et~al.}(2022)\citenamefont
  {Galakhova}, \citenamefont {Hunt}, \citenamefont {Wilbers}, \citenamefont
  {Heyer}, \citenamefont {De~Kock}, \citenamefont {Mansvelder},\ and\
  \citenamefont {Goriounova}}]{galakhova_evolution_2022}%
  \BibitemOpen
  \bibfield  {author} {\bibinfo {author} {\bibfnamefont {A.}~\bibnamefont
  {Galakhova}}, \bibinfo {author} {\bibfnamefont {S.}~\bibnamefont {Hunt}},
  \bibinfo {author} {\bibfnamefont {R.}~\bibnamefont {Wilbers}}, \bibinfo
  {author} {\bibfnamefont {D.}~\bibnamefont {Heyer}}, \bibinfo {author}
  {\bibfnamefont {C.}~\bibnamefont {De~Kock}}, \bibinfo {author} {\bibfnamefont
  {H.}~\bibnamefont {Mansvelder}},\ and\ \bibinfo {author} {\bibfnamefont
  {N.}~\bibnamefont {Goriounova}},\ }\bibfield  {title} {{\selectlanguage
  {English}\bibinfo {title} {Evolution of cortical neurons supporting human
  cognition}},\ }\href {https://doi.org/10.1016/j.tics.2022.08.012} {\bibfield
  {journal} {\bibinfo  {journal} {Trends in Cognitive Sciences}\ }\textbf
  {\bibinfo {volume} {26}},\ \bibinfo {pages} {909} (\bibinfo {year}
  {2022})}\BibitemShut {NoStop}%
\bibitem [{\citenamefont {Song}\ \emph {et~al.}(2022)\citenamefont {Song},
  \citenamefont {Shim},\ and\ \citenamefont
  {Rosenberg}}]{song_large-scale_2022}%
  \BibitemOpen
  \bibfield  {author} {\bibinfo {author} {\bibfnamefont {H.}~\bibnamefont
  {Song}}, \bibinfo {author} {\bibfnamefont {W.~M.}\ \bibnamefont {Shim}},\
  and\ \bibinfo {author} {\bibfnamefont {M.~D.}\ \bibnamefont {Rosenberg}},\
  }\href {https://doi.org/10.1101/2022.11.05.515307} {{\selectlanguage
  {English}\bibinfo {title} {Large-scale neural dynamics in a shared
  low-dimensional state space reflect cognitive and attentional dynamics}}}
  (\bibinfo {year} {2022})\BibitemShut {NoStop}%
\bibitem [{\citenamefont {Zhang}\ \emph {et~al.}(2024)\citenamefont {Zhang},
  \citenamefont {Nolte}, \citenamefont {Sadhukhan}, \citenamefont {Chen},\ and\
  \citenamefont {Bottou}}]{zhang_memory_2024}%
  \BibitemOpen
  \bibfield  {author} {\bibinfo {author} {\bibfnamefont {J.}~\bibnamefont
  {Zhang}}, \bibinfo {author} {\bibfnamefont {N.}~\bibnamefont {Nolte}},
  \bibinfo {author} {\bibfnamefont {R.}~\bibnamefont {Sadhukhan}}, \bibinfo
  {author} {\bibfnamefont {B.}~\bibnamefont {Chen}},\ and\ \bibinfo {author}
  {\bibfnamefont {L.}~\bibnamefont {Bottou}},\ }\href
  {https://doi.org/10.48550/ARXIV.2405.06394} {\bibinfo {title} {Memory
  {Mosaics}}} (\bibinfo {year} {2024}),\ \bibinfo {note} {version Number:
  2}\BibitemShut {NoStop}%
\bibitem [{\citenamefont {Cullen}(2023)}]{cullen_internal_2023}%
  \BibitemOpen
  \bibfield  {author} {\bibinfo {author} {\bibfnamefont {K.~E.}\ \bibnamefont
  {Cullen}},\ }\bibfield  {title} {{\selectlanguage {English}\bibinfo {title}
  {Internal models of self-motion: neural computations by the vestibular
  cerebellum}},\ }\href {https://doi.org/10.1016/j.tins.2023.08.009} {\bibfield
   {journal} {\bibinfo  {journal} {Trends in Neurosciences}\ }\textbf {\bibinfo
  {volume} {46}},\ \bibinfo {pages} {986} (\bibinfo {year} {2023})}\BibitemShut
  {NoStop}%
\bibitem [{\citenamefont {Thornton}\ and\ \citenamefont
  {Tamir}(2024)}]{thornton_neural_2024}%
  \BibitemOpen
  \bibfield  {author} {\bibinfo {author} {\bibfnamefont {M.~A.}\ \bibnamefont
  {Thornton}}\ and\ \bibinfo {author} {\bibfnamefont {D.~I.}\ \bibnamefont
  {Tamir}},\ }\bibfield  {title} {{\selectlanguage {English}\bibinfo {title}
  {Neural representations of situations and mental states are composed of sums
  of representations of the actions they afford}},\ }\href
  {https://doi.org/10.1038/s41467-024-44870-7} {\bibfield  {journal} {\bibinfo
  {journal} {Nat Commun}\ }\textbf {\bibinfo {volume} {15}},\ \bibinfo {pages}
  {620} (\bibinfo {year} {2024})}\BibitemShut {NoStop}%
\bibitem [{\citenamefont {{Zhang, M. (L.), Levy, J., d'Ascoli, S., Rapin, J.,
  Alario, F.-X., Bourdillon, P., Pinet, S., \& King, J.
  R.}}(2025)}]{zhang_m_l_levy_j_dascoli_s_rapin_j_alario_f-x_bourdillon_p_pinet_s__king_j_r_thought_2025}%
  \BibitemOpen
  \bibfield  {author} {\bibinfo {author} {\bibnamefont {{Zhang, M. (L.), Levy,
  J., d'Ascoli, S., Rapin, J., Alario, F.-X., Bourdillon, P., Pinet, S., \&
  King, J. R.}}},\ }\href@noop {} {\bibinfo {title} {From thought to action:
  {How} a hierarchy of neural dynamics supports language productio}} (\bibinfo
  {year} {2025})\BibitemShut {NoStop}%
\bibitem [{\citenamefont {Lévy}\ \emph {et~al.}(2025)\citenamefont {Lévy},
  \citenamefont {Zhang}, \citenamefont {Pinet}, \citenamefont {Rapin},
  \citenamefont {Banville}, \citenamefont {d'Ascoli},\ and\ \citenamefont
  {King}}]{levy_brain--text_nodate}%
  \BibitemOpen
  \bibfield  {author} {\bibinfo {author} {\bibfnamefont {J.}~\bibnamefont
  {Lévy}}, \bibinfo {author} {\bibfnamefont {M.}~\bibnamefont {Zhang}},
  \bibinfo {author} {\bibfnamefont {S.}~\bibnamefont {Pinet}}, \bibinfo
  {author} {\bibfnamefont {J.}~\bibnamefont {Rapin}}, \bibinfo {author}
  {\bibfnamefont {H.}~\bibnamefont {Banville}}, \bibinfo {author}
  {\bibfnamefont {S.}~\bibnamefont {d'Ascoli}},\ and\ \bibinfo {author}
  {\bibfnamefont {J.-R.}\ \bibnamefont {King}},\ }\href
  {https://doi.org/10.48550/arXiv.2502.17480} {\bibinfo {title}
  {Brain-to-{Text} {Decoding}: {A} {Non}-invasive {Approach} via {Typing}}}
  (\bibinfo {year} {2025}),\ \bibinfo {note} {arXiv:2502.17480
  [eess]}\BibitemShut {NoStop}%
\bibitem [{\citenamefont {Goldstein}\ \emph {et~al.}(2025)\citenamefont
  {Goldstein}, \citenamefont {Wang}, \citenamefont {Niekerken}, \citenamefont
  {Schain}, \citenamefont {Zada}, \citenamefont {Aubrey}, \citenamefont
  {Sheffer}, \citenamefont {Nastase}, \citenamefont {Gazula}, \citenamefont
  {Singh}, \citenamefont {Rao}, \citenamefont {Choe}, \citenamefont {Kim},
  \citenamefont {Doyle}, \citenamefont {Friedman}, \citenamefont {Devore},
  \citenamefont {Dugan}, \citenamefont {Hassidim}, \citenamefont {Brenner},
  \citenamefont {Matias}, \citenamefont {Devinsky}, \citenamefont {Flinker},\
  and\ \citenamefont {Hasson}}]{goldstein_unified_2025}%
  \BibitemOpen
  \bibfield  {author} {\bibinfo {author} {\bibfnamefont {A.}~\bibnamefont
  {Goldstein}}, \bibinfo {author} {\bibfnamefont {H.}~\bibnamefont {Wang}},
  \bibinfo {author} {\bibfnamefont {L.}~\bibnamefont {Niekerken}}, \bibinfo
  {author} {\bibfnamefont {M.}~\bibnamefont {Schain}}, \bibinfo {author}
  {\bibfnamefont {Z.}~\bibnamefont {Zada}}, \bibinfo {author} {\bibfnamefont
  {B.}~\bibnamefont {Aubrey}}, \bibinfo {author} {\bibfnamefont
  {T.}~\bibnamefont {Sheffer}}, \bibinfo {author} {\bibfnamefont {S.~A.}\
  \bibnamefont {Nastase}}, \bibinfo {author} {\bibfnamefont {H.}~\bibnamefont
  {Gazula}}, \bibinfo {author} {\bibfnamefont {A.}~\bibnamefont {Singh}},
  \bibinfo {author} {\bibfnamefont {A.}~\bibnamefont {Rao}}, \bibinfo {author}
  {\bibfnamefont {G.}~\bibnamefont {Choe}}, \bibinfo {author} {\bibfnamefont
  {C.}~\bibnamefont {Kim}}, \bibinfo {author} {\bibfnamefont {W.}~\bibnamefont
  {Doyle}}, \bibinfo {author} {\bibfnamefont {D.}~\bibnamefont {Friedman}},
  \bibinfo {author} {\bibfnamefont {S.}~\bibnamefont {Devore}}, \bibinfo
  {author} {\bibfnamefont {P.}~\bibnamefont {Dugan}}, \bibinfo {author}
  {\bibfnamefont {A.}~\bibnamefont {Hassidim}}, \bibinfo {author}
  {\bibfnamefont {M.}~\bibnamefont {Brenner}}, \bibinfo {author} {\bibfnamefont
  {Y.}~\bibnamefont {Matias}}, \bibinfo {author} {\bibfnamefont
  {O.}~\bibnamefont {Devinsky}}, \bibinfo {author} {\bibfnamefont
  {A.}~\bibnamefont {Flinker}},\ and\ \bibinfo {author} {\bibfnamefont
  {U.}~\bibnamefont {Hasson}},\ }\bibfield  {title} {{\selectlanguage
  {English}\bibinfo {title} {A unified acoustic-to-speech-to-language embedding
  space captures the neural basis of natural language processing in everyday
  conversations}},\ }\href {https://doi.org/10.1038/s41562-025-02105-9}
  {\bibfield  {journal} {\bibinfo  {journal} {Nat Hum Behav}\ ,\ \bibinfo
  {pages} {1}} (\bibinfo {year} {2025})},\ \bibinfo {note} {publisher: Nature
  Publishing Group}\BibitemShut {NoStop}%
\bibitem [{\citenamefont {Kuhnke}\ \emph {et~al.}(2022)\citenamefont {Kuhnke},
  \citenamefont {Beaupain}, \citenamefont {Arola}, \citenamefont {Kiefer},\
  and\ \citenamefont {Hartwigsen}}]{kuhnke_meta-analytic_2022}%
  \BibitemOpen
  \bibfield  {author} {\bibinfo {author} {\bibfnamefont {P.}~\bibnamefont
  {Kuhnke}}, \bibinfo {author} {\bibfnamefont {M.~C.}\ \bibnamefont
  {Beaupain}}, \bibinfo {author} {\bibfnamefont {J.}~\bibnamefont {Arola}},
  \bibinfo {author} {\bibfnamefont {M.}~\bibnamefont {Kiefer}},\ and\ \bibinfo
  {author} {\bibfnamefont {G.}~\bibnamefont {Hartwigsen}},\ }\href
  {https://doi.org/10.1101/2022.11.05.515278} {{\selectlanguage
  {English}\bibinfo {title} {Meta-analytic evidence for a novel hierarchical
  model of conceptual processing}}} (\bibinfo {year} {2022})\BibitemShut
  {NoStop}%
\bibitem [{\citenamefont {Mongillo}\ and\ \citenamefont
  {Tsodyks}(2024)}]{mongillo_synaptic_2024}%
  \BibitemOpen
  \bibfield  {author} {\bibinfo {author} {\bibfnamefont {G.}~\bibnamefont
  {Mongillo}}\ and\ \bibinfo {author} {\bibfnamefont {M.}~\bibnamefont
  {Tsodyks}},\ }\href {https://doi.org/10.1101/2024.01.11.575157}
  {{\selectlanguage {English}\bibinfo {title} {Synaptic {Theory} of {Working}
  {Memory} for {Serial} {Order}}}} (\bibinfo {year} {2024})\BibitemShut
  {NoStop}%
\bibitem [{\citenamefont {Schaat}\ \emph {et~al.}(2014)\citenamefont {Schaat},
  \citenamefont {Wendt}, \citenamefont {Jakubec}, \citenamefont {Gelbard},
  \citenamefont {Herret},\ and\ \citenamefont {Dietrich}}]{schaat_ars_2014}%
  \BibitemOpen
  \bibfield  {author} {\bibinfo {author} {\bibfnamefont {S.}~\bibnamefont
  {Schaat}}, \bibinfo {author} {\bibfnamefont {A.}~\bibnamefont {Wendt}},
  \bibinfo {author} {\bibfnamefont {M.}~\bibnamefont {Jakubec}}, \bibinfo
  {author} {\bibfnamefont {F.}~\bibnamefont {Gelbard}}, \bibinfo {author}
  {\bibfnamefont {L.}~\bibnamefont {Herret}},\ and\ \bibinfo {author}
  {\bibfnamefont {D.}~\bibnamefont {Dietrich}},\ }\bibfield  {title}
  {{\selectlanguage {English}\bibinfo {title} {{ARS}: {An} {AGI} {Agent}
  {Architecture}}},\ }in\ \href {https://doi.org/10.1007/978-3-319-09274-4_15}
  {{\selectlanguage {English}\emph {\bibinfo {booktitle} {Artificial {General}
  {Intelligence}}}}},\ \bibinfo {editor} {edited by\ \bibinfo {editor}
  {\bibfnamefont {B.}~\bibnamefont {Goertzel}}, \bibinfo {editor}
  {\bibfnamefont {L.}~\bibnamefont {Orseau}},\ and\ \bibinfo {editor}
  {\bibfnamefont {J.}~\bibnamefont {Snaider}}}\ (\bibinfo  {publisher}
  {Springer International Publishing},\ \bibinfo {address} {Cham},\ \bibinfo
  {year} {2014})\ pp.\ \bibinfo {pages} {155--164}\BibitemShut {NoStop}%
\bibitem [{\citenamefont {Aminifar}\ \emph {et~al.}(2024)\citenamefont
  {Aminifar}, \citenamefont {Huang}, \citenamefont {Abtahi},\ and\
  \citenamefont {Aminifar}}]{aminifar_lightff_2024}%
  \BibitemOpen
  \bibfield  {author} {\bibinfo {author} {\bibfnamefont {A.}~\bibnamefont
  {Aminifar}}, \bibinfo {author} {\bibfnamefont {B.}~\bibnamefont {Huang}},
  \bibinfo {author} {\bibfnamefont {A.}~\bibnamefont {Abtahi}},\ and\ \bibinfo
  {author} {\bibfnamefont {A.}~\bibnamefont {Aminifar}},\ }\href
  {https://doi.org/10.48550/ARXIV.2404.05241} {\bibinfo {title} {{LightFF}:
  {Lightweight} {Inference} for {Forward}-{Forward} {Algorithm}}} (\bibinfo
  {year} {2024}),\ \bibinfo {note} {version Number: 6}\BibitemShut {NoStop}%
\bibitem [{\citenamefont {Sekar}\ \emph {et~al.}(2024)\citenamefont {Sekar},
  \citenamefont {Cantwell}, \citenamefont {Liao}, \citenamefont {Berton},
  \citenamefont {Jacquart},\ and\ \citenamefont {Abu
  Al-Rub}}]{sekar_additively_2024}%
  \BibitemOpen
  \bibfield  {author} {\bibinfo {author} {\bibfnamefont {V.}~\bibnamefont
  {Sekar}}, \bibinfo {author} {\bibfnamefont {W.~J.}\ \bibnamefont {Cantwell}},
  \bibinfo {author} {\bibfnamefont {K.}~\bibnamefont {Liao}}, \bibinfo {author}
  {\bibfnamefont {B.}~\bibnamefont {Berton}}, \bibinfo {author} {\bibfnamefont
  {P.-M.}\ \bibnamefont {Jacquart}},\ and\ \bibinfo {author} {\bibfnamefont
  {R.~K.}\ \bibnamefont {Abu Al-Rub}},\ }\bibfield  {title} {\bibinfo {title}
  {Additively manufactured metamaterials for acoustic absorption: a review},\
  }\href {https://doi.org/10.1080/17452759.2024.2435562} {\bibfield  {journal}
  {\bibinfo  {journal} {Virtual and Physical Prototyping}\ }\textbf {\bibinfo
  {volume} {19}},\ \bibinfo {pages} {e2435562} (\bibinfo {year} {2024})},\
  \bibinfo {note} {publisher: Taylor \& Francis \_eprint:
  https://doi.org/10.1080/17452759.2024.2435562}\BibitemShut {NoStop}%
\bibitem [{\citenamefont {Durney}\ \emph {et~al.}(2023)\citenamefont {Durney},
  \citenamefont {Wilson}, \citenamefont {McGregor}, \citenamefont {Armand},
  \citenamefont {Smith}, \citenamefont {Gray}, \citenamefont {Morris},\ and\
  \citenamefont {Fleming}}]{durney_grasses_2023}%
  \BibitemOpen
  \bibfield  {author} {\bibinfo {author} {\bibfnamefont {C.~H.}\ \bibnamefont
  {Durney}}, \bibinfo {author} {\bibfnamefont {M.~J.}\ \bibnamefont {Wilson}},
  \bibinfo {author} {\bibfnamefont {S.}~\bibnamefont {McGregor}}, \bibinfo
  {author} {\bibfnamefont {J.}~\bibnamefont {Armand}}, \bibinfo {author}
  {\bibfnamefont {R.~S.}\ \bibnamefont {Smith}}, \bibinfo {author}
  {\bibfnamefont {J.~E.}\ \bibnamefont {Gray}}, \bibinfo {author}
  {\bibfnamefont {R.~J.}\ \bibnamefont {Morris}},\ and\ \bibinfo {author}
  {\bibfnamefont {A.~J.}\ \bibnamefont {Fleming}},\ }\bibfield  {title}
  {{\selectlanguage {English}\bibinfo {title} {Grasses exploit geometry to
  achieve improved guard cell dynamics}},\ }\href
  {https://doi.org/10.1016/j.cub.2023.05.051} {\bibfield  {journal} {\bibinfo
  {journal} {Current Biology}\ }\textbf {\bibinfo {volume} {33}},\ \bibinfo
  {pages} {2814} (\bibinfo {year} {2023})}\BibitemShut {NoStop}%
\bibitem [{\citenamefont {Zhang}\ \emph {et~al.}(2018)\citenamefont {Zhang},
  \citenamefont {Butepage}, \citenamefont {Kjellstrom},\ and\ \citenamefont
  {Mandt}}]{zhang_advances_2018}%
  \BibitemOpen
  \bibfield  {author} {\bibinfo {author} {\bibfnamefont {C.}~\bibnamefont
  {Zhang}}, \bibinfo {author} {\bibfnamefont {J.}~\bibnamefont {Butepage}},
  \bibinfo {author} {\bibfnamefont {H.}~\bibnamefont {Kjellstrom}},\ and\
  \bibinfo {author} {\bibfnamefont {S.}~\bibnamefont {Mandt}},\ }\href
  {https://doi.org/10.48550/arXiv.1711.05597} {\bibinfo {title} {Advances in
  {Variational} {Inference}}} (\bibinfo {year} {2018}),\ \bibinfo {note}
  {arXiv:1711.05597 [cs]}\BibitemShut {NoStop}%
\bibitem [{\citenamefont {Yang}(2024)}]{yang_overview_2024}%
  \BibitemOpen
  \bibfield  {author} {\bibinfo {author} {\bibfnamefont {Y.}~\bibnamefont
  {Yang}},\ }\bibfield  {title} {\bibinfo {title} {Overview of the {Current}
  {State} of {Research} on {Metamaterials} in {Biomedicine}},\ }\href
  {https://doi.org/10.1051/bioconf/202414203020} {\bibfield  {journal}
  {\bibinfo  {journal} {BIO Web Conf.}\ }\textbf {\bibinfo {volume} {142}},\
  \bibinfo {pages} {03020} (\bibinfo {year} {2024})}\BibitemShut {NoStop}%
\bibitem [{\citenamefont {Delgado}\ \emph {et~al.}(2023)\citenamefont
  {Delgado}, \citenamefont {Yang}, \citenamefont {Madaio},\ and\ \citenamefont
  {Yang}}]{delgado_participatory_2023}%
  \BibitemOpen
  \bibfield  {author} {\bibinfo {author} {\bibfnamefont {F.}~\bibnamefont
  {Delgado}}, \bibinfo {author} {\bibfnamefont {S.}~\bibnamefont {Yang}},
  \bibinfo {author} {\bibfnamefont {M.}~\bibnamefont {Madaio}},\ and\ \bibinfo
  {author} {\bibfnamefont {Q.}~\bibnamefont {Yang}},\ }\bibfield  {title}
  {{\selectlanguage {English}\bibinfo {title} {The {Participatory} {Turn} in
  {AI} {Design}: {Theoretical} {Foundations} and the {Current} {State} of
  {Practice}}},\ }in\ \href {https://doi.org/10.1145/3617694.3623261}
  {{\selectlanguage {English}\emph {\bibinfo {booktitle} {Equity and {Access}
  in {Algorithms}, {Mechanisms}, and {Optimization}}}}}\ (\bibinfo  {publisher}
  {ACM},\ \bibinfo {address} {Boston MA USA},\ \bibinfo {year} {2023})\ pp.\
  \bibinfo {pages} {1--23}\BibitemShut {NoStop}%
\bibitem [{\citenamefont {Li}\ and\ \citenamefont
  {Collins}(2024)}]{li_algorithmic_2024}%
  \BibitemOpen
  \bibfield  {author} {\bibinfo {author} {\bibfnamefont {J.-J.}\ \bibnamefont
  {Li}}\ and\ \bibinfo {author} {\bibfnamefont {A.}~\bibnamefont {Collins}},\
  }\href {https://doi.org/10.31234/osf.io/b3xnv} {\bibinfo {title} {An
  algorithmic account for how humans efficiently learn, transfer, and compose
  hierarchically structured decision policies}} (\bibinfo {year}
  {2024})\BibitemShut {NoStop}%
\bibitem [{\citenamefont {Buehler}(2025)}]{buehler_agentic_2025}%
  \BibitemOpen
  \bibfield  {author} {\bibinfo {author} {\bibfnamefont {M.~J.}\ \bibnamefont
  {Buehler}},\ }\href {https://doi.org/10.48550/arXiv.2502.13025} {\bibinfo
  {title} {Agentic {Deep} {Graph} {Reasoning} {Yields} {Self}-{Organizing}
  {Knowledge} {Networks}}} (\bibinfo {year} {2025}),\ \bibinfo {note}
  {arXiv:2502.13025 [cs]}\BibitemShut {NoStop}%
\bibitem [{\citenamefont {Murray}(2017)}]{murray_stoic_2017}%
  \BibitemOpen
  \bibfield  {author} {\bibinfo {author} {\bibfnamefont {G.}~\bibnamefont
  {Murray}},\ }\href {https://doi.org/10.48550/arXiv.1701.02388} {\bibinfo
  {title} {Stoic {Ethics} for {Artificial} {Agents}}} (\bibinfo {year}
  {2017}),\ \bibinfo {note} {arXiv:1701.02388 [cs]}\BibitemShut {NoStop}%
\bibitem [{\citenamefont {{Giovanni Sileno and Matteo
  Pascucc}}(2020)}]{giovanni_sileno_and_matteo_pascucc_disentangling_2020}%
  \BibitemOpen
  \bibfield  {author} {\bibinfo {author} {\bibnamefont {{Giovanni Sileno and
  Matteo Pascucc}}},\ }\bibfield  {title} {\bibinfo {title} {Disentangling
  {Deontic} {Positions} and {Abilities}: a {Modal} {Analysis}},\ }\href
  {https://ceur-ws.org/Vol-2710/paper3.pdf} {\bibfield  {journal} {\bibinfo
  {journal} {Italian Conference on Computational Logic}\ } (\bibinfo {year}
  {2020})}\BibitemShut {NoStop}%
\bibitem [{\citenamefont {Lucentini}\ and\ \citenamefont
  {Gudwin}(2015)}]{lucentini_comparison_2015}%
  \BibitemOpen
  \bibfield  {author} {\bibinfo {author} {\bibfnamefont {D.~F.}\ \bibnamefont
  {Lucentini}}\ and\ \bibinfo {author} {\bibfnamefont {R.~R.}\ \bibnamefont
  {Gudwin}},\ }\bibfield  {title} {\bibinfo {title} {A {Comparison} {Among}
  {Cognitive} {Architectures}: {A} {Theoretical} {Analysis}},\ }\href
  {https://doi.org/10.1016/j.procs.2015.12.198} {\bibfield  {journal} {\bibinfo
   {journal} {Procedia Computer Science}\ }\bibinfo {series} {6th {Annual}
  {International} {Conference} on {Biologically} {Inspired} {Cognitive}
  {Architectures}, {BICA} 2015, 6-8 {November} {Lyon}, {France}},\ \textbf
  {\bibinfo {volume} {71}},\ \bibinfo {pages} {56} (\bibinfo {year}
  {2015})}\BibitemShut {NoStop}%
\bibitem [{\citenamefont {Sukhobokov}\ \emph {et~al.}(2024)\citenamefont
  {Sukhobokov}, \citenamefont {Belousov}, \citenamefont {Gromozdov},
  \citenamefont {Zenger},\ and\ \citenamefont
  {Popov}}]{sukhobokov_universal_2024}%
  \BibitemOpen
  \bibfield  {author} {\bibinfo {author} {\bibfnamefont {A.}~\bibnamefont
  {Sukhobokov}}, \bibinfo {author} {\bibfnamefont {E.}~\bibnamefont
  {Belousov}}, \bibinfo {author} {\bibfnamefont {D.}~\bibnamefont {Gromozdov}},
  \bibinfo {author} {\bibfnamefont {A.}~\bibnamefont {Zenger}},\ and\ \bibinfo
  {author} {\bibfnamefont {I.}~\bibnamefont {Popov}},\ }\bibfield  {title}
  {\bibinfo {title} {A universal knowledge model and cognitive architectures
  for prototyping {AGI}},\ }\href
  {https://doi.org/10.1016/j.cogsys.2024.101279} {\bibfield  {journal}
  {\bibinfo  {journal} {Cognitive Systems Research}\ }\textbf {\bibinfo
  {volume} {88}},\ \bibinfo {pages} {101279} (\bibinfo {year}
  {2024})}\BibitemShut {NoStop}%
\bibitem [{\citenamefont {Bringsjord}\ \emph {et~al.}(2018)\citenamefont
  {Bringsjord}, \citenamefont {Govindarajulu}, \citenamefont {Sen},
  \citenamefont {Peveler}, \citenamefont {Srivastava},\ and\ \citenamefont
  {Talamadupula}}]{bringsjord_tentacular_2018}%
  \BibitemOpen
  \bibfield  {author} {\bibinfo {author} {\bibfnamefont {S.}~\bibnamefont
  {Bringsjord}}, \bibinfo {author} {\bibfnamefont {N.~S.}\ \bibnamefont
  {Govindarajulu}}, \bibinfo {author} {\bibfnamefont {A.}~\bibnamefont {Sen}},
  \bibinfo {author} {\bibfnamefont {M.}~\bibnamefont {Peveler}}, \bibinfo
  {author} {\bibfnamefont {B.}~\bibnamefont {Srivastava}},\ and\ \bibinfo
  {author} {\bibfnamefont {K.}~\bibnamefont {Talamadupula}},\ }\href
  {https://doi.org/10.48550/arXiv.1810.07007} {\bibinfo {title} {Tentacular
  {Artificial} {Intelligence}, and the {Architecture} {Thereof}, {Introduced}}}
  (\bibinfo {year} {2018}),\ \bibinfo {note} {arXiv:1810.07007
  [cs]}\BibitemShut {NoStop}%
\end{thebibliography}

%

\end{document}